\documentclass[10pt,twocolumn,letterpaper]{article}

\usepackage{iccv}
\usepackage{times}
\usepackage{epsfig}
\usepackage{graphicx}
\usepackage{amsmath}
\usepackage{amssymb}

\usepackage{booktabs}
\usepackage{multirow, makecell}
\usepackage{subcaption}
\usepackage{float}
\usepackage[dvipsnames]{xcolor}
\usepackage{rotating}
\usepackage{nicefrac}
\usepackage{pifont}
\usepackage{placeins}
\usepackage{url}

\usepackage[pagebackref=true,breaklinks=false,letterpaper=true,colorlinks,bookmarks=false]{hyperref}

\iccvfinalcopy %

\def\iccvPaperID{3246} %

\ificcvfinal\fi

\newenvironment{tight_itemize}{
	\begin{itemize}
		\setlength{\itemsep}{0pt}
		\setlength{\parskip}{0pt}
	}{\end{itemize}}

\newlength{\tabcolsepdefault}
\newcommand{\cmark}{\ding{51}}%
\newcommand{\xmark}{\ding{55}}%

\DeclareMathOperator*{\argmin}{arg\,min}

\newcommand\customparagraph[1]{\vspace{0.4em}\noindent\textbf{#1}}

\begin{document}
	
	\title{Cross-Descriptor Visual Localization and Mapping}
	
	\author{Mihai Dusmanu \textsuperscript{1}
		\and
		Ondrej Miksik \textsuperscript{2}
		\and
		Johannes L. Sch\"onberger \textsuperscript{2}
		\and
		Marc Pollefeys \textsuperscript{1, 2}
		\and
		\normalsize\textsuperscript{1} Department of Computer Science, ETH Z\"urich
		\and
		\normalsize\textsuperscript{2} Microsoft MR \& AI Lab, Z\"urich
	}
	
	\maketitle

	\begin{abstract}
		Visual localization and mapping is the key technology underlying the majority of mixed reality and robotics systems.
		Most state-of-the-art approaches rely on local features to establish correspondences between images.
		In this paper, we present three novel scenarios for localization and mapping which require the continuous update of feature representations and the ability to match across different feature types.
		While localization and mapping is a fundamental computer vision problem, the traditional setup supposes the same local features are used throughout the evolution of a map.
		Thus, whenever the underlying features are changed, the whole process is repeated from scratch.
		However, this is typically impossible in practice, because raw images are often not stored and re-building the maps could lead to loss of the attached digital content.
		To overcome the limitations of current approaches, we present the first principled solution to cross-descriptor localization and mapping.
		Our data-driven approach is agnostic to the feature descriptor type, has low computational requirements, and scales linearly with the number of description algorithms.
		Extensive experiments demonstrate the effectiveness of our approach on state-of-the-art benchmarks for a variety of handcrafted and learned features.
	\end{abstract}
	
	\vspace{-10pt}
	\section{Introduction}

\begin{figure}[t]
	\centering
	\includegraphics[width=.95\columnwidth]{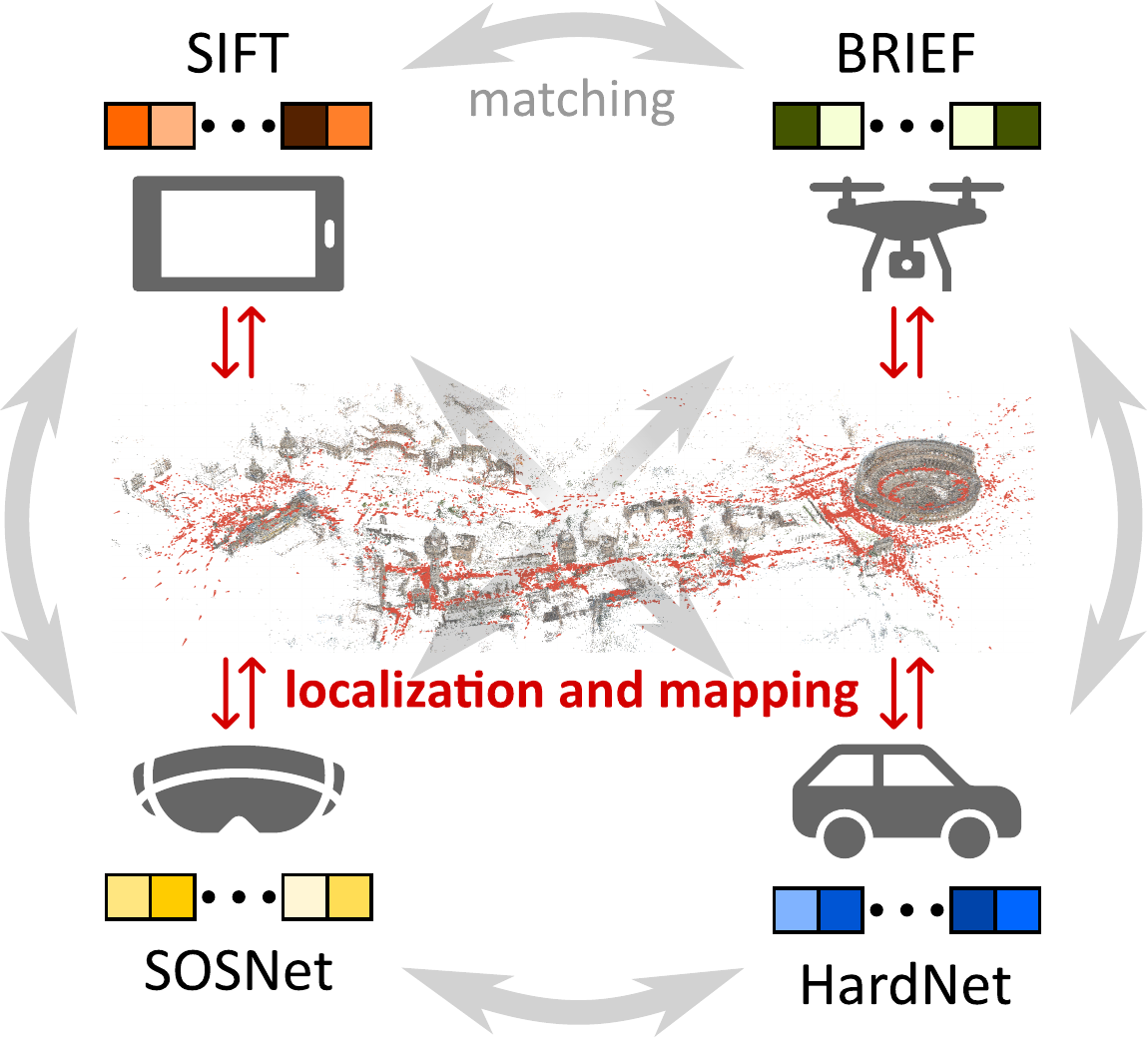}
	\vspace{-7.5pt}
	\caption{
		Mixed reality and robotics systems typically do not store raw images. 
		This complicates deployment of new feature representations for visual localization and mapping.
		Our approach enables the continuous update of feature representations and the ability to match across heterogeneous devices using different features.
		}
	\label{fig:teaser}
	\vspace{-10pt}
\end{figure}

Mixed reality and robotics blend the physical and digital worlds to unlock the next evolution in human, machine, and environment interaction.
This promise largely relies on the technical capability to build and localize against maps of the environment.
For example, collaborative experiences require co-localization of multiple devices into the same coordinate space.
Similarly, re-localization against existing maps enables persistence and retrieval of digital content attached to the real world.
Most systems build upon localization and mapping pipelines predominantly based on vision sensors using local features for image matching.

Traditionally, these systems run in real-time on the device, but the industry has been increasingly moving localization and mapping capabilities to the cloud (\eg, Facebook LiveMaps~\cite{Facebook2019Inside}, Google VPS~\cite{Google2019Google}, Magic Leap's Magicverse~\cite{MagicLeap2019What}, or Microsoft Azure Spatial Anchors~\cite{Microsoft2019Announcing}).
This is for a variety of reasons, such as reducing on-board compute or enabling collaborative experiences and crowd-sourced mapping.
In most settings, images are typically not shared across devices or uploaded to the cloud primarily due to privacy reasons~\cite{Kipman2019Azure}, but also to reduce bandwidth and storage requirements.
Instead, image features are computed locally on the device and only the extracted keypoints and descriptors are shared.
In other words, it is impossible to re-extract features whenever a different representation is required, as the images are no longer available.

There are two fundamental limitations of existing systems. 
First, they cannot adopt new feature algorithms because incompatibilities of feature representations make re-localization against maps built using older features impossible.
In a world of continuous progress on local features in the research community, this severely limits progress on localization and mapping.
One might argue for simply re-building the maps from scratch whenever a significantly improved feature algorithm is available.
However, content attached to the old maps would be lost and mapping is an expensive process, where it could take weeks or even months until the whole area is re-visited.
Second, co-localization and collaborative mapping scenarios with devices using different features is impossible.
The situation is further complicated by the fact that many devices implement specific algorithms in hardware for efficiency reasons, making a client-side upgrade of algorithms impossible.
This also means that many of the existing commercial solutions might be significantly behind the state-of-the-art art in the research community, because they cannot easily upgrade their algorithms and representations.

\begin{figure*}[t!]
    \centering
    \hfill
    \begin{subfigure}{0.40\textwidth}
		\centering
        \includegraphics[width=\textwidth]{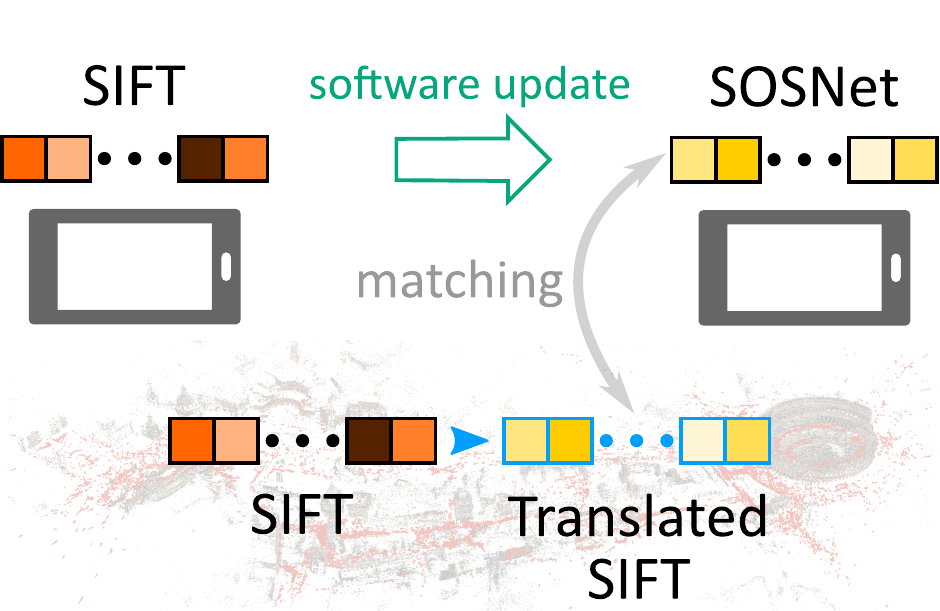}
		\caption{}
    \end{subfigure}
	\hfill
    \begin{subfigure}{0.40\textwidth}
		\centering
        \includegraphics[width=\textwidth]{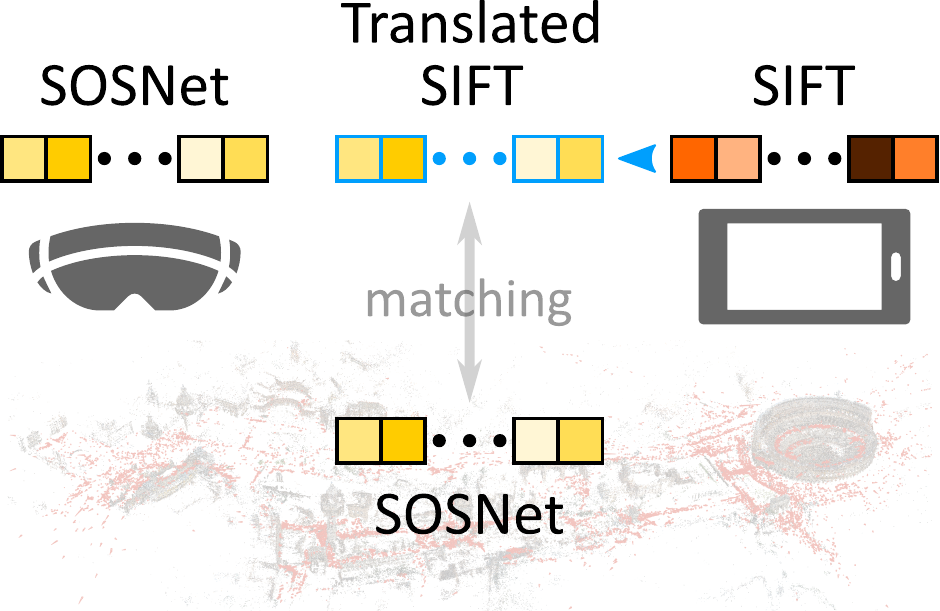}
		\caption{}
    \end{subfigure}
	\hfill
	\vspace{-7.5pt}
	\caption{{\bf Continuous deployment and cross-device localization.} (a) Translating the local descriptors associated with the map enables continuous deployment of new feature representations. 
    (b) Translating the query descriptors enables cross-device localization when queries and map use different feature representations.}
	\label{fig:scenarios}
	\vspace{-10pt}
\end{figure*}

In this paper, we first define three novel scenarios addressing the challenges in a world of changing local feature representations (\cf~Figures~\ref{fig:teaser} and~\ref{fig:scenarios}):
\vspace{-5pt}
\begin{tight_itemize} 
	\item \emph{Continuous deployment} of feature representations without requiring an explicit re-mapping phase.
	\item \emph{Cross-device localization} when the localization and mapping devices use different features.
	\item \emph{Collaborative mapping} with multiple heterogeneous devices and features.
\end{tight_itemize}
\vspace{-5pt}
Note that these scenarios are completely different from existing industrial or academic setups, where generally a \emph{fixed} local feature extraction algorithm is used for co-localization and throughout the evolution of a map.
Section~\ref{sec:scenarios} introduces the scenarios in more detail.

As a first step towards enabling these scenarios, we focus on local feature descriptors and propose what to our knowledge is the first principled and scalable approach to the underlying challenges.
Our learned approach translates descriptors from one representation to another without any assumptions about the structure of the feature vectors and enables matching of features with incompatible dimensionality as well as distance metrics.
For instance, we can match 512 dimensional binary BRIEF~\cite{Calonder2010BRIEF} against 128 dimensional floating point SIFT~\cite{Lowe2004Distinctive} or even deep learning models such as HardNet~\cite{Mishchuk2017Working} or SOSNet~\cite{Tian2019SOSNet} and vice versa.
Our method has linear scalability in the number of algorithms and is specifically designed to have a small computational footprint to enable deployment on low-compute devices.
The training data is generated automatically by computing different descriptors from the same image patches.

We evaluate our approach on relevant geometric tasks in the context of the newly proposed scenarios.
We first consider localization (pose estimation) on the Aachen Day \& Night benchmark~\cite{Sattler2017Benchmarking}.
Next, we assess the performance for 3D mapping from crowd-sourced images using the benchmark of Sch\"{o}nberger~\etal~\cite{Schonberger2017Comparative}.
In the supplementary material, we also show results on the HPatches descriptor benchmark~\cite{Balntas2017HPatches}, the Image Matching Workshop challenge~\cite{Jin2020Image}, and the InLoc Indoor Visual Localization dataset~\cite{Taira2018InLoc}.
Our experiments demonstrate the effectiveness and high practical relevance of our method for real-world localization and mapping systems.

To summarize our contributions, we i) introduce three novel scenarios to localization and mapping in a world of changing feature representations, ii) propose the first principled and scalable approach tackling these newly introduced scenarios, and iii) demonstrate the effectiveness of our method on challenging real-world datasets.

\section{Related work}

\customparagraph{Local image descriptors.}
Local descriptors are typically extracted from normalized images patches defined by local feature frames~\cite{Lowe2004Distinctive, Mikolaczyk2003Scale,Rublee2011ORB}. 
The most notable handcrafted descriptors include binary BRIEF~\cite{Calonder2010BRIEF}, gradient-based SIFT~\cite{Lowe2004Distinctive}, and their variants~\cite{Rublee2011ORB,Mikolajczyk2005Performance,Van2009Evaluating,Arandjelovic2012Three}. 
Recently, the community has moved to data-driven models leveraging large-scale datasets, various triplet losses, and hard-negative mining techniques~\cite{Balntas2016Learning,Mishchuk2017Working,Luo2018GeoDesc,Tian2019SOSNet,Ebel2019Beyond}.
A related line of research aims to reformulate the entire pipeline in an end-to-end trainable fashion~\cite{Yi2016LIFT,Ono2019LFNet}.
Finally, the describe-then-detect approach extracts local features from dense feature maps~\cite{Noh2017Largescale,DeTone2018SuperPoint,Dusmanu2019D2,Revaud2019R2D2,Luo2020ASLFeat}.
While it is possible to handcraft relations between some of these descriptors (\eg SIFT~\cite{Lowe2004Distinctive} $\leftrightarrow$ \mbox{Hue-SIFT}~\cite{Van2009Evaluating}), they are typically not designed for inter-compatibility.
In contrast, our approach aims to enable matching of local descriptors with incompatible feature representations, dimensionalities or metrics.

\customparagraph{Visual localization and mapping.}
Given the efficiency of local features and the well-established theory of (sparse) multi-view geometry, a majority of large-scale mapping~\cite{Agarwal2011Building,Heinly2015Reconstructing} and localization~\cite{Li2012Worldwide,Sattler2012Improving} is based on local image features.
Furthermore, there is a growing number of open-source tools for Simultaneous Localization and Mapping (\eg, ORB-SLAM~\cite{Mur2015ORBSLAM,Mur2017ORBSLAM2}), visual odometry (\eg, LIBVISO~\cite{Kitt2010Visual,Geiger2011StereoScan}) or Structure-from-Motion (\eg, COLMAP~\cite{Schoenberger2016Structure}, OpenMVG~\cite{MoulonOpenMVG}, Theia~\cite{SweeneyTheia}) rendering this research domain more accessible.
Recently, these pipelines were adopted for evaluation of visual localization and mapping -- Sch\"onberger~\etal~\cite{Schonberger2017Comparative} and Jin~\etal~\cite{Jin2020Image} considered mapping and localization performance using photo-tourism collections, while Sattler~\etal~\cite{Sattler2017Benchmarking} introduced several datasets for localization under challenging scenarios (\eg, seasonal / temporal changes).
Also related to our work, the problem of collaborative mapping between multiple agents (\eg, robots, drones) has been thoroughly studied~\cite{Fox2000Probabilistic,Cieslewski2015Map,Morrison2016MOARSLAM} in the field of robotics.
However, all these systems and associated benchmarks use the same local image features throughout the whole process and are thus unable to tackle any of the scenarios discussed in Section~\ref{sec:scenarios}.

\customparagraph{Domain translation.}
Several local feature inversion techniques~\cite{Dosovitskiy2016Inverting,Pittaluga2019} reconstructing the original image from its features were proposed.
One way to enable descriptor translation would be to reconstruct the image and then follow the traditional pipeline to extract a different descriptor. 
In our early experiments with feature inversion networks, we noticed significant limitations in generalizing to new scenes and poor quality of recovered low-level gradient information.
As such, it was not possible to successfully match images using this feature inversion based approach.

Image-to-image translation~\cite{Zhu2016Generative,Isola2017Image,Zhu2017Unpaired} attempts to convert one representation of a scene to another (\eg, RGB image to semantic labels).
One could imagine adapting these models to translate between different local feature representations.
However, similarly to feature inversion techniques, these approaches typically have very large computational and memory footprint, as they employ deep convolutional neural networks to produce full resolution images.
In contrast, we propose to use shallow multi-layer perceptrons, taking a single descriptor as input and predicting its translation -- a solution more suitable for low-compute devices.

Finally, research on domain adaptation (\cf surveys of Csurka~\cite{Csurka2017Comprehensive} and Zhuang~\etal~\cite{Zhuang2019Comprehensive}) tackles the problem of adapting algorithms learned on a source data distribution to a dataset with a related but different distribution.
In the case of deep learning, this is particularly interesting when the source domain has enough annotations to allow training while the target domain has little to no annotations available.
In our case, instead of adapting to a different input data distribution, we try to adapt the output distributions to make them compatible for matching.

	\section{Scenarios}
\label{sec:scenarios}

As our first contribution, we identify, introduce, and formalize three novel scenarios for localization and mapping.
In all scenarios, we assume there are no images stored in the resulting maps.
Therefore, it is impossible to re-extract or replace the underlying feature representation. 

\subsection{Continuous deployment}
\label{sec:scenarios-continuous}

Given a sparse 3D map, associated local feature descriptors and no access to the original image data, the goal is to develop a mechanism enabling the continuous deployment of new feature representations (Figure~\ref{fig:scenarios} (a)) without the need for re-mapping.
In other words, we aim to \emph{translate} the features of the map from one representation to another.
For instance, it should be possible to switch from handcrafted to learned descriptors (\eg, SIFT~\cite{Lowe2004Distinctive} $\rightarrow$ SOSNet~\cite{Tian2019SOSNet}), change their dimensionality, or update the model weights (\eg, different training data, loss, or architecture).

Matching the new feature representation against a database of translated descriptors might lead to worse performance when compared to matching using a single feature representation.
This is, however, only a transient issue and not of serious concern in practice, since the translated descriptors can be used in a boot-strapping fashion -- as users re-visit the environment, the translated descriptors are gradually replaced by the newly extracted ones.
Thus, the map will eventually contain the latest descriptors to close the performance gap.
The key points are that i) the devices run only a single algorithm (due to real-time constraints) or cannot be easily updated (due to hardware implementation), ii) the scenario avoids an explicit re-mapping phase, which is generally very expensive and time-consuming. %

\subsection{Cross-device localization}
The previous scenario assumes the same feature extraction algorithm is running on all client devices.
However, this is generally not the case, as legacy devices often cannot be updated due to hardware limitations.
In this case, a mechanism for backwards-compatibility is required to match across different device versions (see Figure~\ref{fig:scenarios} (b)).
Similarly, specialized devices (\eg, headsets, autonomous cars, mapping platforms) could take advantage of additional on-board compute for a better localization performance compared to devices such as mobile phones or light-weight drones.
It would be beneficial to co-localize these devices inside the same maps to avoid a fragmented scene representation.
The same functionality could be extended to enable localization between devices of different vendors, but, in this situation, new challenges arise due to differences in previous steps of the pipeline -- notably feature detection, as discussed in detail in Section~\ref{sec:limitations}.

\subsection{Collaborative mapping} 
\label{sec:crowdmapping}
Finally, we consider multiple heterogeneous devices \emph{collaboratively} mapping the same environment.
For the aforementioned reasons, it is fairly unlikely for all devices to use the same features as in a standard academic setup, even when produced by a single vendor.
Consequently, we end up with each device providing different types of features from which we need to build a single coherent map.
Therefore, we need a mechanism for translating all descriptors into a common representation that can be used for establishing correspondences (see Figure~\ref{fig:teaser}).
	\section{Method}
\label{sec:method}

\begin{figure*}[t!]
	\includegraphics[width=\textwidth]{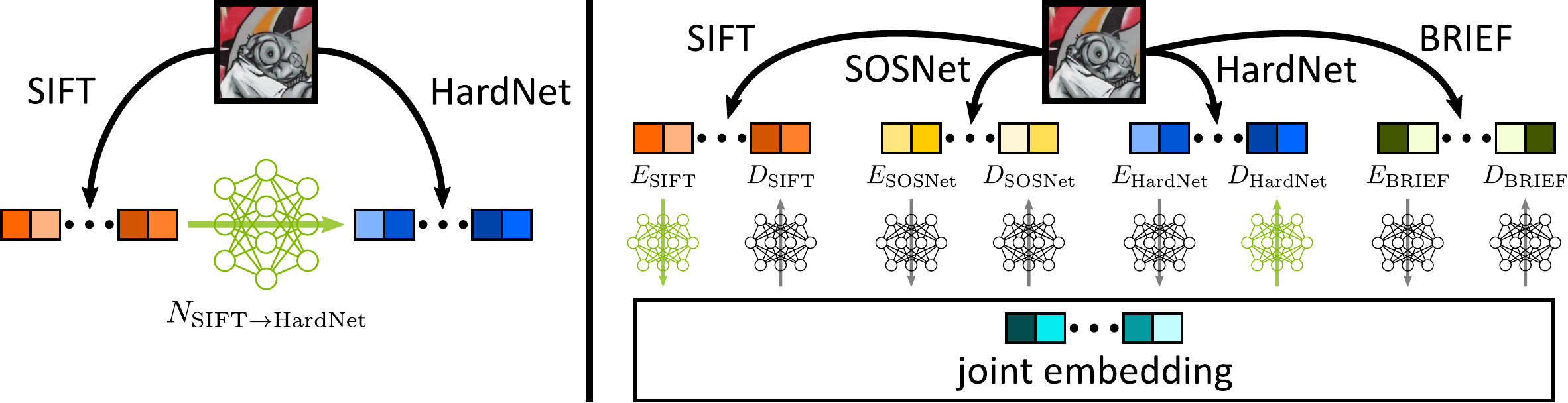}
	\\
	\begin{minipage}{0.34\textwidth}
		\centering
		(a)
	\end{minipage}
	\hfill
	\begin{minipage}{0.595\textwidth}
		\centering
		(b)
	\end{minipage}
	\vspace{-7.5pt}
	\caption{{\bf Method overview.} (a) The pair network is trained for each description algorithm pair independently. (b) The encoder-decoder network is trained for all description algorithms at once. All descriptors are mapped to a joint embedding space. In \textcolor{OliveGreen}{green}, we highlight the networks that need to be used in order to translate from SIFT to HardNet.}
	\label{fig:method}
	\vspace{-10pt}
\end{figure*}

In this section, we present the first principled solution to cross-descriptor localization and mapping.
We start by formalizing the descriptor translation problem.
We then propose to use a separate multi-layer perceptron (MLP) for each pair of description algorithms, trained using a translation loss (\cf~Figure~\ref{fig:method}~(a)).
Finally, we extend this formulation by using an auto-encoder inspired model that embeds all descriptors from different algorithms into a joint embedding space (\cf~Figure~\ref{fig:method}~(b)).
To make sure that the common embedding is suitable for establishing correspondences between different algorithms, we leverage additional supervision through a matching loss.

\subsection{Descriptor translation}
\label{sec:translation}
A feature description algorithm $A$ defines a handcrafted or learned function mapping images to vectors as $A: \mathcal{I}_{h \times w} \rightarrow \mathbb{R}^n$, where $\mathcal{I}_{h \times w}$ is the set of images of size $h \times w$ and $n$ is the embedding dimension.
In the case of local image descriptors, the domain is the set of patches $\mathcal{I}$ normalized based on the estimated feature geometry (position, scale, orientation, affine shape).

Let $A_1: \mathcal{I} \rightarrow \mathbb{R}^{n_1}$ and $A_2: \mathcal{I} \rightarrow \mathbb{R}^{n_2}$ be two description algorithms.
A translation function $t_{1 \rightarrow 2}: \mathbb{R}^{n_1} \rightarrow \mathbb{R}^{n_2}$ maps a feature vector of $A_1$ onto the manifold of $A_2$, satisfying $t_{1 \rightarrow 2}(A_1(p)) = A_2(p)$ for all patches $p \in \mathcal{I}$.
Our experiments (Section~\ref{sec:evaluation}) empirically demonstrate that using MLPs to approximate the translation functions $t_{1 \rightarrow 2}$ is highly effective in enabling matching between different descriptors which, in turn, facilitates the scenarios introduced in Section~\ref{sec:scenarios}.
Note that we do not propose a matching algorithm but instead directly map descriptors from one space to another.
Our approach thus relies on traditional descriptor comparison strategies, lending itself to efficient matching implementations (\eg, approximate nearest neighbor).
Apart from the corresponding descriptors being extracted from the same patches, we make no additional assumptions regarding the descriptor dimension or the underlying algorithm (handcrafted or learned, binary or floating point).

In the following, we consider a set of description algorithms $\mathcal{A} = \{ A_1, A_2, \dots \}$.
Our training batches consist of normalized images patches $\mathcal{P} = \{p_1, p_2, \dots\}$.
Each description algorithm can be applied independently on a patch $p \in \mathcal{P}$ to obtain the associated descriptor, \ie, $A_i(p) = a_i$.

\subsection{Pair network}
\label{sec:pair}

Given a pair of different description algorithms $(A_i, A_j)$, one could approximate the translation function $t_{i \rightarrow j}$ by a multi-layer perceptron $N_{i \rightarrow j}$ (\cf Figure~\ref{fig:method} (a)).
This model translates features extracted using algorithm $A_i$ into the feature representation of algorithm $A_j$.
To this end, we define the translation loss as the $\ell_2$ prediction error:
\begin{equation}
	\mathcal{L}^T_{i \rightarrow j} = \frac{1}{\lvert \mathcal{P} \rvert} \sum_{p \in \mathcal{P}} \lVert N_{i \rightarrow j} (A_i(p)) - A_j(p) \rVert \enspace .
\end{equation}
If the target algorithm $A_j$ produces binary descriptors, the regression loss is replaced by a classification loss:
\begin{equation}
	\mathcal{L}^T_{i \rightarrow j} = \frac{1}{\lvert \mathcal{P} \rvert} \sum_{p \in \mathcal{P}}  \text{BCE} \left( N_{i \rightarrow j} (A_i(p)), A_j(p) \right) \enspace ,
\end{equation}
where $\text{BCE}$ is the binary cross entropy loss function defined as $\text{BCE} \left( x, y \right) = - y \log (x) - (1 - y) \log (1 - x)$.

To compute the distance $d$ between two descriptors $a_i, a_j$ extracted using algorithms $A_i, A_j$, respectively, one can translate from $A_i$ to $A_j$ by using the associated MLP ($d = \lVert N_{i \rightarrow j} (a_i) - a_j \rVert$) or conversely ($d = \lVert N_{j \rightarrow i} (a_j) - a_i \rVert$).
In most practical applications, the direction of translation is known.
In the case of continuous software updates, the outdated map descriptors are ``migrated" to the new ones, while for cross-device localization the query descriptors are always translated to map descriptors.
However, it is not obvious how to use this approach in the collaborative mapping scenario as there are $\mathcal{O}(\lvert \mathcal{A} \rvert^2)$ possible translation directions.
One could use a ``progressive'' strategy that translates the weaker to the better descriptor for each image pair, but deciding which descriptor is more suitable is not straightforward as it generally depends on the specific scene.\footnote{Please refer to the supplementary material for a scene-by-scene breakdown on the HPatches Sequences dataset~\cite{Balntas2017HPatches} that supports this argument.}

\subsection{Encoder-decoder network}
\label{sec:enc-dec}

To address this issue, we turn our attention towards auto-encoders.
Each description algorithm $A_i$ now has an associated encoder $E_i$ and decoder $D_i$ that ideally satisfy $D_i (E_i (a_i)) = a_i$ for all descriptors $a_i$ extracted using $A_i$.
We propose an extension to the identity loss traditionally used for training auto-encoders~\cite{Bourlard1988Auto}, such that the codes (\ie, the output of encoders / the input of decoders) live in a joint space for all algorithms.
Thus, for a pair of descriptor extractors $A_i, A_j$, one can chain the encoder of $A_i$ and the decoder of $A_j$ to obtain the mapping from $A_i$ to $A_j$ and vice versa, \ie, $N_{i \rightarrow j} = D_j \circ E_i$ and $N_{j \rightarrow i} = D_i \circ E_j$ (see Fig.~\ref{fig:method} (b)).
In this way, we reduce the number of required networks to $\mathcal{O}(\lvert \mathcal{A} \rvert)$ and achieve linear scalability.

Furthermore, we use a triplet ranking loss to facilitate direct matching in the joint embedding space.
Therefore, one can compare two descriptors $a_i, a_j$ extracted using different algorithms $A_i, A_j$ by mapping both descriptors to the joint space and computing the distance as $d = \lVert E_i (a_i) - E_j (a_j)\rVert$, instead of using the directional translation mentioned above.
This is of particular interest in the collaborative mapping scenario, as it offers an elegant way to match descriptors coming from different algorithms by translating everything to the joint embedding space.
Moreover, it allows compatibility with existing pipelines that suppose all images use the same feature.

\customparagraph{Translation loss.}
As above, we consider the translation loss $\mathcal{L}^T_{i \rightarrow j}$ for each algorithm pair $A_i, A_j \in \mathcal{A}^2$.
Note that, for $i = j$, this loss is equivalent to the traditional auto-encoder loss~\cite{Bourlard1988Auto}.
The overall translation loss is defined as the average over all algorithm pairs as:
\begin{equation}
	\mathcal{L}^T = \frac{1}{\lvert \mathcal{A} \rvert^2} \sum_{A_i, A_j \in \mathcal{A}^2} \mathcal{L}^T_{i \rightarrow j} \enspace .
\end{equation}

\customparagraph{Matching loss.} 
We use a triplet margin loss for each description algorithm pair $A_i, A_j \in \mathcal{A}^2$ to allow matching in the joint embedding space:
\begin{equation}
\mathcal{L}^M_{i \rightarrow j} = \frac{1}{\lvert \mathcal{P} \rvert} \sum_{p \in \mathcal{P}} \max \left( m + \text{pos}(p) - \text{neg}(p) , 0 \right) \enspace ,
\end{equation}
where $\text{pos}(p)$ and $\text{neg}(p)$ are the distances to the positive and negative samples, respectively.
Given two corresponding descriptors $A_i(p) = a_i, A_j(p) = a_j$, the positive sample for the embedding of $a_i$ is simply the embedding of $a_j$:
\begin{equation}
	\text{pos}(p) = \lVert E_i (A_i(p)) - E_j (A_j(p)) \rVert \enspace .
\end{equation}
Following HardNet~\cite{Mishchuk2017Working}, we use in-batch hardest negative mining by selecting the closest non-matching embedding of a descriptor extracted using algorithm $A_j$ as negative sample for the embedding of the current descriptor $a_i$:
\begin{equation}
	\text{neg}(p) = \argmin_{p^\prime \in \mathcal{P}, p^\prime \neq p} \lVert E_i (A_i(p)) - E_j (A_j(p^\prime)) \rVert \enspace .
\end{equation}
The overall matching loss is then defined as the sum over all description algorithm pairs as:
\begin{equation}
\mathcal{L}^M = \frac{1}{\lvert \mathcal{A} \rvert^2} \sum_{A_i, A_j \in \mathcal{A}^2} \mathcal{L}^M_{i \rightarrow j} \enspace .
\end{equation}

\customparagraph{Final loss.} The final loss is a weighted sum of the translation and matching losses as:	$\mathcal{L} = \mathcal{L}^T + \alpha \mathcal{L}^M$.

\subsection{Implementation details}

\customparagraph{Training dataset.} Our training dataset contains $7.4$ million patches from $3190$ random internet images of the Oxford-Paris revisited retrieval dataset distractors~\cite{Radenovic2018Revisiting}.
The normalized image patches are extracted around Difference-of-Gaussians (DoG) detections according to the estimated scale and orientation. 
Training descriptors are computed by applying all description algorithms on each patch to obtain a set of corresponding descriptors.
For DoG keypoint extraction we use COLMAP~\cite{Schoenberger2016Structure} with default parameters.

\customparagraph{Training methodology.} We train all networks for $5$ epochs using Adam~\cite{Kingma2014Adam} as optimizer with a learning rate of $10^{-3}$ and a batch size of $1024$.
Similar to Mishchuk~\etal~\cite{Mishchuk2017Working}, we use a margin $m=1$.
According to validation results, we fix the weighting of the matching loss $\alpha=0.1$.
The encoders and decoders are MLPs with $2$ hidden layers of sizes $1024$ and $256$ for hand-crafted and learned features, respectively.
We use ReLU followed by batch normalization~\cite{Ioffe2015Batch} after each linear layer apart from the last.
For consistency with state-of-the-art floating point local descriptors~\cite{Lowe2004Distinctive,Mishchuk2017Working,Tian2019SOSNet}, the joint embedding is $128$-dimensional and $\ell_2$ normalized.
For binary descriptors~\cite{Calonder2010BRIEF}, we use sigmoid activation after the last linear layer and, at test time, we use a threshold of $0.5$ for binarization to allow efficient matching using bit-wise operations.
Similarly, if needed~\cite{Lowe2004Distinctive,Mishchuk2017Working,Tian2019SOSNet}, the output of the network is $\ell_2$ normalized.
Please refer to the supplementary material for additional details.

	\section{Experimental evaluation}
\label{sec:evaluation}

We evaluate our approach on the tasks of visual localization and mapping in the context of the scenarios introduced in Section~\ref{sec:scenarios}.
Please refer to the supplementary material for additional results on the HPatches descriptor benchmark~\cite{Balntas2017HPatches}, the Image Matching Workshop challenge~\cite{Jin2020Image}, and the InLoc Indoor Visual Localization dataset~\cite{Taira2018InLoc}.
We consider four well-established widely used descriptors -- two handcrafted descriptors, with BRIEF~\cite{Calonder2010BRIEF} as a binary and SIFT~\cite{Lowe2004Distinctive} as a gradient-based approach as well as two state-of-the-art learned descriptors, HardNet~\cite{Mishchuk2017Working} and SOSNet~\cite{Tian2019SOSNet}.
In our initial experiments, we found the pair network to perform on par with the encoder-decoder approach in terms of accuracy and runtime.
However, the encoder-decoder architecture provides us with a joint embedding and covers all three discussed scenarios.
Thus, all results presented in this section were obtained using the encoder-decoder approach.
Please refer to the supplementary material for a comparison of the two approaches.

\subsection{Visual localization}

\setlength{\tabcolsep}{2.0pt}
\begin{table}
	\footnotesize
	\centering
	\begin{tabular}{c c | c | l l l l }
		\toprule
		\multirow{3}{*}{\rotatebox[origin=c]{90}{Scenario}} & \multirowcell{3}{Database\\descriptor} & \multirowcell{3}{Query\\descriptor} & \multicolumn{4}{c}{\% localized queries} \\
		& & & \multicolumn{2}{c}{Day (824 images)} & \multicolumn{2}{c}{Night (98 images)} \\
		& & & $0.25m, 2^\circ$ & $0.5m, 5^\circ$ & $0.25m, 2^\circ$ & $0.5m, 5^\circ$ \\ \midrule\midrule
		\multirow{4}{*}{\rotatebox[origin=c]{90}{Standard}} & BRIEF & BRIEF & \textcolor{OrangeRed}{76.1} & \textcolor{OrangeRed}{81.4} & \textcolor{OrangeRed}{32.7} & \textcolor{OrangeRed}{36.7} \\ 
		& SIFT & SIFT & \textcolor{Cerulean}{82.5} & \textcolor{Cerulean}{88.7} & \textcolor{Cerulean}{52.0} & \textcolor{Cerulean}{61.2} \\
		& HardNet & HardNet & \textcolor{OliveGreen}{86.2} & \textcolor{OliveGreen}{92.2} & \textcolor{OliveGreen}{64.3} & \textcolor{OliveGreen}{72.4} \\
		& SOSNet & SOSNet & \textcolor{RoyalPurple}{86.4} & \textcolor{RoyalPurple}{92.7} & \textcolor{RoyalPurple}{65.3} & \textcolor{RoyalPurple}{75.5} \\
		\midrule\midrule
		\multirow{13.5}{*}{\rotatebox[origin=c]{90}{Continuous deployment}} & \multirowcell{3}{BRIEF $\rightarrow$} & SIFT & 74.9 \textsuperscript{\textcolor{OrangeRed}{-1.2}} & 80.5 \textsuperscript{\textcolor{OrangeRed}{-0.9}} & 31.6 \textsuperscript{\textcolor{OrangeRed}{-1.1}} & 36.7 \textsuperscript{\textcolor{OrangeRed}{0.0}} \\
		& & HardNet & 81.4 \textsuperscript{\textcolor{OrangeRed}{+5.3}} & 86.7 \textsuperscript{\textcolor{OrangeRed}{+5.3}} & 44.9 \textsuperscript{\textcolor{OrangeRed}{+12.2}} & 49.0 \textsuperscript{\textcolor{OrangeRed}{+12.3}} \\
		& & SOSNet & 81.6 \textsuperscript{\textcolor{OrangeRed}{+5.5}} & 86.9 \textsuperscript{\textcolor{OrangeRed}{+5.5}} & 42.9 \textsuperscript{\textcolor{OrangeRed}{+10.2}} & 46.9 \textsuperscript{\textcolor{OrangeRed}{+10.2}} \\ \cmidrule{2-7}
		& \multirowcell{3}{SIFT $\rightarrow$} & BRIEF & 66.6 \textsuperscript{\textcolor{Cerulean}{-15.9}} & 73.1 \textsuperscript{\textcolor{Cerulean}{-15.6}} & 19.4 \textsuperscript{\textcolor{Cerulean}{-32.6}} & 23.5 \textsuperscript{\textcolor{Cerulean}{-37.7}} \\
		& & HardNet & 83.4 \textsuperscript{\textcolor{Cerulean}{+0.9}} & 90.9 \textsuperscript{\textcolor{Cerulean}{+2.2}} & 59.2 \textsuperscript{\textcolor{Cerulean}{+7.2}} & 66.3 \textsuperscript{\textcolor{Cerulean}{+5.1}} \\
		& & SOSNet & 84.2 \textsuperscript{\textcolor{Cerulean}{+1.7}} & 91.4 \textsuperscript{\textcolor{Cerulean}{+2.7}} & 55.1 \textsuperscript{\textcolor{Cerulean}{+3.1}} & 62.2 \textsuperscript{\textcolor{Cerulean}{+1.0}} \\ \cmidrule{2-7}
		& \multirowcell{3}{HardNet $\rightarrow$} & BRIEF & 70.5 \textsuperscript{\textcolor{OliveGreen}{-15.7}} & 76.7 \textsuperscript{\textcolor{OliveGreen}{-15.5}} & 22.4 \textsuperscript{\textcolor{OliveGreen}{-41.9}} & 26.5 \textsuperscript{\textcolor{OliveGreen}{-45.9}} \\
		& & SIFT & 81.2 \textsuperscript{\textcolor{OliveGreen}{-5.0}} & 88.0 \textsuperscript{\textcolor{OliveGreen}{-4.2}} & 41.8 \textsuperscript{\textcolor{OliveGreen}{-22.5}} & 51.0 \textsuperscript{\textcolor{OliveGreen}{-21.4}} \\
		& & SOSNet & 85.8 \textsuperscript{\textcolor{OliveGreen}{-0.4}} & 92.4 \textsuperscript{\textcolor{OliveGreen}{+0.2}} & 61.2 \textsuperscript{\textcolor{OliveGreen}{-3.1}} & 68.4 \textsuperscript{\textcolor{OliveGreen}{-4.0}} \\ \cmidrule{2-7}
		& \multirowcell{3}{SOSNet $\rightarrow$} & BRIEF & 68.8 \textsuperscript{\textcolor{RoyalPurple}{-17.6}} & 74.8 \textsuperscript{\textcolor{RoyalPurple}{-17.9}} & 18.4 \textsuperscript{\textcolor{RoyalPurple}{-46.9}} & 20.4 \textsuperscript{\textcolor{RoyalPurple}{-55.1}} \\
		& & SIFT & 81.7 \textsuperscript{\textcolor{RoyalPurple}{-4.7}} & 87.5 \textsuperscript{\textcolor{RoyalPurple}{-5.2}} & 42.9 \textsuperscript{\textcolor{RoyalPurple}{-22.4}} & 49.0 \textsuperscript{\textcolor{RoyalPurple}{-26.5}} \\
		& & HardNet & 85.9 \textsuperscript{\textcolor{RoyalPurple}{-0.5}} & 92.4 \textsuperscript{\textcolor{RoyalPurple}{-0.3}} & 63.3 \textsuperscript{\textcolor{RoyalPurple}{-2.0}} & 69.4 \textsuperscript{\textcolor{RoyalPurple}{-6.1}} \\ \midrule\midrule
		\multirow{13.5}{*}{\rotatebox[origin=c]{90}{Cross-device}} & \multirowcell{3}{BRIEF $\leftarrow$} & SIFT & 65.8 \textsuperscript{\textcolor{OrangeRed}{-10.3}} & 71.6 \textsuperscript{\textcolor{OrangeRed}{-9.8}} & 22.4 \textsuperscript{\textcolor{OrangeRed}{-10.3}} & 24.5 \textsuperscript{\textcolor{OrangeRed}{-12.2}} \\
		& & HardNet & 68.8 \textsuperscript{\textcolor{OrangeRed}{-7.3}} & 74.0 \textsuperscript{\textcolor{OrangeRed}{-7.4}} & 20.4 \textsuperscript{\textcolor{OrangeRed}{-12.3}} & 26.5 \textsuperscript{\textcolor{OrangeRed}{-10.2}} \\
		& & SOSNet & 66.6 \textsuperscript{\textcolor{OrangeRed}{-9.5}} & 71.4 \textsuperscript{\textcolor{OrangeRed}{-10.0}} & 22.4 \textsuperscript{\textcolor{OrangeRed}{-10.3}} & 24.5 \textsuperscript{\textcolor{OrangeRed}{-12.2}} \\ \cmidrule{2-7}
		& \multirowcell{3}{SIFT $\leftarrow$} & BRIEF & 75.6 \textsuperscript{\textcolor{Cerulean}{-6.9}} & 80.8 \textsuperscript{\textcolor{Cerulean}{-7.9}} & 28.6 \textsuperscript{\textcolor{Cerulean}{-23.4}} & 36.7 \textsuperscript{\textcolor{Cerulean}{-24.5}} \\
		& & HardNet & 81.2 \textsuperscript{\textcolor{Cerulean}{-1.3}} & 87.3 \textsuperscript{\textcolor{Cerulean}{-1.4}} & 46.9 \textsuperscript{\textcolor{Cerulean}{-5.1}} & 55.1 \textsuperscript{\textcolor{Cerulean}{-6.1}} \\
		& & SOSNet & 80.1 \textsuperscript{\textcolor{Cerulean}{-2.4}} & 87.3 \textsuperscript{\textcolor{Cerulean}{-1.4}} & 42.9 \textsuperscript{\textcolor{Cerulean}{-9.1}} & 46.9 \textsuperscript{\textcolor{Cerulean}{-14.3}} \\ \cmidrule{2-7}
		& \multirowcell{3}{HardNet $\leftarrow$} & BRIEF & 82.8 \textsuperscript{\textcolor{OliveGreen}{-3.4}} & 88.8 \textsuperscript{\textcolor{OliveGreen}{-3.4}} & 43.9 \textsuperscript{\textcolor{OliveGreen}{-20.4}} & 49.0 \textsuperscript{\textcolor{OliveGreen}{-23.4}} \\
		& & SIFT & 84.7 \textsuperscript{\textcolor{OliveGreen}{-1.5}} & 91.0 \textsuperscript{\textcolor{OliveGreen}{-1.2}} & 58.2 \textsuperscript{\textcolor{OliveGreen}{-6.1}} & 67.3 \textsuperscript{\textcolor{OliveGreen}{-5.1}} \\
		& & SOSNet & 86.2 \textsuperscript{\textcolor{OliveGreen}{0.0}} & 92.7 \textsuperscript{\textcolor{OliveGreen}{+0.5}} & 64.3 \textsuperscript{\textcolor{OliveGreen}{0.0}} & 69.4 \textsuperscript{\textcolor{OliveGreen}{-3.0}} \\ \cmidrule{2-7}
		& \multirowcell{3}{SOSNet $\leftarrow$} & BRIEF & 82.5 \textsuperscript{\textcolor{RoyalPurple}{-3.9}} & 88.2 \textsuperscript{\textcolor{RoyalPurple}{-4.5}} & 44.9 \textsuperscript{\textcolor{RoyalPurple}{-20.4}} & 49.0 \textsuperscript{\textcolor{RoyalPurple}{-26.5}} \\ 
		& & SIFT & 84.0 \textsuperscript{\textcolor{RoyalPurple}{-2.4}} & 91.1 \textsuperscript{\textcolor{RoyalPurple}{-1.6}} & 51.0 \textsuperscript{\textcolor{RoyalPurple}{-14.3}} & 57.1 \textsuperscript{\textcolor{RoyalPurple}{-18.0}} \\
		& & HardNet & 85.3 \textsuperscript{\textcolor{RoyalPurple}{-1.1}} & 91.9 \textsuperscript{\textcolor{RoyalPurple}{-0.8}} & 66.3 \textsuperscript{\textcolor{RoyalPurple}{+1.0}} & 72.4 \textsuperscript{\textcolor{RoyalPurple}{-3.1}} \\ \midrule\midrule
		\multirow{4}{*}{\rotatebox[origin=c]{90}{Collab.}} & \multirowcell{4}{Embed  $\leftarrow$} & BRIEF & 80.6 \textsuperscript{\textcolor{OrangeRed}{+4.5}} & 86.7 \textsuperscript{\textcolor{OrangeRed}{+5.3}} & 48.0 \textsuperscript{\textcolor{OrangeRed}{+15.3}} & 50.0 \textsuperscript{\textcolor{OrangeRed}{+13.3}} \\
		& & SIFT & 82.8 \textsuperscript{\textcolor{Cerulean}{+0.3}} & 89.0 \textsuperscript{\textcolor{Cerulean}{+0.3}} & 50.0 \textsuperscript{\textcolor{Cerulean}{-2.0}} & 57.1 \textsuperscript{\textcolor{Cerulean}{-4.1}} \\
		& & HardNet & 85.1 \textsuperscript{\textcolor{OliveGreen}{-1.1}} & 91.7 \textsuperscript{\textcolor{OliveGreen}{-0.5}} & 55.1 \textsuperscript{\textcolor{OliveGreen}{-9.2}} & 61.2 \textsuperscript{\textcolor{OliveGreen}{-11.2}} \\
		& & SOSNet & 84.8 \textsuperscript{\textcolor{RoyalPurple}{-1.6}} & 90.9 \textsuperscript{\textcolor{RoyalPurple}{-1.8}} & 57.1 \textsuperscript{\textcolor{RoyalPurple}{-8.2}} & 60.2 \textsuperscript{\textcolor{RoyalPurple}{-15.3}} \\ \bottomrule
	\end{tabular}
	\vspace{-7.5pt}
	\caption{{\bf Localization under continuous deployment.} A reference map is built using the database description algorithm. The descriptors of this map are translated to a target query descriptor. {\bf Cross-device localization.} A reference map is built using the database description algorithm. The descriptors of query images are translated to be compatible with the map. {\bf Localization to collaborative maps.} The database images are partitioned in $4$ balanced subsets, one for each description algorithm. Both database and query descriptors are mapped to the common embedding space. The absolute performance difference \textsuperscript{(superscript)} is color-coded according to the baseline used as reference.}
	\label{tab:aachen}
	\vspace{-10pt}
\end{table}
\setlength{\tabcolsep}{\tabcolsepdefault}

We consider the task of visual localization against pre-built maps on the challenging Aachen Day \& Night localization benchmark~\cite{Sattler2017Benchmarking}.
We start by matching each database image with its $20$ nearest spatial neighbors.
COLMAP~\cite{Schoenberger2016Structure} is used for triangulation using the provided camera poses and intrinsics.
Next, each query image is matched against its top $50$ retrieval results according to NetVLAD~\cite{Arandjelovic2016NetVLAD}.
Finally, we use COLMAP's image registrator for localization with known intrinsics.
According to standard procedure, the poses are submitted to the evaluation system~\cite{VisualLocalization} and we report the percentage of localized images at different real-world thresholds in Table~\ref{tab:aachen}.

\customparagraph{Localization under continuous deployment.}
To evaluate this scenario, we build the reference map from a single description algorithm (\eg, SIFT).
Then, all the feature descriptors of the map are translated to a target feature representation (\eg, SIFT $\rightarrow$ HardNet).
For the queries, we use descriptors extracted directly by the target feature description algorithm (\eg, HardNet) and match them against the translated features.
The resulting matches are used for localization.
Our approach not only enables updating descriptors without re-mapping the environment, but in many cases also unexpectedly improves the re-localization performance. 
In particular, hand-crafted descriptors~\cite{Calonder2010BRIEF,Lowe2004Distinctive} are significantly better after translating them to learned ones~\cite{Mishchuk2017Working,Tian2019SOSNet}.
For SIFT, the increase is of more than $2\%$ and $5\%$ in absolute performance at the largest threshold for day and night queries, respectively, and for BRIEF the difference is even more accentuated with $5\%$ and $10\%$.
While there is still a gap w.r.t. re-building the map from scratch using the learned descriptors, the performance can be further improved using a bootstrapping strategy (\cf Section~\ref{sec:scenarios-continuous}).

\customparagraph{Cross-device localization.}
Inverse to the previous experiment, we now build maps using a description algorithm (\eg, SIFT) and instead translate the query descriptors (\eg, HardNet $\rightarrow$ SIFT).
Similarly to the previous case, compared to standard localization, we notice increased performance when translating hand-crafted descriptors to learned ones.
Maps built from state-of-the-art descriptors (HardNet, SOSNet) preserve their advantages no matter the description algorithm used for queries.
Not surprisingly, translating other descriptors to BRIEF~\cite{Calonder2010BRIEF} yields worse performance due to the more limited discriminative power of binary descriptors.
Overall, our approach enables localization with a good performance in this previously infeasible scenario.

\customparagraph{Localization to collaborative maps.}
For the final scenario, we randomly split the database images of the map into $4$ balanced subsets (one for each description algorithm) and translate all descriptors to the joint embedding space (\ie, SIFT, HardNet, \ldots $\rightarrow$ embedding).
Similarly, all query descriptors are translated to the joint embedding space for matching.
Although the map is built from heterogeneous descriptors, the localization performance at day is comparable to the best results in the previous experiments.
Moreover, note that the performance of HardNet and SOSNet in this scenario is just slightly below the current state-of-the-art patch-based descriptor~\cite{Tian2019SOSNet}.
In addition, BRIEF queries achieve a substantial improvement of more than $5\%$ for both day and night folds at the coarsest threshold.
This is a very encouraging result for cloud-based solutions, as we enable multiple devices implementing different algorithms to not only collaborate on mapping a location, but also remain competitive with the state-of-the-art in localization.

\subsection{Collaborative mapping}

\setlength{\tabcolsep}{2.0pt}
\begin{table*}
	\centering
	\footnotesize
	\begin{tabular}{c c | c c c c c c | c c c c c c | c c c c c c}
		\toprule
		\multicolumn{2}{c | }{Dataset} & \multicolumn{6}{c |}{\textit{Madrid Metropolis} -- $453$ images} & \multicolumn{6}{c |}{\textit{Gendarmenmarkt} -- $985$ images} & \multicolumn{6}{c}{\textit{Tower of London} -- $730$ images} \\ 
		\midrule
		\multicolumn{2}{c |}{\multirow{3}{*}{Method}} & \multicolumn{3}{c}{\% localized images} & \multirowcell{3}{Num.\\3D pts.} & \multirowcell{3}{Track\\length} & \multirowcell{3}{Reproj.\\error} & \multicolumn{3}{c}{\% localized images} & \multirowcell{3}{Num.\\3D pts.} & \multirowcell{3}{Track\\length} & \multirowcell{3}{Reproj.\\error} & \multicolumn{3}{c}{\% localized images} & \multirowcell{3}{Num.\\3D pts.} & \multirowcell{3}{Track\\length} & \multirowcell{3}{Reproj.\\error} \\
		& & $0.25m$ & $0.5m$ & \multirowcell{2}{$\infty$} & & & & $0.25m$ & $0.5m$ & \multirowcell{2}{$\infty$} & & & & $0.25m$ & $0.5m$ & \multirowcell{2}{$\infty$} & & & \\
		& & $2^\circ$ & $5^\circ$ & & & & & $2^\circ$ & $5^\circ$ & & & & & $2^\circ$ & $5^\circ$ & & & & \\ \midrule
		\multirow{4}{*}{\rotatebox[origin=c]{90}{Standard}} & BRIEF & 57.0 & 64.2 & 72.4 & 18.3K & 6.84 & 0.63 & 59.9 & 68.7 & 80.7 & 52.3K & 6.31 & 0.82 & 64.9 & 68.5 & 74.2 & 48.1K & 7.70 & 0.66 \\
		& SIFT & 78.1 & 83.7 & 95.1 & 39.4K & 6.71 & 0.83 & 68.8 & 77.1 & 95.6 & 121.4K & 5.53 & 0.95 & 74.2 & 76.7 & 97.1 & 90.0K & 7.14 & 0.81 \\
		& HardNet & 89.2 & 95.4 & 100 & 47.4K & 7.15 & 0.92 & 84.2 & 90.7 & 99.8 & 135.4K & 6.40 & 1.01 & 83.0 & 87.7 & 100 & 104.5K & 7.56 & 0.87 \\
		& SOSNet & 92.7 & 96.3 & 100 & 46.0K & 7.22 & 0.92 & 85.5 & 89.2 & 99.9 & 128.4K & 6.56 & 1.02 & 85.2 & 89.2 & 100 & 101.3K & 7.67 & 0.86 \\ \midrule
		\multirow{4}{*}{\rotatebox[origin=c]{90}{Real-world}} & BRIEF & 3.1 & 4.9 & 6.0 & 1.9K & 4.79 & 0.53 & 2.9 & 3.7 & 10.4 & 4.5K & 4.36 & 0.72 & 10.4 & 11.5 & 11.8 & 5.8K & 4.65 & 0.57 \\
		& SIFT & 14.1 & 16.6 & 21.6 & 6.1K & 4.70 & 0.77 & 5.4 & 6.8 & 20.9 & 15.7K & 4.29 & 0.89 & 13.0 & 16.0 & 17.1 & 15.2K & 4.55 & 0.74 \\
		& HardNet & 10.8 & 13.9 & 21.4 & 8.8K & 5.34 & 0.87 & 13.8 & 15.9 & 23.2 & 25.7K & 4.67 & 0.95 & 16.0 & 17.8 & 18.6 & 22.1K & 5.23 & 0.80 \\
		& SOSNet & 16.8 & 19.4 & 21.2 & 7.5K & 4.70 & 0.85 & 13.6 & 14.8 & 23.5 & 26.7K & 4.95 & 0.96 & 17.3 & 18.8 & 23.0 & 22.9K & 5.32 & 0.80 \\ \midrule
		\multirow{2.5}{*}{\rotatebox[origin=c]{90}{Ours}} & Embed & 80.6 & 84.3 & 92.1 & 36.3K & 7.26 & 0.86 & 74.1 & 82.3 & 95.6 & 103.4K & 6.45 & 0.98 & 77.3 & 81.2 & 97.9 & 88.5K & 7.57 & 0.81 \\ \cmidrule{2-20}
		& Progressive & 77.0 & 82.8 & 88.7 & 31.2K & 7.59 & 0.86 & 76.2 & 82.7 & 94.8 & 92.2K & 6.58 & 0.98 & 79.2 & 83.2 & 96.6 & 76.0K & 7.81 & 0.82 \\ \bottomrule
	\end{tabular}
	\vspace{-7.5pt}
	\caption{{\bf Collaborative mapping.} We report different reconstruction statistics from internet collected images. The first four rows represent the standard evaluation protocol where each description algorithm has access to all images. The next four rows present more realistic scenarios where each algorithm only has access to a quarter of the data. Finally, the last two rows represent variants of our method enabling cross-descriptor reconstruction using the same splits as above.}
	\label{tab:sfm}
	\vspace{-10pt}
\end{table*}
\setlength{\tabcolsep}{\tabcolsepdefault}

Next, we consider the large-scale Structure-from-Motion benchmark of Sch\"{o}nberger~\etal~\cite{Schonberger2017Comparative}.
Similar to other datasets~\cite{Sattler2017Benchmarking, Jin2020Image}, to evaluate the camera poses, we generate pseudo-ground-truth intrinsics and extrinsics via an initial SfM process.
Only the images registered in this step are considered.\footnote{Please refer to the supplementary material for additional details.}
For each method, we exhaustively match all remaining images using a mutual nearest-neighbors matcher with symmetric second nearest neighbor ratio test~\cite{Lowe2004Distinctive} (the threshold is set to $0.9$ for all methods).
Given the feature matches, COLMAP~\cite{Schoenberger2016Structure} is used for geometric verification and sparse reconstruction with known intrinsics.
In the context of collaborative mapping (\cf Section~\ref{sec:crowdmapping}), we randomly split the database into balanced sets corresponding to different description algorithms.

The resulting point-cloud statistics are reported in Table~\ref{tab:sfm}.
We evaluate two variants of our method: i) \textbf{Embed} taking advantage of the joint embedding space and ii) \textbf{Progressive} using the hierarchy BRIEF $\rightarrow$ SIFT $\rightarrow$ HardNet $\rightarrow$ SOSNet based on the rankings reported in~\cite{Balntas2017HPatches,Tian2019SOSNet}.
In this variant, the translation direction is chosen \emph{online} towards the strongest in each pair.
Given two images with descriptors X and Y respectively, we translate the lower one to the higher one in the order and then run matching on the obtained descriptors.
For instance, in an image pair with BRIEF and HardNet, BRIEF would be translated to HardNet, while in an image pair with only SIFT descriptors, there would be no translation.

First, we compare the joint embedding approach with progressive translation.
Both models achieve comparable results, but the jointly embedded map consistently registers more images and reconstructs more 3D points, albeit at the cost of shorter track lengths.
Furthermore, the joint embedding also significantly simplifies the translation and matching in the reconstruction pipeline.

Next, we compare our approach with the traditional setup, where one has access to the same feature descriptor for all dataset images.   
While these baselines are not able to tackle collaborative mapping with heterogeneous descriptors, it serves as an upper-bound for our method.
Both our approaches outperform maps built from only BRIEF or SIFT, but achieve slightly inferior results than state-of-the-art HardNet and SOSNet.
However, it is important to keep in mind that, in our case, the dataset is split into 4 random subsets.
Therefore, only half of the images have HardNet or SOSNet features.   
Finally, we compare against baselines that only have access to their associated subset of dataset images to demonstrate the actual merit of our work.
This corresponds to the realistic setup in which 4 types of devices with incompatible descriptors map the same environment but the mapping system cannot jointly leverage them.
In this case, both our approaches enable collaboration across devices and significantly outperform all baselines -- we consistently register four times more images.

\begin{figure}
	\begin{minipage}{0.47\columnwidth}
		\centering
		\includegraphics[width=\textwidth]{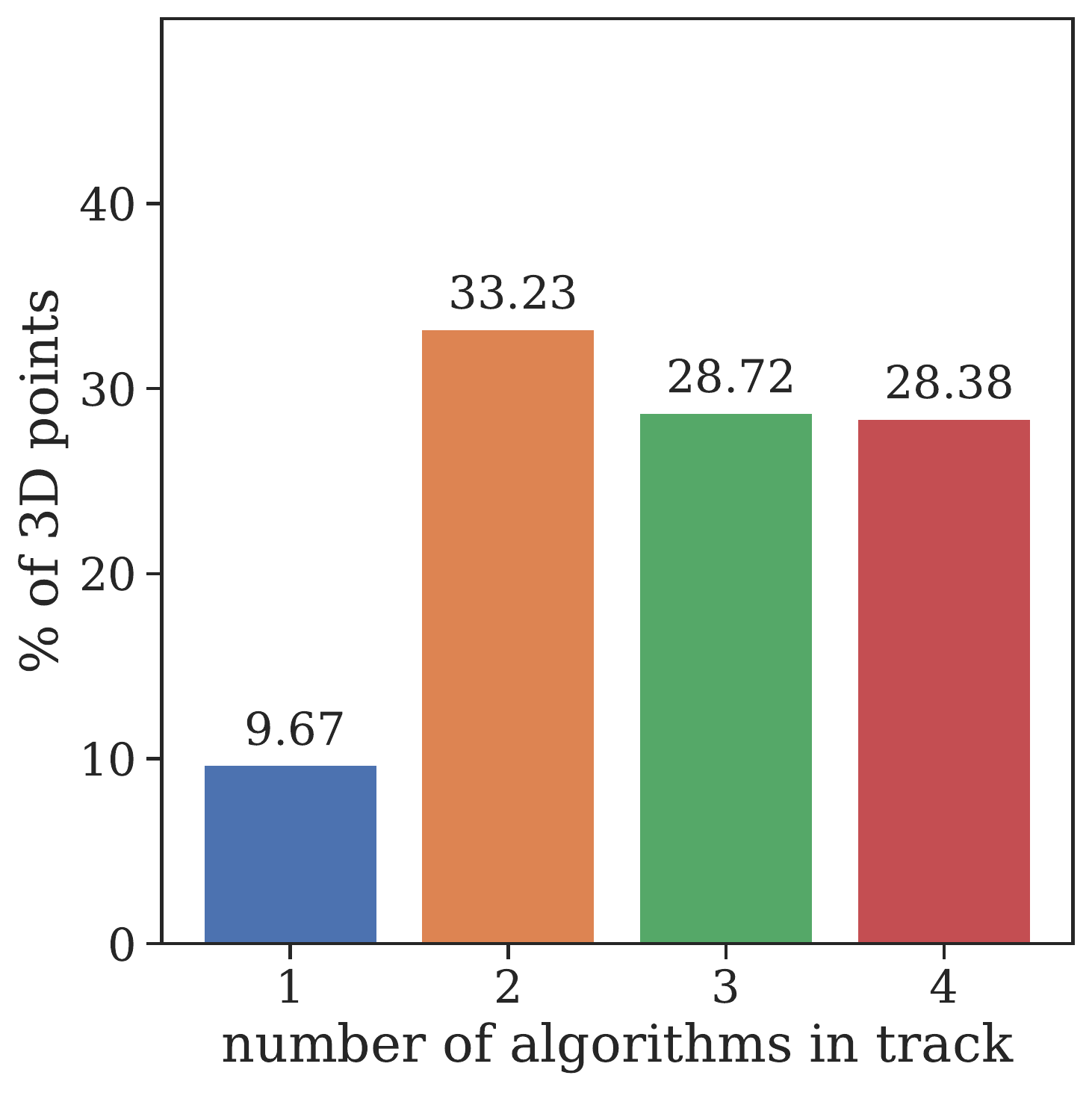}
	\end{minipage}
	\hfill
	\begin{minipage}{0.51\columnwidth}
		\centering
		\includegraphics[width=\textwidth]{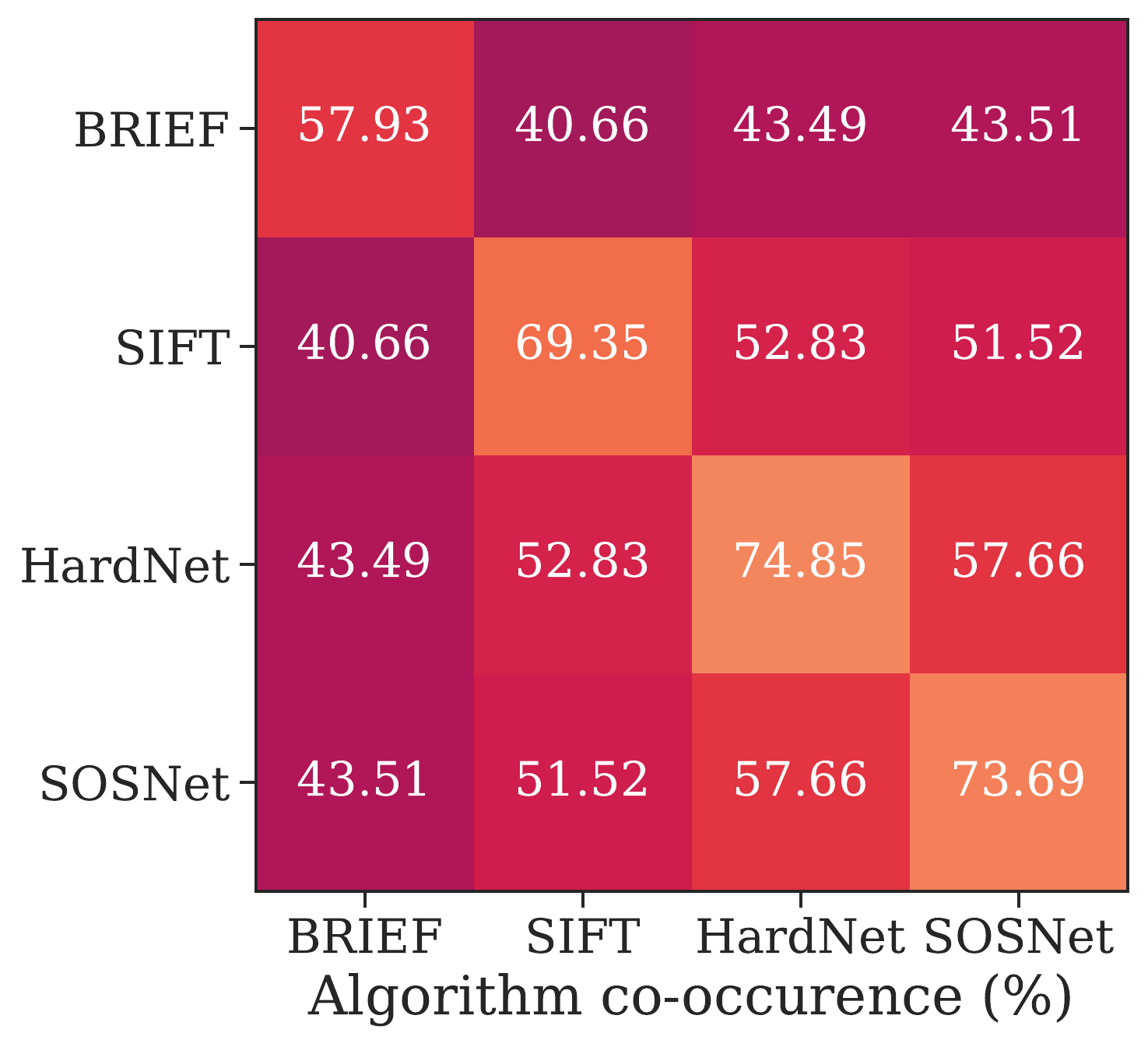}
	\end{minipage}
	\vspace{-7.5pt}
	\caption{{\bf Co-visibility statistics.} For the \textbf{Embed} approach on the \textit{Tower of London} dataset, we report the \% of 3D points containing $1-4$ distinct algorithms in their tracks on the left. On the right, we visualize the co-occurence, \ie, the percentage of 3D points containing descriptors originating from $2$ given description algorithms in their tracks.}
	\label{fig:sfm-stats}
	\vspace{-10pt}
\end{figure}

Figure~\ref{fig:sfm-stats} provides co-visibility statistics of the 3D points for the \textbf{Embed} model.
An overwhelming majority of the 3D points have at least $2$ different description algorithms in their tracks and more than $50\%$ have $3$ or $4$.
Furthermore, the co-occurence matrix demonstrates that all descriptors contributed (almost) equally to the map.
We refer to the supplementary material for statistics on the other datasets.

	\section{Discussion}
\label{sec:limitations}

The main limitation comes from the \textbf{same keypoint} assumption, as our work only considers the compatibility of different description algorithms.
While our assumption is likely applicable for devices produced by a single manufacturer that has control over the entire pipeline, different manufacturers likely use different keypoint detection algorithms.
Possible directions for future research include studying cross-detector repeatability or improving feature inversion techniques for better generalization with different local feature detection pipelines.
Despite our proposed solution being only a first step, it still has significant real-world applications: i) migration of existing maps from, \eg, handcrafted descriptors to learned ones and, similarly, continuous updates of learned descriptors in case of novel architectures / losses / datasets, ii) support of legacy devices inside maps built with newer descriptors, and iii) collaborative mapping between devices of the same manufacturer with different compute capabilities.
Furthermore, while we have made a significant step towards enabling cross-descriptor localization and mapping, several challenges still remain.
\textbf{Information loss.}
Each description algorithm only encodes a subset of the information available in the input patches and these subsets might be distinct for different methods.
This is further exacerbated in efficient real-world applications taking advantage of quantization and dimensionality reduction.
\textbf{One-to-many associations.}
A one-to-one mapping may not even exist as two given patches might lead to a similar descriptor for one algorithm but completely different feature vectors for another one.
Similarly to other data-driven approaches, translation performance has to be verified experimentally as there are no formal guarantees.
To address these limitations, future research could try to take advantage of local or global context to better disambiguate the visual information.

	\section{Conclusion}

We have identified, introduced, and formalized three novel scenarios for localization and mapping in the presence of heterogeneous feature representations.
Towards addressing the open challenges underlying these scenarios, we presented the first principled solution to cross-descriptor localization and mapping.
We demonstrated the effectiveness of our approach on representative and well-established benchmarks.
We believe our work will not only spark new research on the topic of translating local image features but will also make an immediate impact on commercial applications in the area of cloud-based localization and mapping.

{\small \noindent {\bf Acknowledgments.} This work was supported by the Microsoft MR \& AI Lab Z\"urich PhD scholarship.}
	
	\appendix
	\part*{Supplementary Material}
	This supplementary material provides the following information:
first, we report additional experimental results on the Image Matching Workshop challenge~\cite{Jin2020Image}, the HPatches benchmark~\cite{Balntas2017HPatches}, the InLoc Indoor Visual Localization dataset~\cite{Taira2018InLoc}, and the Aachen Day-Night dataset~\cite{Sattler2017Benchmarking}.
Second, we provide further implementation details.
Third, we explain the protocol for generating pseudo-ground-truth intrinsics and extrinsics for the Local Feature Evaluation benchmark~\cite{Schonberger2017Comparative}.
Fourth, we show co-visibility statistics for the collaborative mapping experiments.
Finally, we present an ablation study comparing different architectures and loss formulations.

	\section{Additional experimental results}

In this section, we report additional experimental results.
First, we evaluate our method on the Image Matching Workshop (IMW) challenge~\cite{Jin2020Image} as well as the HPatches descriptor evaluation benchmark~\cite{Balntas2017HPatches}.
Next, we provide a per-scene breakdown on the full sequences of the HPatches dataset~\cite{Balntas2017HPatches}.
Then, we study the impact of the joint embedding dimension in the scenario of localization to a collaborative map on the Aachen Day-Night dataset~\cite{Sattler2017Benchmarking}.

\subsection{Image Matching Workshop challenge}

We evaluate the performance of descriptor translation on the stereo and multi-view tasks of the IMW challenge~\cite{Jin2020Image}.
Given the large number of methods to consider\footnote{The benchmark rules limit each team to a maximum of 2 submissions per week to avoid parameter tuning on the test set.}, we restrict the evaluation to the $3$ validation scenes as follows: the smallest one (Reichstag) is used for parameter tuning (thresholds for the ratio test and RANSAC), while the other two (Sacre Coeur and Saint Peter's Square) are used for evaluation.
We use the $2048$ OpenCV SIFT keypoints with default parameters provided by the authors.
For consistency, we retrain the encoder-decoder approach on patches extracted according to OpenCV SIFT keypoints on the same 3190 random internet images part of the Oxford-Paris revisited retrieval dataset distractors~\cite{Radenovic2018Revisiting}.

\setlength{\tabcolsep}{2.0pt}
\begin{table}[t]
	\footnotesize
	\centering
	\begin{tabular}{c c | l l | l l c }
		\toprule
		\multirow{3}{*}{\rotatebox[origin=c]{90}{}} & \multirowcell{3}{Descriptor} & \multicolumn{2}{c |}{Stereo} & \multicolumn{3}{c}{Multi-view} \\
		& & \multicolumn{2}{c |}{AUC (\%)} & \multicolumn{2}{c}{AUC (\%)} & \multirow{2}{*}{\rotatebox[origin=c]{90}{Real.}} \\
		& & \multicolumn{1}{c}{$5^\circ$} & \multicolumn{1}{c |}{$10^\circ$} & \multicolumn{1}{c}{$5^\circ$} & \multicolumn{1}{c}{$10^\circ$} & \\ \midrule\midrule
		\multirow{4}{*}{\rotatebox[origin=c]{90}{Standard}} & BRIEF & \textcolor{OrangeRed}{35.3} & \textcolor{OrangeRed}{41.8} & \textcolor{OrangeRed}{31.9} & \textcolor{OrangeRed}{36.5} & \cmark \\ 
		& SIFT & \textcolor{Cerulean}{41.4} & \textcolor{Cerulean}{49.2} & \textcolor{Cerulean}{41.4} & \textcolor{Cerulean}{48.7} & \cmark\\
		& HardNet & \textcolor{OliveGreen}{51.4} & \textcolor{OliveGreen}{59.9} & \textcolor{OliveGreen}{55.9} & \textcolor{OliveGreen}{63.5} & \cmark \\
		& SOSNet & \textcolor{RoyalPurple}{51.4} & \textcolor{RoyalPurple}{60.1} & \textcolor{RoyalPurple}{58.6} & \textcolor{RoyalPurple}{66.2} & \cmark\\
		\midrule\midrule
		\multirow{7}{*}{\rotatebox[origin=c]{90}{Directional}} & BRIEF $\rightarrow$ SIFT & 25.2 & 31.5 & 14.9 & 17.6 & \xmark \\
		& BRIEF $\rightarrow$ HardNet & 35.3 & 42.7 & 36.5 & 40.8 & \xmark \\
		& BRIEF $\rightarrow$ SOSNet & 39.8 & 47.5 & 40.3 & 46.9 & \xmark \\ \cmidrule{2-7}
		& SIFT $\rightarrow$ HardNet & 42.7 & 51.1 & 48.7 & 55.4 & \xmark \\
		& SIFT $\rightarrow$ SOSNet & 45.1 & 53.5 & 47.3 & 55.2 & \xmark \\ \cmidrule{2-7}
		& HardNet $\rightarrow$ SOSNet & 49.4 & 57.8 & 56.9 & 64.3 & \xmark \\ \midrule\midrule
		\multirow{7.5}{*}{\rotatebox[origin=c]{90}{Embed}} & BRIEF, SIFT, \nicefrac{1}{2} & 39.5 & 47.1 & 41.6 & 48.1 & \cmark \\
		& BRIEF, HardNet, \nicefrac{1}{2} & 42.4 & 50.3 & 46.2 & 52.8 & \cmark \\
		& BRIEF, SOSNet, \nicefrac{1}{2} & 41.3 & 48.9 & 45.2 & 51.9 & \cmark \\
		& SIFT, HardNet, \nicefrac{1}{2} & 46.8 & 55.1 & 53.4 & 61.3 & \cmark \\
		& SIFT, SOSNet, \nicefrac{1}{2} & 46.2 & 54.4 & 49.9 & 57.6 & \cmark \\
		& HardNet, SOSNet, \nicefrac{1}{2} & 50.4 & 58.9 & 57.6 & 64.9 & \cmark \\ \cmidrule{2-7}
		& All, \nicefrac{1}{4} & 42.3 & 50.1 & 46.7 & 53.5 & \cmark \\
		\bottomrule
	\end{tabular}
	\vspace{-7.5pt}
	\caption{{\bf Image Matching Workshop challenge.} We report results on the IMW challenge under two evaluation protocols: directional translation and collaborative mapping using the joint embedding.}
	\label{tab:imw}
\end{table}
\setlength{\tabcolsep}{\tabcolsepdefault}

We report results under two evaluation protocols in Table~\ref{tab:imw}.
First, we consider the case of directional translation.
For a given direction ($A \rightarrow B$), in each image pair, we use the target description algorithm ($B$) in the first image and we translate source descriptors to target ones in the second image.
Note that this does not correspond to a realistic scenario on the multi-view task, as the same image might use different descriptors in different image pairs.
Second, we consider the case of collaborative mapping using the joint embedding.
To this end, we randomly split the images of each dataset into balanced subsets, one for each description algorithm.
Following the original evaluation protocol, we run each method three times and report the average over all runs.
Once again, we notice an increase in performance when translating handcrafted to learned descriptors and matching them against natively extracted ones.
Further, in the binary collaborative scenario, the results are generally in between the results of the individual descriptors.

\subsection{HPatches benchmark}

\begin{figure*}
	\begin{subfigure}[t]{0.33\textwidth}
		\centering
		\includegraphics[width=\textwidth]{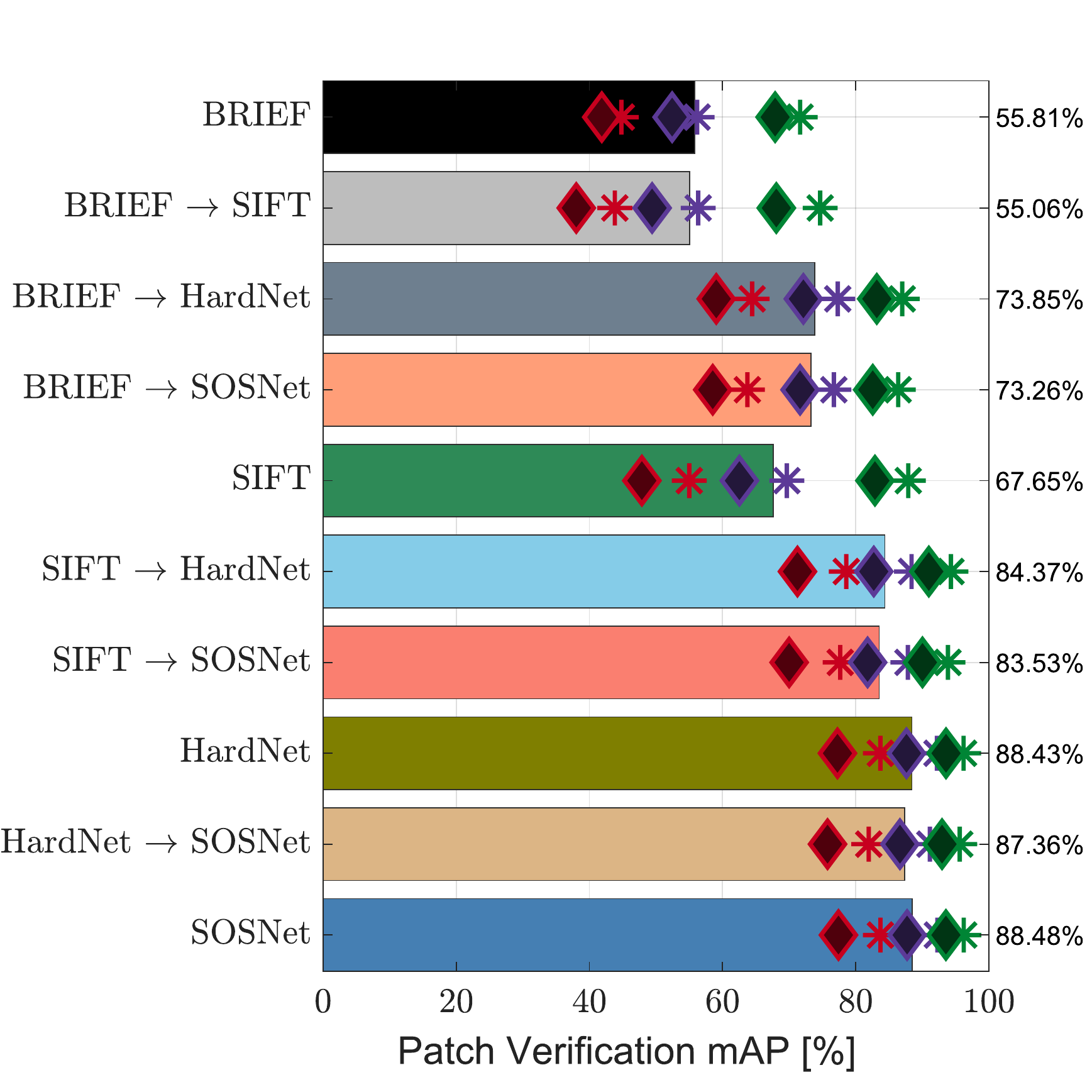}
	\end{subfigure}
	\hfill
	\begin{subfigure}[t]{0.33\textwidth}
		\centering
		\includegraphics[width=\textwidth]{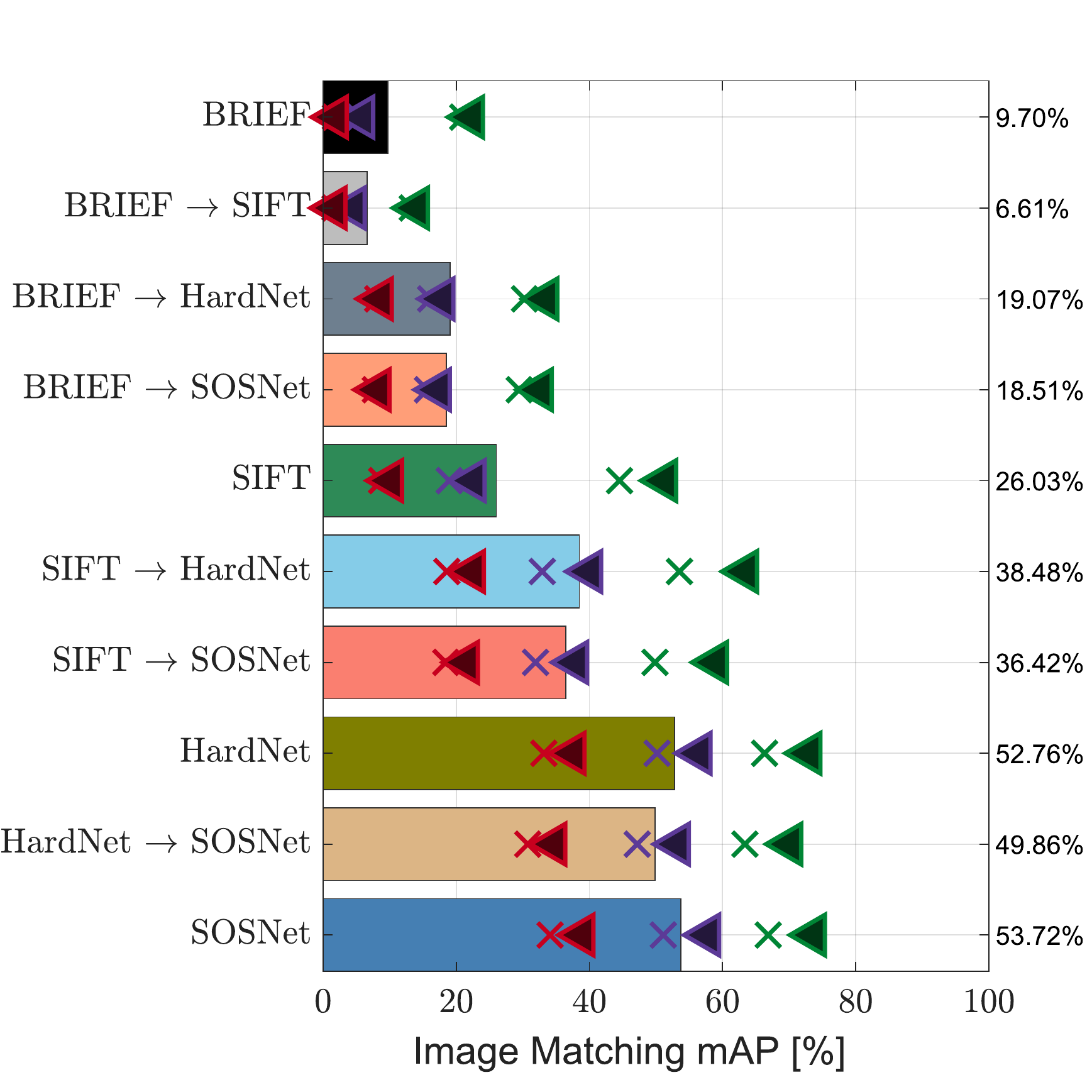}
	\end{subfigure}
	\hfill
	\begin{subfigure}[t]{0.33\textwidth}
		\centering
		\includegraphics[width=\textwidth]{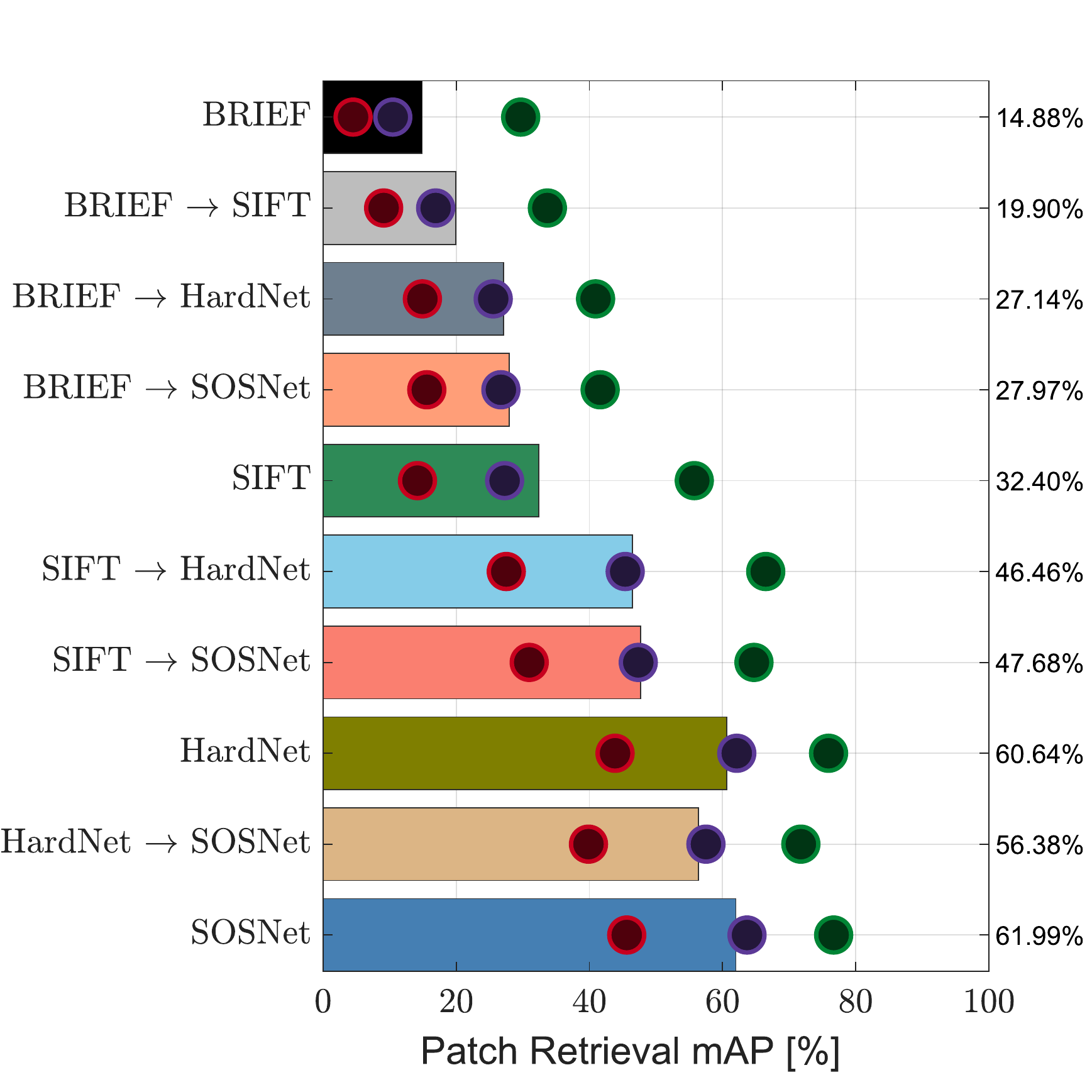}
	\end{subfigure}
	\caption{{\bf HPatches.} We report verification, matching, and retrieval results on the HPatches dataset. Color of the marker indicates \textcolor{Green}{Easy}, \textcolor{purple}{Hard}, and \textcolor{red}{Tough} noise. For the patch verification task, diamonds and stars show results with negatives from the same sequence and from different sequences, respectively. For the image matching task, crosses and triangles denote illumination and viewpoint results, respectively.}
	\label{fig:hpatches}
\end{figure*}

To analyze the raw matching performance between original and translated descriptors, we evaluate our method on the HPatches benchmark~~\cite{Balntas2017HPatches}.
There are three different tasks, notably patch verification, image matching and patch retrieval.
For the translated methods (denoted $A \rightarrow B$), we use the target description algorithm ($B$) directly in the reference patches and we translate source descriptors to target ones for all other patches.
Results are reported in Figure~\ref{fig:hpatches}.
As in our previous experiment, we notice an improvement in performance when translating handcrafted to learned descriptor.
Furthermore, while some translated descriptors achieve worse performance than the baselines (\eg, HardNet $\rightarrow$ SOSNet), all three tasks are possible in this previously unfeasible cross-descriptor scenario.

\setlength{\tabcolsep}{2.0pt}
\begin{table}
	\footnotesize
	\centering
	\begin{tabular}{c c | c | l l l l}
		\toprule
		\multirow{3}{*}{\rotatebox[origin=c]{90}{Scenario}} & \multirowcell{3}{Database\\descriptor} & \multirowcell{3}{Query\\descriptor} & \multicolumn{4}{c}{\% localized queries} \\
		& & & \multicolumn{2}{c}{DUC1} & \multicolumn{2}{c}{DUC2} \\
		& & & $0.25m$ & $0.5m$ & $0.25m$ & $0.5m$ \\ \midrule\midrule
		\multirow{4}{*}{\rotatebox[origin=c]{90}{Standard}} & BRIEF & BRIEF & \textcolor{OrangeRed}{25.3} & \textcolor{OrangeRed}{36.4} & \textcolor{OrangeRed}{22.9} & \textcolor{OrangeRed}{41.2} \\ 
		& SIFT & SIFT & \textcolor{Cerulean}{32.3} & \textcolor{Cerulean}{47.5} & \textcolor{Cerulean}{27.5} & \textcolor{Cerulean}{45.0} \\
		& HardNet & HardNet & \textcolor{OliveGreen}{36.4} & \textcolor{OliveGreen}{52.5} & \textcolor{OliveGreen}{30.5} & \textcolor{OliveGreen}{54.2} \\
		& SOSNet & SOSNet & \textcolor{RoyalPurple}{34.8} & \textcolor{RoyalPurple}{50.5} & \textcolor{RoyalPurple}{30.5} & \textcolor{RoyalPurple}{53.4} \\
		\midrule\midrule
		\multirow{7}{*}{\rotatebox[origin=c]{90}{Cont. deployment}} & \multirowcell{3}{BRIEF $\rightarrow$} & SIFT & 28.3 & 39.9 & 22.1 & 40.5 \\
		& & HardNet & 29.8 & 43.9 & 30.5 & 40.5 \\
		& & SOSNet & 31.8 & 43.4 & 23.7 & 40.5 \\ \cmidrule{2-7}
		& \multirowcell{2}{SIFT $\rightarrow$} & HardNet & 36.4 & 50.0 & 31.3 & 50.4 \\
		& & SOSNet & 36.4 & 53.5 & 33.6 & 50.4 \\ \cmidrule{2-7}
		& \multirowcell{1}{HardNet $\rightarrow$} & SOSNet & 33.3 & 48.5 & 30.5 & 55.7 \\ \midrule\midrule
		\multirow{7}{*}{\rotatebox[origin=c]{90}{Cross-device}} & \multirowcell{1}{SIFT $\leftarrow$} & BRIEF & 29.3 & 40.9 & 25.2 & 42.0 \\ \cmidrule{2-7}
		& \multirowcell{2}{HardNet $\leftarrow$} & BRIEF & 30.3 & 46.5 & 27.5 & 48.1 \\
		& & SIFT & 36.4 & 51.0 & 33.6 & 55.7 \\ \cmidrule{2-7}
		& \multirowcell{3}{SOSNet $\leftarrow$} & BRIEF & 29.8 & 44.9 & 29.0 & 45.0 \\
		& & SIFT & 34.8 & 51.0 & 33.6 & 53.4 \\
		& & HardNet & 37.4 & 50.5 & 29.0 & 49.6 \\ \bottomrule
	\end{tabular}
	\vspace{-7.5pt}
	\caption{{\bf InLoc Indoor Visual Localization.} {\bf Localization under continuous deployment.} A reference map is built using the database description algorithm. The descriptors of this map are translated to a target query descriptor. {\bf Cross-device localization.} A reference map is built using the database description algorithm. The descriptors of query images are translated to be compatible with the map.}
	\label{tab:inloc}
\end{table}
\setlength{\tabcolsep}{\tabcolsepdefault}

\subsection{InLoc Indoor Visual Localization dataset}

We also evaluate descriptor translation on the challenging InLoc Indoor Visual Localization dataset~\cite{Taira2018InLoc}.
We follow the regular evaluation protocol for local features~\cite{Taira2018InLoc}.
For each query, we retrieve top 10 related images according to NetVLAD~\cite{Arandjelovic2016NetVLAD} global descriptors.
2D-2D matches are established between the query image and each retrieved image.
Next, keypoints in database images are back-projected to 3D using the ground-truth LiDAR scans to obtain a set of 2D-3D matches.
Finally, RANSAC pose estimation is ran for each set of 2D-3D matches and the pose with the highest number of inliers is selected.
For this experiment, we use DoG keypoints extracted using COLMAP.

The results are shown in Table~\ref{tab:inloc}.
As with previous experiments, we notice a significant uptick in performance when matching translated hand-crafted descriptors against natively extracted learned ones, notably for the lower localization threshold.

\subsection{Localization to collaborative maps}

We train our encoder-decoder approach with varied joint embedding dimensions.
We present results in the case of localization to crowd-sourced maps on the Aachen Day-Night dataset~\cite{Sattler2017Benchmarking} in Table~\ref{tab:aachen-emb-dim}.
To recall, in this scenario, the set of database images is split in $4$ balanced subsets, one for each description algorithm.
For query images, we extract SOSNet~\cite{Tian2019SOSNet} descriptors.
Both query and database features are then translated to the joint space for matching.
Increasing the dimensionality past $128$ does not have any benefits in terms of performance.
Interestingly, the $64$-dimensional variant performs better than SIFT~\cite{Lowe2004Distinctive} despite using a heterogeneous map.
Finally, even the $32$-dimensional variant drastically outperforms native BRIEF~\cite{Calonder2010BRIEF} localization.

\setlength{\tabcolsep}{2.0pt}
\begin{table}[t]
	\footnotesize
	\centering
	\begin{tabular}{c c | l l l l }
		\toprule
		\multirow{3}{*}{\rotatebox[origin=c]{90}{Scenario}} & \multirowcell{3}{Descriptor} & \multicolumn{4}{c}{\% localized queries} \\
		& & \multicolumn{2}{c}{Day (824 images)} & \multicolumn{2}{c}{Night (98 images)} \\
		& & $0.25m, 2^\circ$ & $0.5m, 5^\circ$ & $0.25m, 2^\circ$ & $0.5m, 5^\circ$ \\ \midrule\midrule
		\multirow{4}{*}{\rotatebox[origin=c]{90}{Standard}} & BRIEF & \textcolor{OrangeRed}{76.1} & \textcolor{OrangeRed}{81.4} & \textcolor{OrangeRed}{32.7} & \textcolor{OrangeRed}{36.7} \\ 
		& SIFT & \textcolor{Cerulean}{82.5} & \textcolor{Cerulean}{88.7} & \textcolor{Cerulean}{52.0} & \textcolor{Cerulean}{61.2} \\
		& HardNet & \textcolor{OliveGreen}{86.2} & \textcolor{OliveGreen}{92.2} & \textcolor{OliveGreen}{64.3} & \textcolor{OliveGreen}{72.4} \\
		& SOSNet & \textcolor{RoyalPurple}{86.4} & \textcolor{RoyalPurple}{92.7} & \textcolor{RoyalPurple}{65.3} & \textcolor{RoyalPurple}{75.5} \\
		\midrule\midrule
		\multirow{5}{*}{\rotatebox[origin=c]{90}{Collaborative}} & Embed 256 & 84.6 \textsuperscript{\textcolor{RoyalPurple}{-1.8}} & 90.8 \textsuperscript{\textcolor{RoyalPurple}{-1.9}} & 58.2 \textsuperscript{\textcolor{RoyalPurple}{-7.1}} & 64.3 \textsuperscript{\textcolor{RoyalPurple}{-11.2}} \\
		& {\bf Embed 128} & 84.8 \textsuperscript{\textcolor{RoyalPurple}{-1.6}} & 90.9 \textsuperscript{\textcolor{RoyalPurple}{-1.8}} & 57.1 \textsuperscript{\textcolor{RoyalPurple}{-8.2}} & 60.2 \textsuperscript{\textcolor{RoyalPurple}{-15.3}} \\
		& Embed 64 & 83.7 \textsuperscript{\textcolor{RoyalPurple}{-2.7}} & 89.1 \textsuperscript{\textcolor{RoyalPurple}{-3.6}} & 56.1 \textsuperscript{\textcolor{RoyalPurple}{-9.2}} & 61.2 \textsuperscript{\textcolor{RoyalPurple}{-14.3}} \\
		& Embed 32 & 80.3 \textsuperscript{\textcolor{RoyalPurple}{-6.1}} & 86.8 \textsuperscript{\textcolor{RoyalPurple}{-5.9}} & 45.9 \textsuperscript{\textcolor{RoyalPurple}{-19.4}} & 49.0 \textsuperscript{\textcolor{RoyalPurple}{-26.5}} \\
		& Embed 16 & 75.1 \textsuperscript{\textcolor{RoyalPurple}{-11.3}} & 80.8 \textsuperscript{\textcolor{RoyalPurple}{-11.9}} & 30.6 \textsuperscript{\textcolor{RoyalPurple}{-34.7}} & 33.7 \textsuperscript{\textcolor{RoyalPurple}{-41.8}} \\ \bottomrule
	\end{tabular}
	\vspace{-7.5pt}
	\caption{{\bf Localization to collaborative maps -- embedding dimension.} The database images are partitioned in $4$ balanced subsets, one for each description algorithm. We use SOSNet for query images. Both database and query descriptors are translated to the joint embedding space.}
	\label{tab:aachen-emb-dim}
\end{table}
\setlength{\tabcolsep}{\tabcolsepdefault}

\subsection{HPatches sequences}

\begin{figure}
	\begin{subfigure}{\columnwidth}
		\centering
		\includegraphics[width=.8\textwidth]{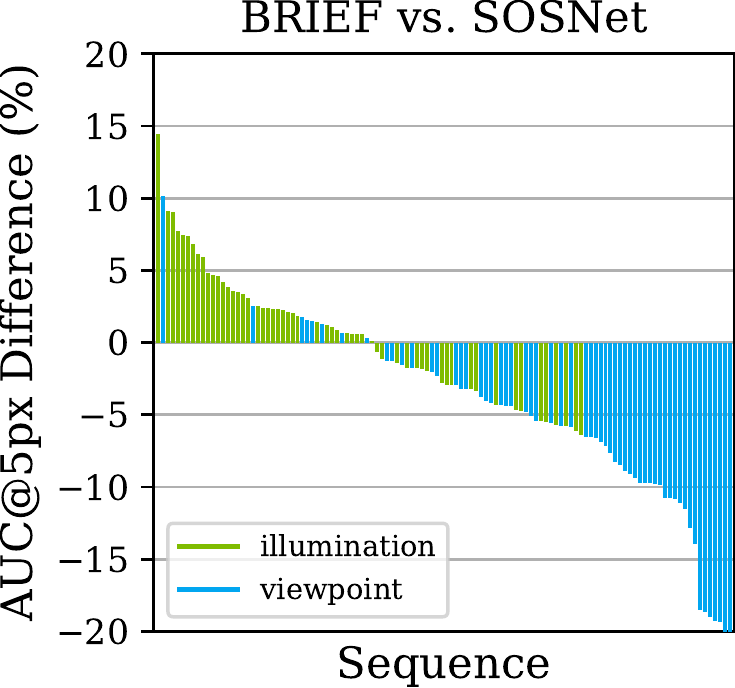}
	\end{subfigure}
	
	\vspace{7.5pt}
	\begin{subfigure}{\columnwidth}
		\centering
		\includegraphics[width=.8\textwidth]{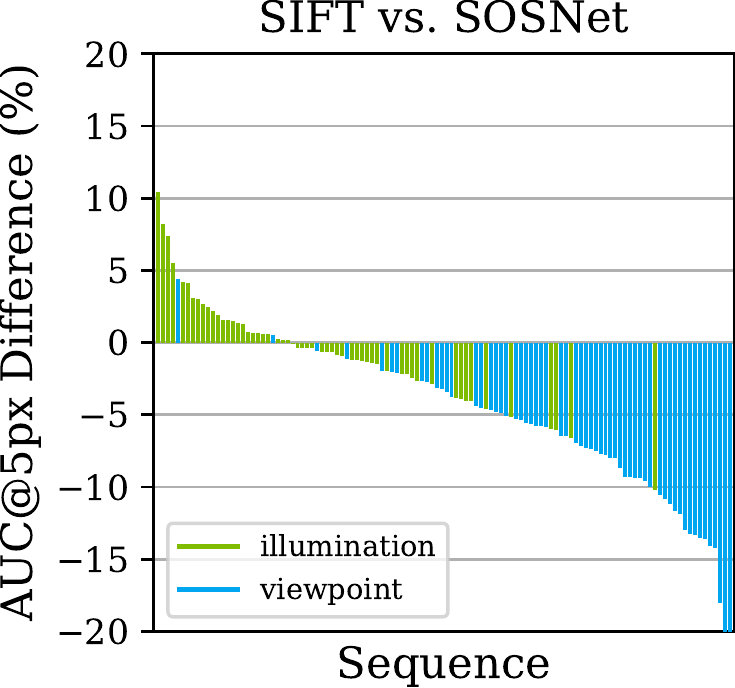}
	\end{subfigure}
	
	\vspace{7.5pt}
	\begin{subfigure}{\columnwidth}
		\centering
		\includegraphics[width=.8\textwidth]{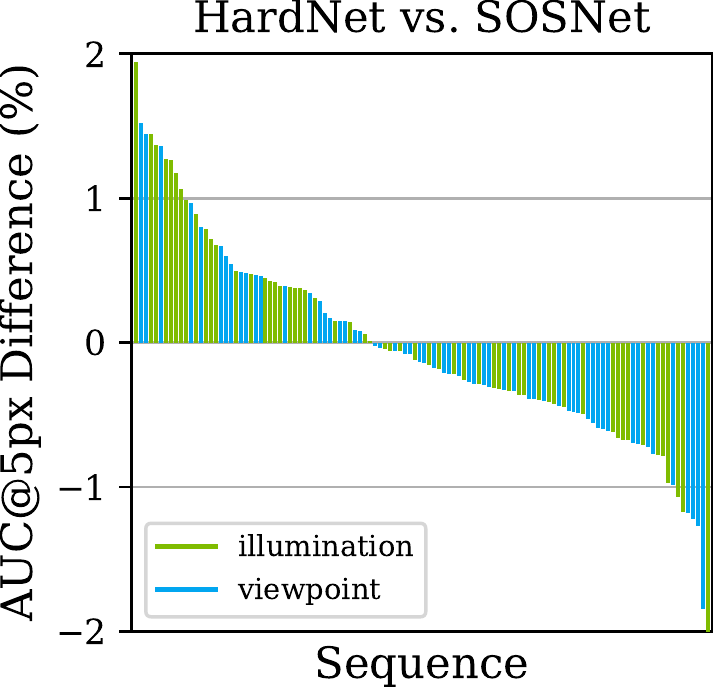}
	\end{subfigure}
	\caption{{\bf HPatches sequences breakdown.} We report the per-scene absolute difference in the area under the mean matching accuracy curve up to 5 pixels between different descriptors. While SOSNet has a better overall performance, it does not outperform BRIEF or SIFT on all scenes. Similarly, while smaller, there are still scene-to-scene differences between the two learned descriptors.}
	\label{fig:hseq}
\end{figure}

We present a per-scene comparison between state-of-the-art descriptors on the full HPatches sequences~\cite{Balntas2017HPatches} following the evaluation protocol of Dusmanu~\etal~\cite{Dusmanu2019D2}.
We report the absolute difference between the area under the mean matching accuracy curve up to $5$ pixels for different pairs of descriptors in Figure~\ref{fig:hseq}.
Despite SOSNet~\cite{Tian2019SOSNet} drastically outperforming BRIEF~\cite{Calonder2010BRIEF} and SIFT~\cite{Lowe2004Distinctive} overall, the handcrafted descriptors achieve better matching accuracy on a significant number of scenes.
Similarly, while HardNet~\cite{Mishchuk2017Working} and SOSNet have a similar overall performance, there are still small scene-to-scene variations.
Thus, it is unclear whether existing learned descriptors (\eg, SOSNet) are the best under all conditions.

	\section{Further implementation details}

{\bf To encourage and facilitate future research on the topic of collaborative localization and mapping from heterogeneous devices, the code of our method and the evaluation protocols will be released as open source at \url{https://github.com/mihaidusmanu/cross-descriptor-vis-loc-map}.}

The architectures used throughout our experiments are detailed in Table~\ref{tab:architectures}.
Our approach was implemented in Python using PyTorch~\cite{Paszke2019PyTorch} and Kornia~\cite{Riba2019Kornia}.
For the learned descriptors, we use the official Liberty~\cite{Winder2007Learning} pre-trained weights released by the authors.
Training the encoder-decoder approach for all $4$ description algorithms takes around $30$ minutes on a single NVIDIA RTX 2080Ti.

\begin{table*}[t!]
	\begin{minipage}{0.30\textwidth}
		\centering
		\footnotesize
		\begin{tabular}{c c c c}
			\toprule
			\multicolumn{4}{c}{BRIEF} \\ \midrule
			\multirowcell{2}{Layer} & \multirowcell{2}{Batch\\norm.} & \multirowcell{2}{Activation} & \multirowcell{2}{Output\\dim.} \\ 
			& & & \\ \midrule\midrule
			\multicolumn{4}{c}{Encoder} \\ \midrule
			\texttt{input} & & & $512$ \\
			\texttt{hidden1} & \checkmark & ReLU & $1024$ \\
			\texttt{hidden2} & \checkmark & ReLU & $1024$ \\
			\texttt{embed} & & & $128$ \\ \midrule\midrule
			\multicolumn{4}{c}{Decoder} \\ \midrule
			\texttt{embed} & & & $128$ \\
			\texttt{hidden1} & \checkmark & ReLU & $1024$ \\
			\texttt{hidden2} & \checkmark & ReLU & $1024$ \\
			\texttt{output} & & Sigmoid & $512$ \\
			\bottomrule
		\end{tabular}
	\end{minipage}
	\hfill
	\begin{minipage}{0.30\textwidth}
		\centering
		\footnotesize
		\begin{tabular}{c c c c}
			\toprule
			\multicolumn{4}{c}{SIFT} \\ \midrule
			\multirowcell{2}{Layer} & \multirowcell{2}{Batch\\norm.} & \multirowcell{2}{Activation} & \multirowcell{2}{Output\\dim.} \\ 
			& & & \\ \midrule\midrule
			\multicolumn{4}{c}{Encoder} \\ \midrule
			\texttt{input} & & & $128$ \\
			\texttt{hidden1} & \checkmark & ReLU & $1024$ \\
			\texttt{hidden2} & \checkmark & ReLU & $1024$ \\
			\texttt{embed} & & & $128$ \\ \midrule\midrule
			\multicolumn{4}{c}{Decoder} \\ \midrule
			\texttt{embed} & & & $128$ \\
			\texttt{hidden1} & \checkmark & ReLU & $1024$ \\
			\texttt{hidden2} & \checkmark & ReLU & $1024$ \\
			\texttt{output} & & ReLU & $128$ \\
			\bottomrule
		\end{tabular}
	\end{minipage}
	\hfill
	\begin{minipage}{0.30\textwidth}
		\centering
		\footnotesize
		\begin{tabular}{c c c c}
			\toprule
			\multicolumn{4}{c}{HardNet / SOSNet} \\ \midrule
			\multirowcell{2}{Layer} & \multirowcell{2}{Batch\\norm.} & \multirowcell{2}{Activation} & \multirowcell{2}{Output\\dim.} \\ 
			& & & \\ \midrule\midrule
			\multicolumn{4}{c}{Encoder} \\ \midrule
			\texttt{input} & & & $128$ \\
			\texttt{hidden1} & \checkmark & ReLU & $256$ \\
			\texttt{hidden2} & \checkmark & ReLU & $256$ \\
			\texttt{embed} & & & $128$ \\ \midrule\midrule
			\multicolumn{4}{c}{Decoder} \\ \midrule
			\texttt{embed} & & & $128$ \\
			\texttt{hidden1} & \checkmark & ReLU & $256$ \\
			\texttt{hidden2} & \checkmark & ReLU & $256$ \\
			\texttt{output} & & & $128$ \\
			\bottomrule
		\end{tabular}
	\end{minipage}
	\caption{{\bf Architectures.} We use shallow MLPs with $2$ hidden layers for all methods. For the handcrafted algorithms (BRIEF~\cite{Calonder2010BRIEF}, SIFT~\cite{Lowe2004Distinctive}) we use larger hidden layers as these descriptors encode lower level image structures. The joint embedding is $\ell_2$ normalized and so is the output if required (\ie, in the case of SIFT~\cite{Lowe2004Distinctive}, HardNet~\cite{Mishchuk2017Working}, SOSNet~\cite{Tian2019SOSNet}).}
	\label{tab:architectures}
\end{table*}
	\section{Pseudo-ground-truth generation}

Similar to other datasets~\cite{Sattler2017Benchmarking,Jin2020Image}, we generate pseudo-ground-truth intrinsics and extrinsics for the Local Feature Evaluation benchmark~\cite{Schonberger2017Comparative} via an initial Structure-from-Motion process.
For each dataset, there are four steps:
\begin{tight_itemize}
	\item We extract SOSNet~\cite{Tian2019SOSNet} descriptors around DoG~\cite{Lowe2004Distinctive} keypoints obtained using COLMAP~\cite{Schoenberger2016Structure} with default parameters. We exhaustively match all images using a mutual nearest neighbors matcher enforcing the ratio test~\cite{Lowe2004Distinctive} with a threshold of $0.9$.
	\item We run COLMAP~\cite{Schoenberger2016Structure} for geometric verification and mapping. All images with less than $100$ 3D points are not considered during the next steps.
	\item We run geometric verification and mapping again on the remaining images -- this time all intrinsics are fixed to the estimates from the previous step.
	\item We rescale the final model with respect to Google Maps by manual correspondence clicking to obtain final pseudo-ground-truth metric poses.
\end{tight_itemize}
	\section{Collaborative mapping -- co-visibility}

\begin{figure*}[p]
	\begin{minipage}{.98\columnwidth}
		\centering
		Madrid Metropolis\\
		\begin{minipage}{0.47\columnwidth}
			\centering
			\includegraphics[width=\textwidth]{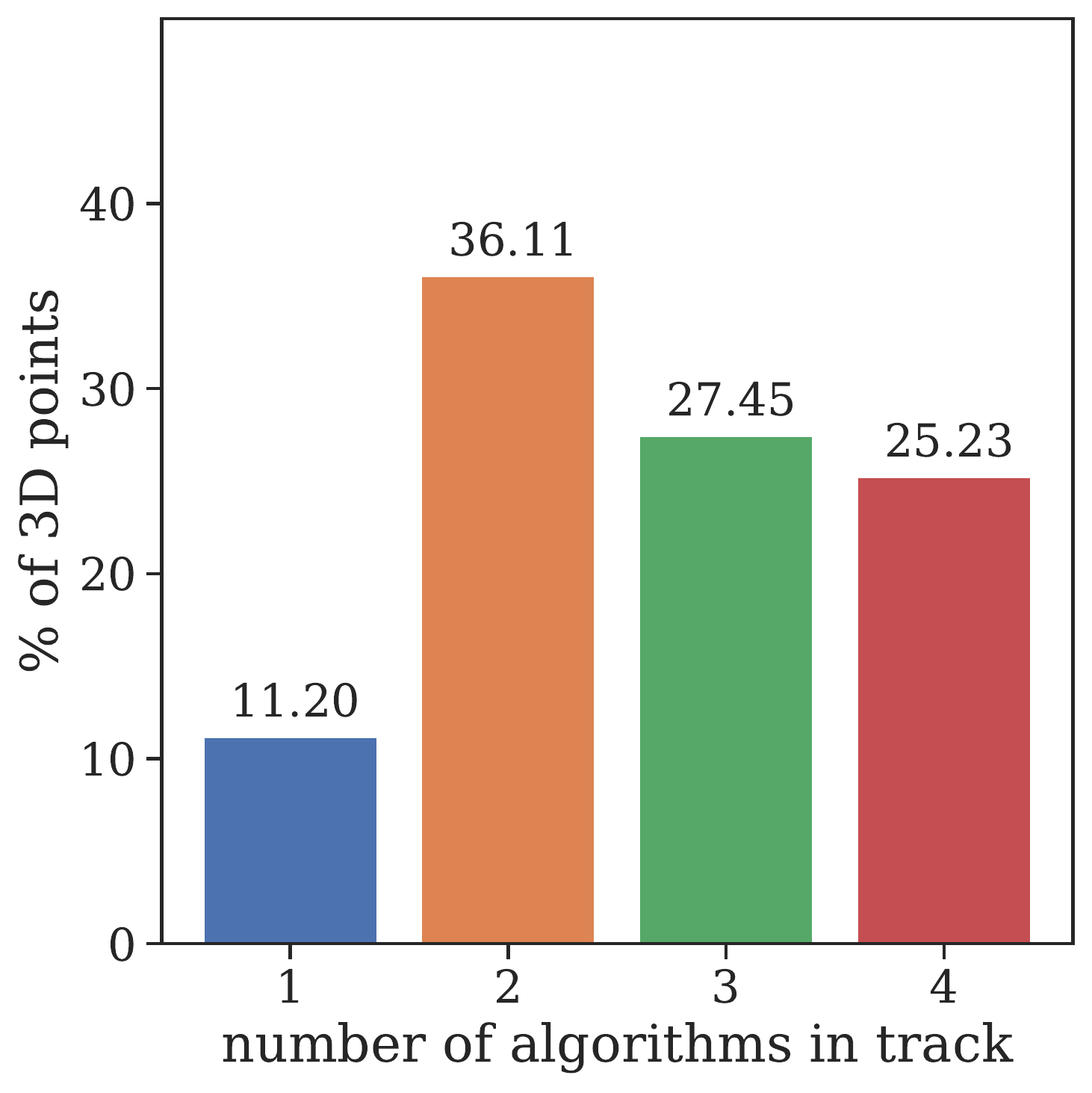}
		\end{minipage}
		\hfill
		\begin{minipage}{0.51\columnwidth}
			\centering
			\includegraphics[width=\textwidth]{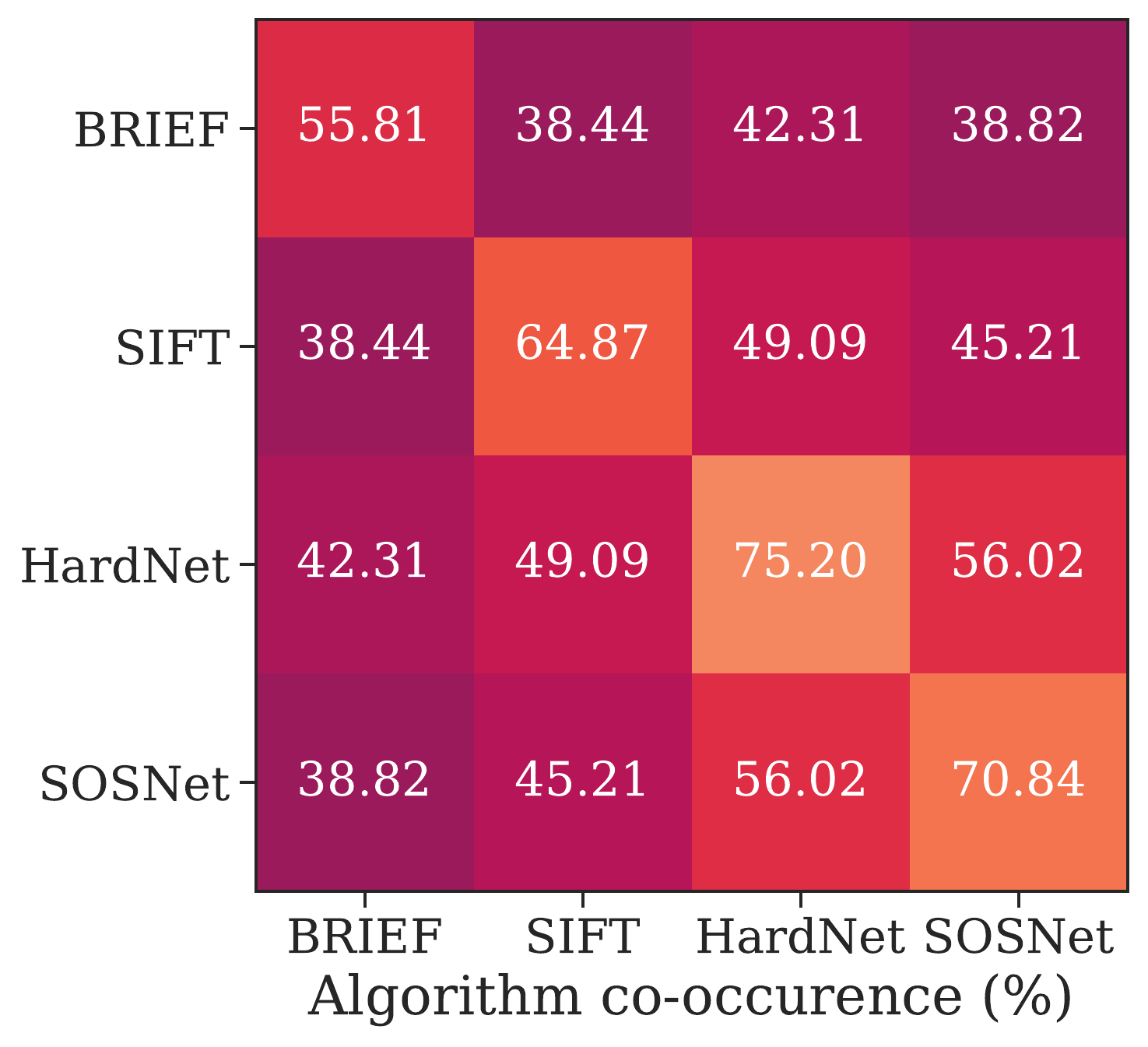}
		\end{minipage}\\
		\vspace{5pt}Gendarmenmarkt\\
		\begin{minipage}{0.47\columnwidth}
			\centering
			\includegraphics[width=\textwidth]{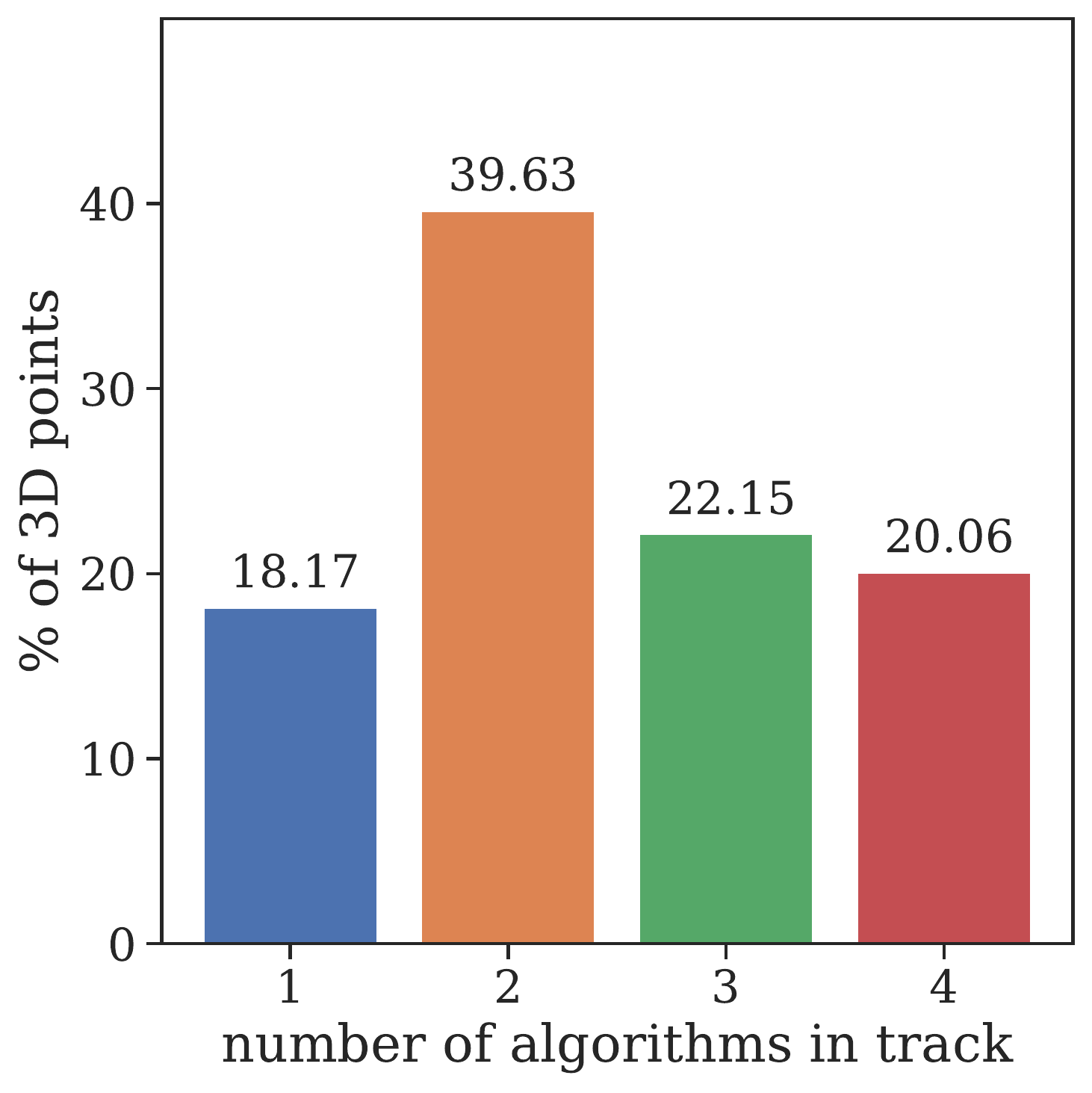}
		\end{minipage}
		\hfill
		\begin{minipage}{0.51\columnwidth}
			\centering
			\includegraphics[width=\textwidth]{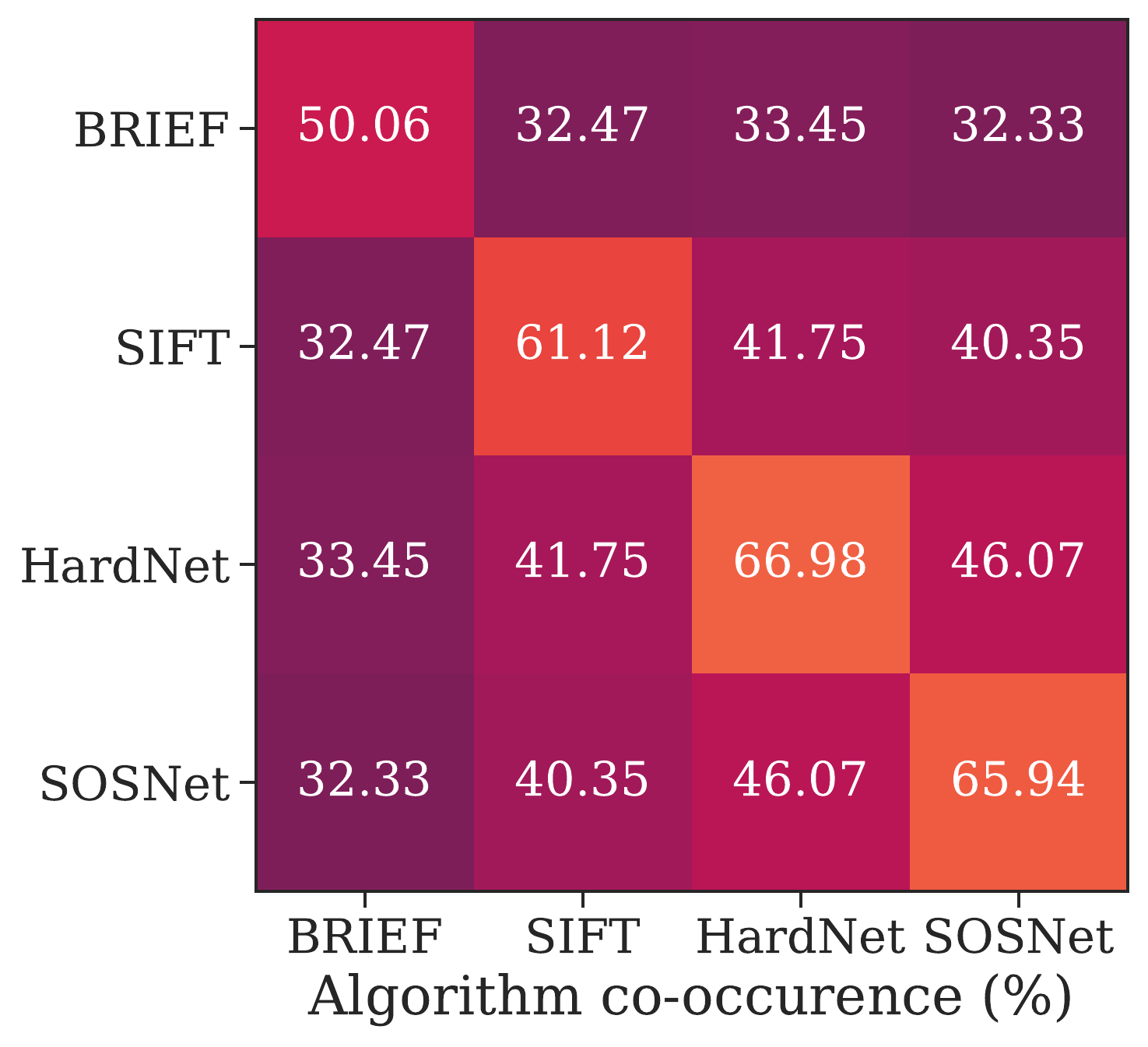}
		\end{minipage}\\
		\vspace{5pt}Tower of London\\
		\begin{minipage}{0.47\columnwidth}
			\centering
			\includegraphics[width=\textwidth]{assets/sfm/nb_feats_tower}
		\end{minipage}
		\hfill
		\begin{minipage}{0.51\columnwidth}
			\centering
			\includegraphics[width=\textwidth]{assets/sfm/cooc_tower}
		\end{minipage}
		\vspace{-7.5pt}
		\captionof{figure}{{\bf Co-visibility statistics -- ``Embed".} For the ``Embed" approach, we report the \% of 3D points containing $1-4$ distinct algorithms in their tracks on the left. On the right, we visualize the co-occurence, \ie, the percentage of 3D points containing descriptors originating from $2$ given description algorithms in their tracks.}
		\label{fig:sfm-stats-all}
	\end{minipage}
	\hfill
	\begin{minipage}{.98\columnwidth}
		\centering
		Madrid Metropolis\\
		\begin{minipage}{0.47\columnwidth}
			\centering
			\includegraphics[width=\textwidth]{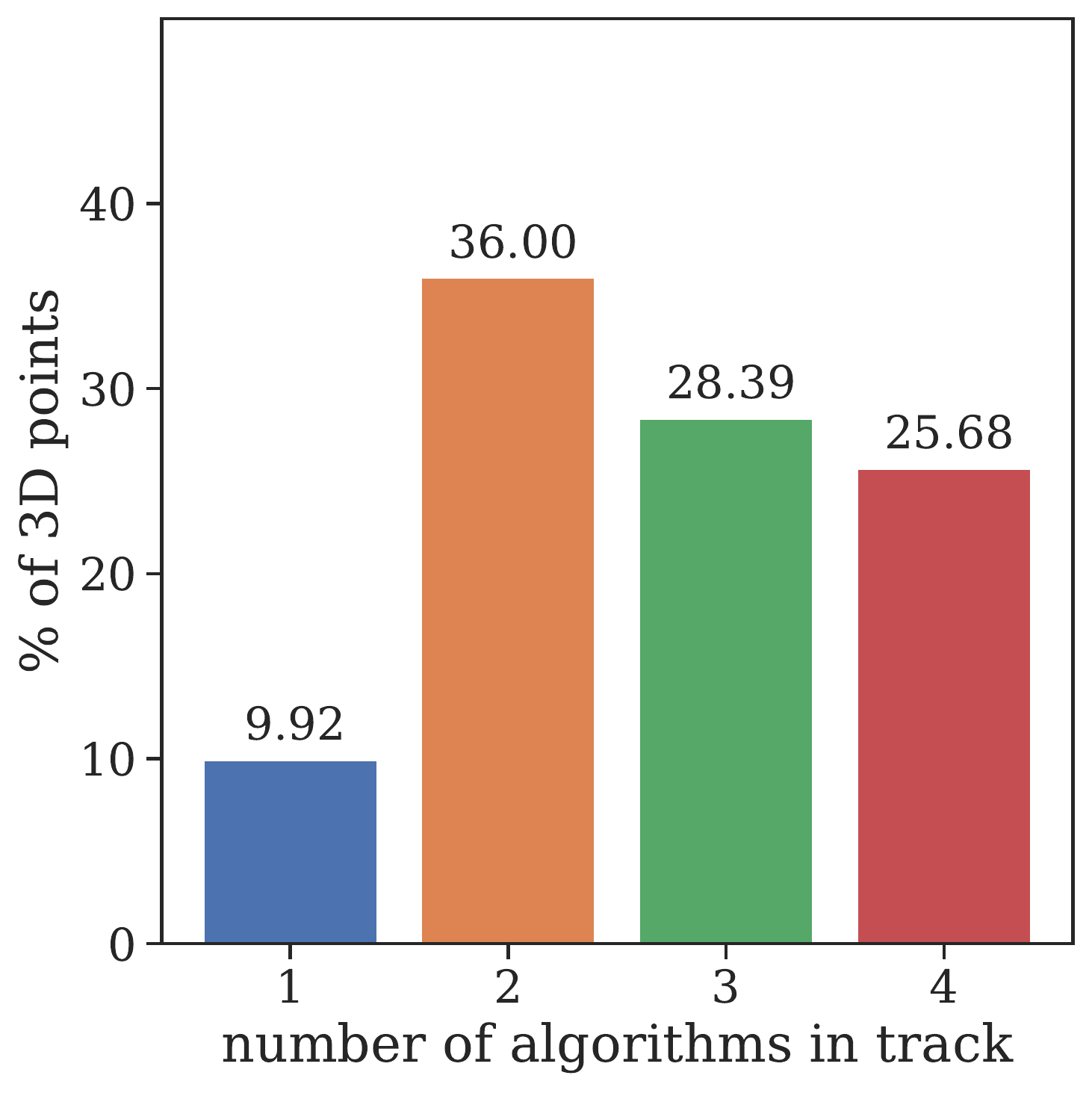}
		\end{minipage}
		\hfill
		\begin{minipage}{0.51\columnwidth}
			\centering
			\includegraphics[width=\textwidth]{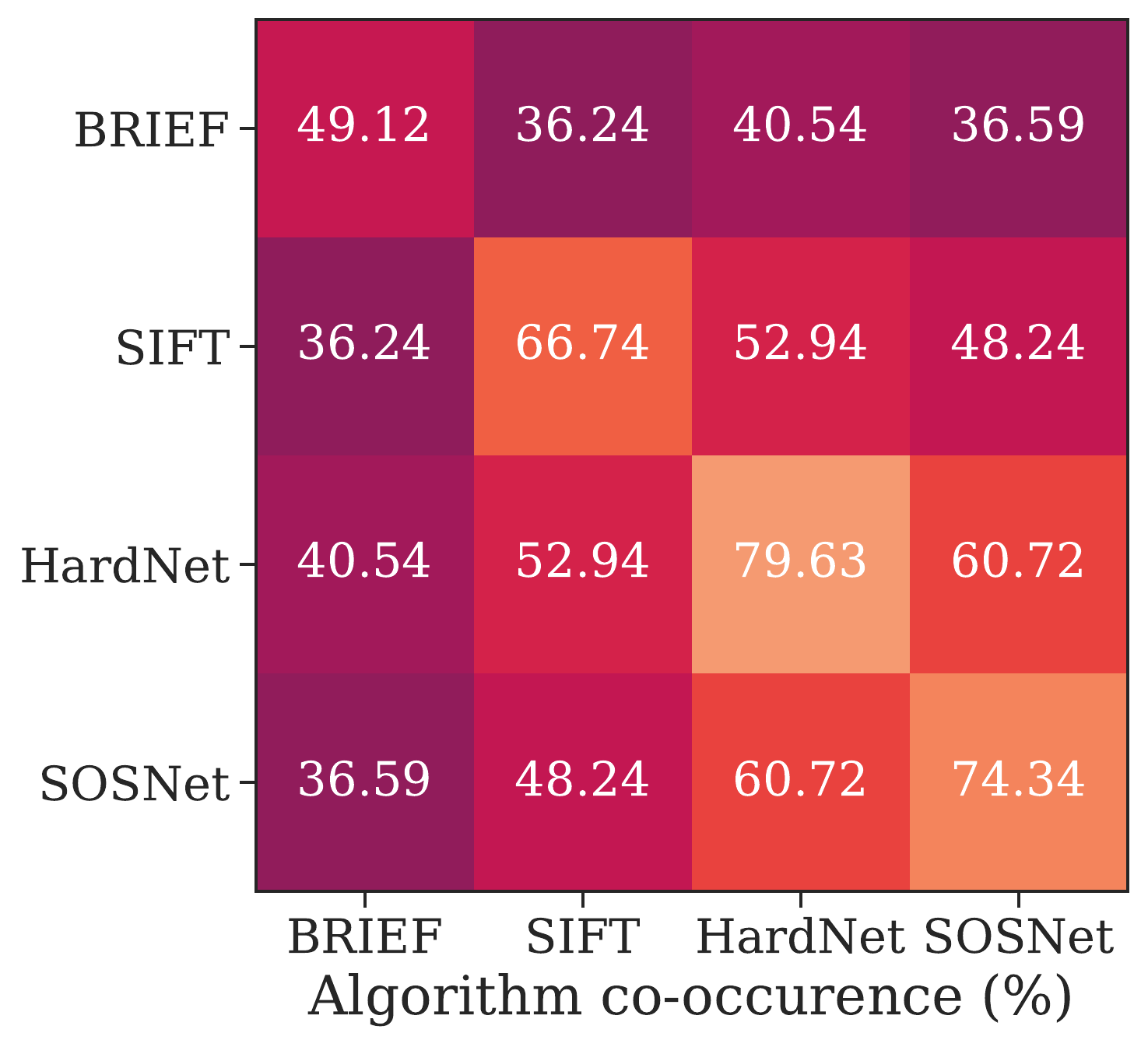}
		\end{minipage}\\
		\vspace{5pt}Gendarmenmarkt\\
		\begin{minipage}{0.47\columnwidth}
			\centering
			\includegraphics[width=\textwidth]{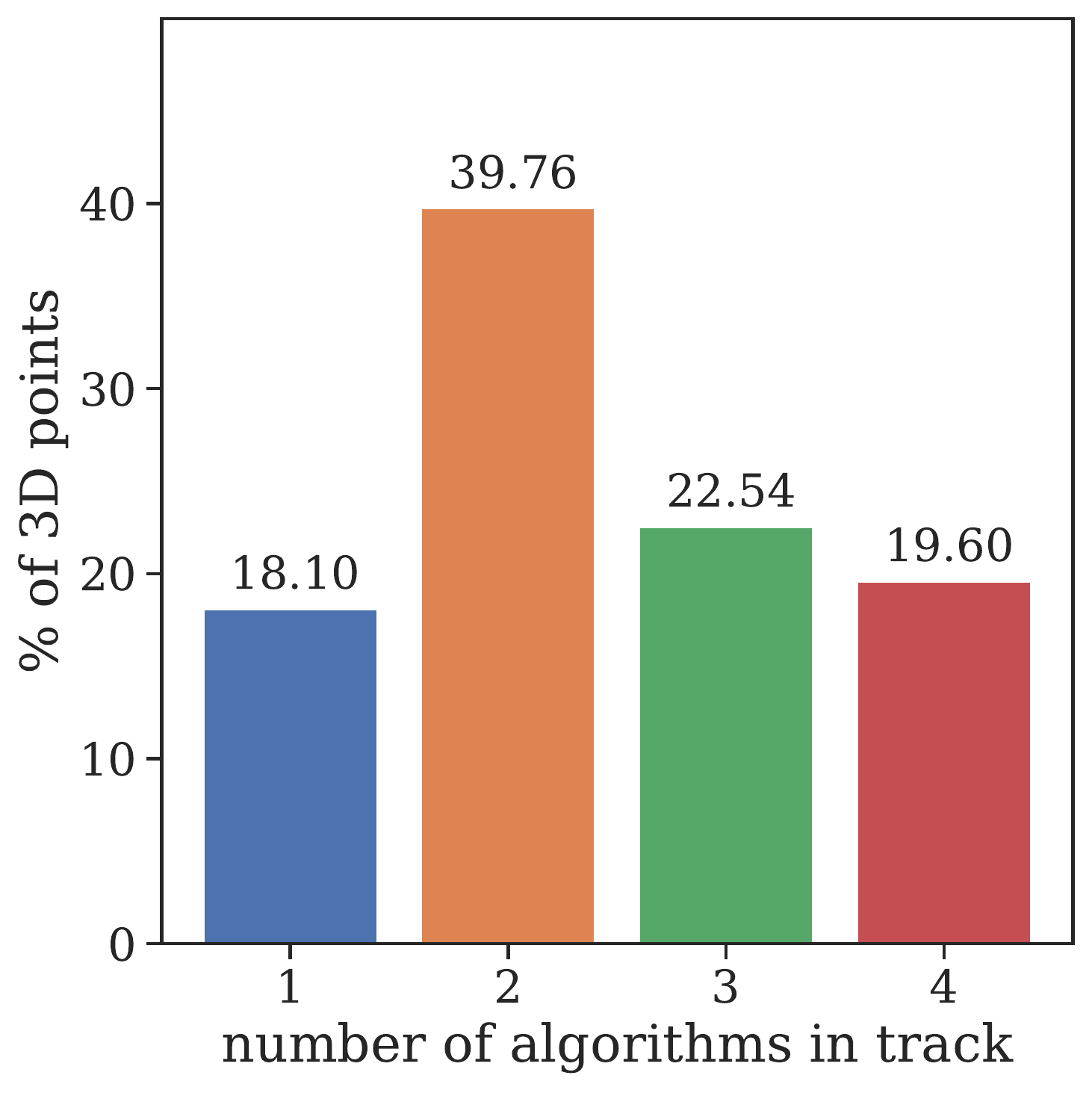}
		\end{minipage}
		\hfill
		\begin{minipage}{0.51\columnwidth}
			\centering
			\includegraphics[width=\textwidth]{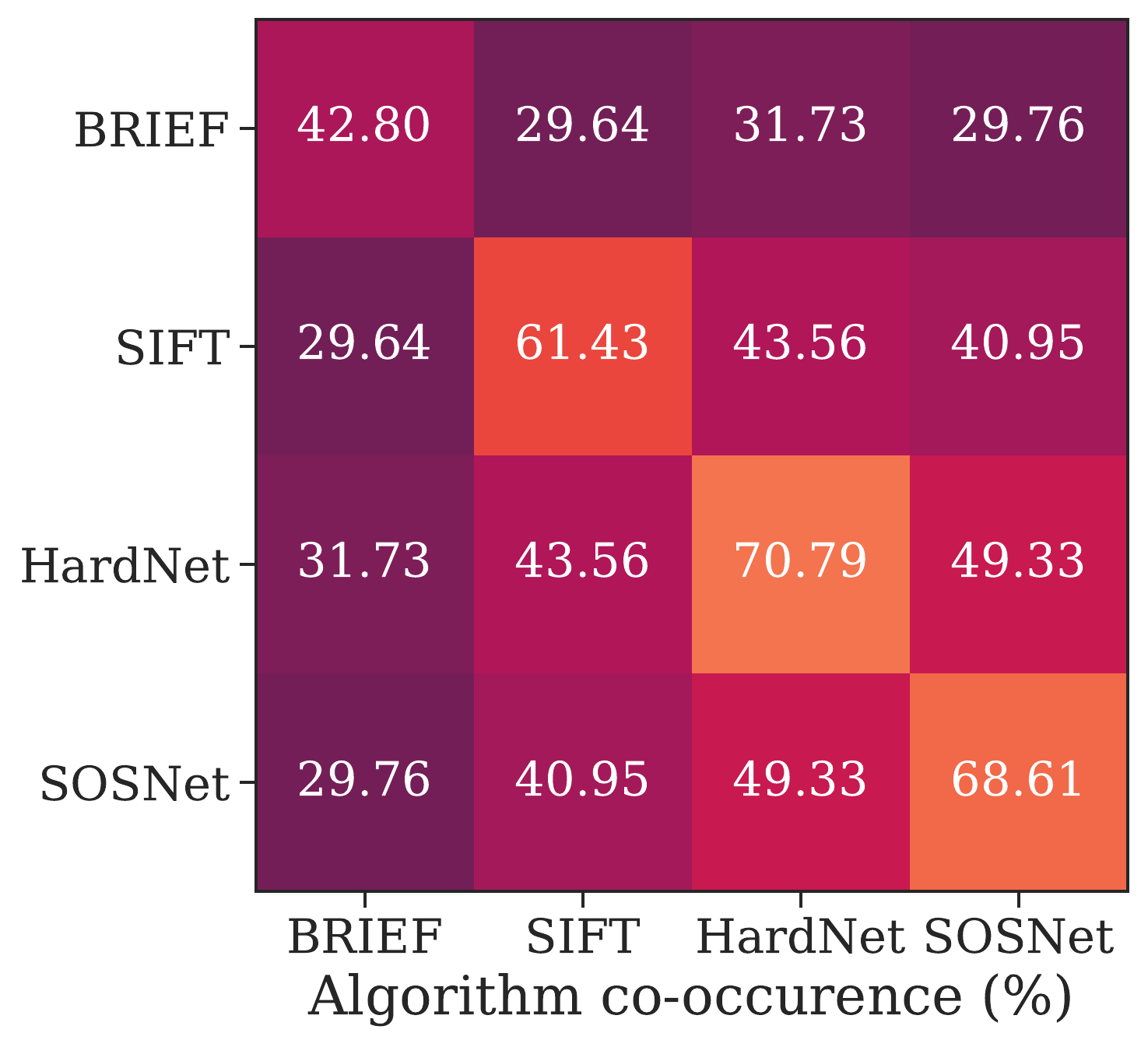}
		\end{minipage}\\
		\vspace{5pt}Tower of London\\
		\begin{minipage}{0.47\columnwidth}
			\centering
			\includegraphics[width=\textwidth]{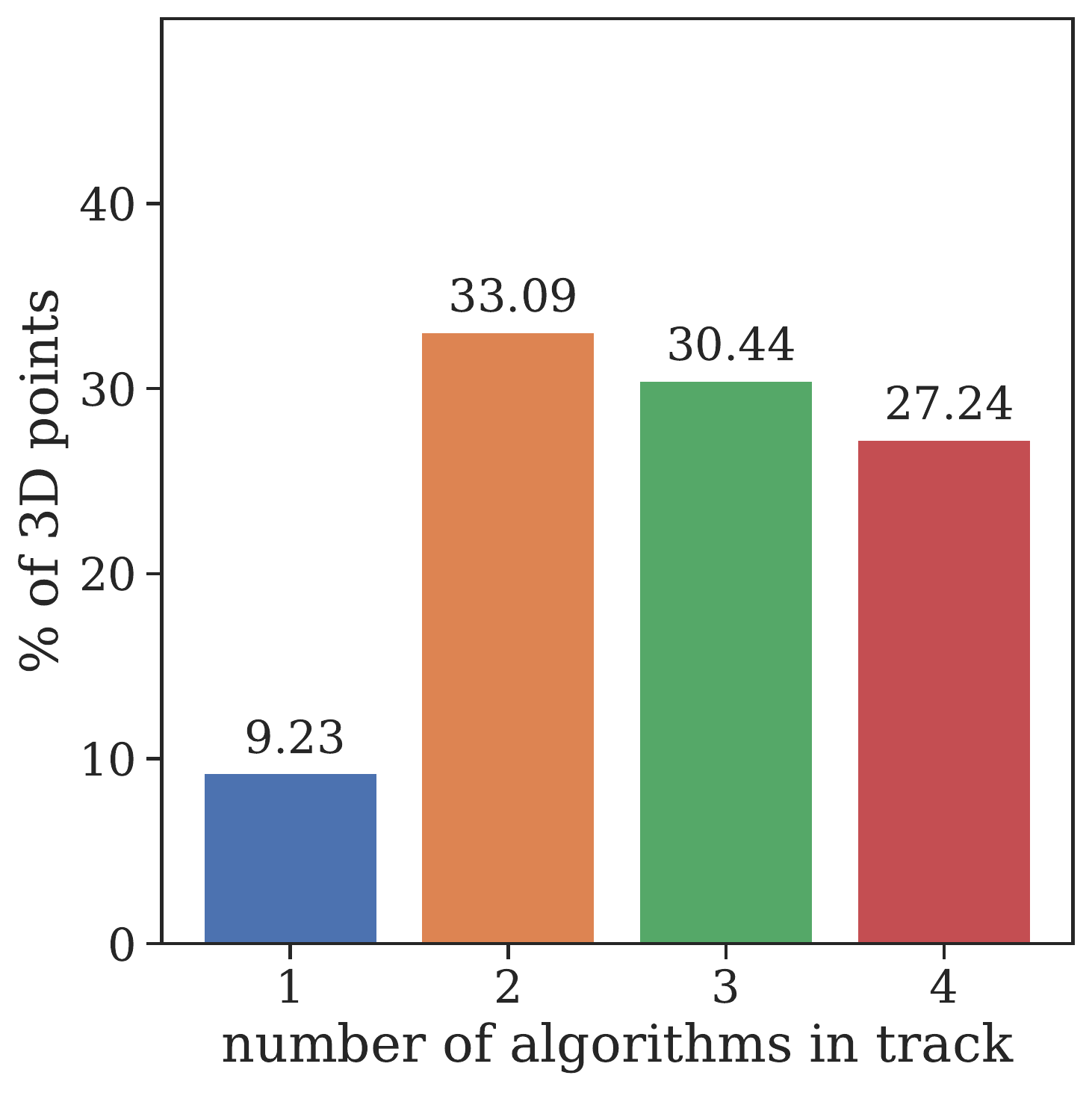}
		\end{minipage}
		\hfill
		\begin{minipage}{0.51\columnwidth}
			\centering
			\includegraphics[width=\textwidth]{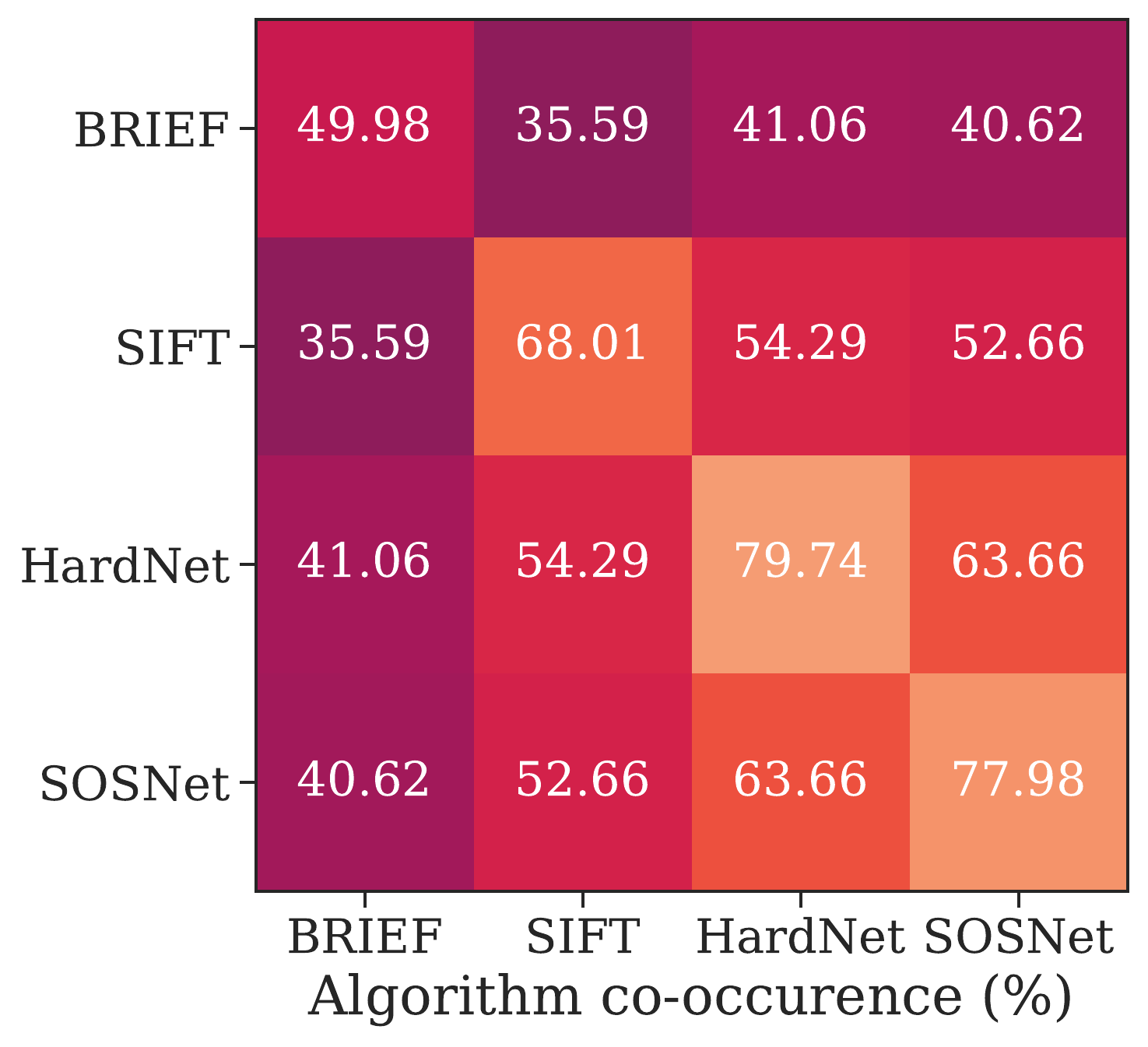}
		\end{minipage}
		\vspace{-7.5pt}
		\captionof{figure}{{\bf Co-visibility statistics -- ``Progressive".} For the ``Progressive" approach, we report the \% of 3D points containing $1-4$ distinct algorithms in their tracks on the left. On the right, we visualize the co-occurence, \ie, the percentage of 3D points containing descriptors originating from $2$ given description algorithms in their tracks.}
		\label{fig:sfm-stats-progressive-all}
	\end{minipage}
\end{figure*}

Figures~\ref{fig:sfm-stats-all} and~\ref{fig:sfm-stats-progressive-all} report additional co-visibility statistics for our approaches to collaborative mapping from heterogeneous descriptors on the benchmark of Sch\"onberger~\etal~\cite{Schonberger2017Comparative}.
The ``Embed" approach translating everything to the joint embedding space generally manages to have more balanced models.
This is especially noticeable in the percentage of 3D points containing at least one BRIEF descriptor in their tracks.
However, this comes at a cost, as learned features (\ie, HardNet, SOSNet) are less represented than in the ``Progressive" approach.

\begin{sidewaysfigure*}
	\begin{tabular}{c c c c c}
		\multicolumn{5}{c}{\large Madrid Metropolis}\\
		BRIEF & SIFT & HardNet & SOSNet & ``Embed"\\ \includegraphics[width=0.19\textwidth]{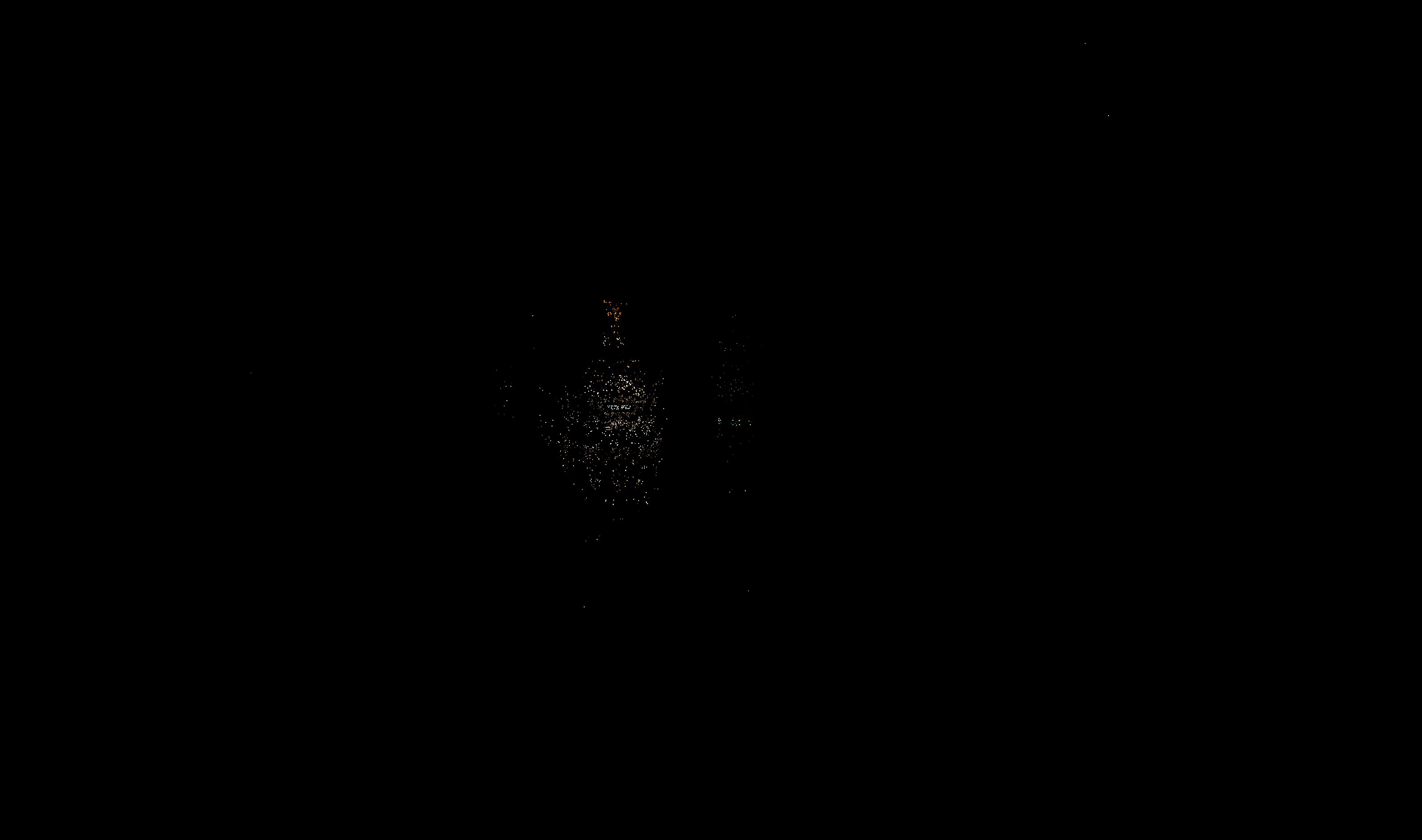} & \includegraphics[width=0.19\textwidth]{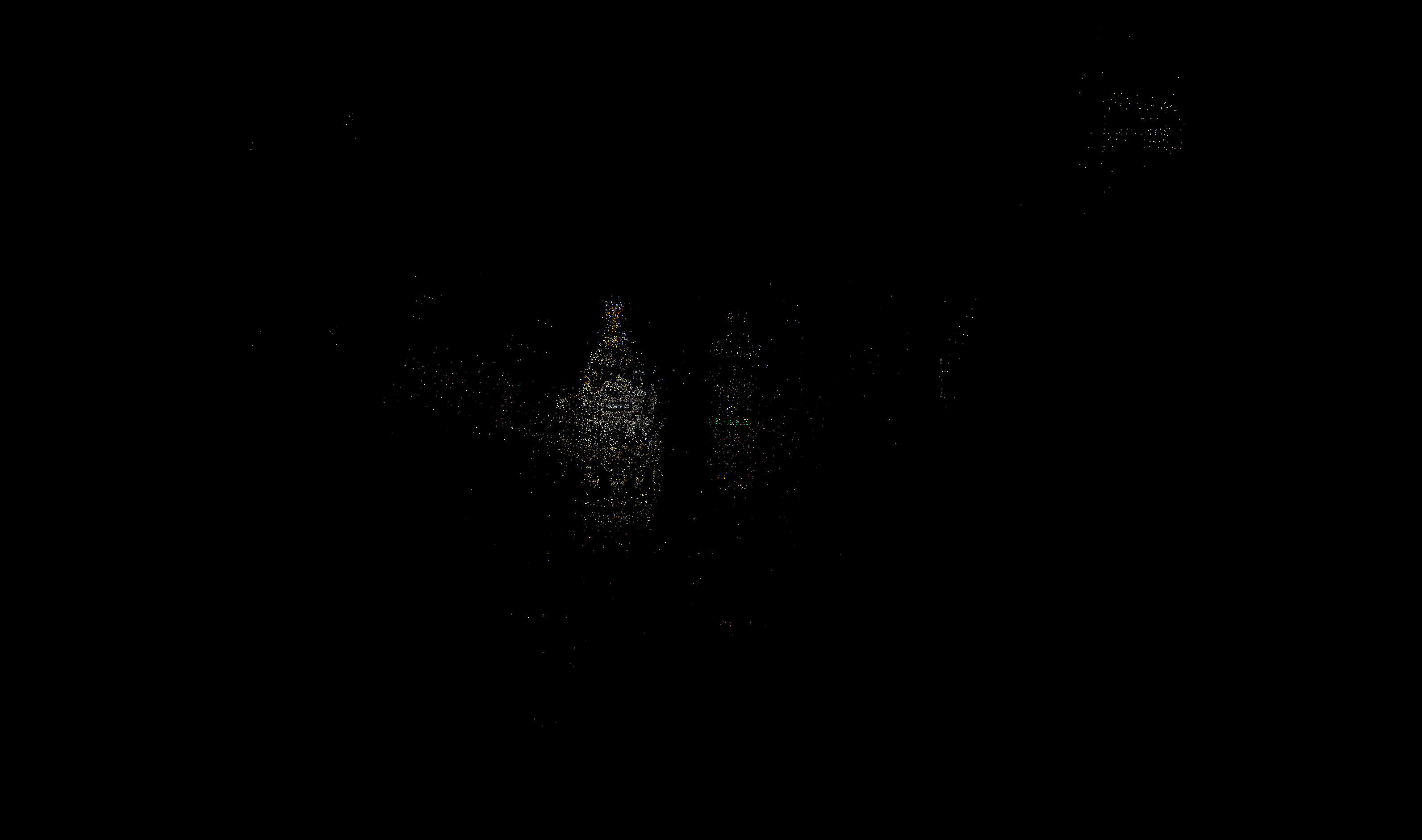} & \includegraphics[width=0.19\textwidth]{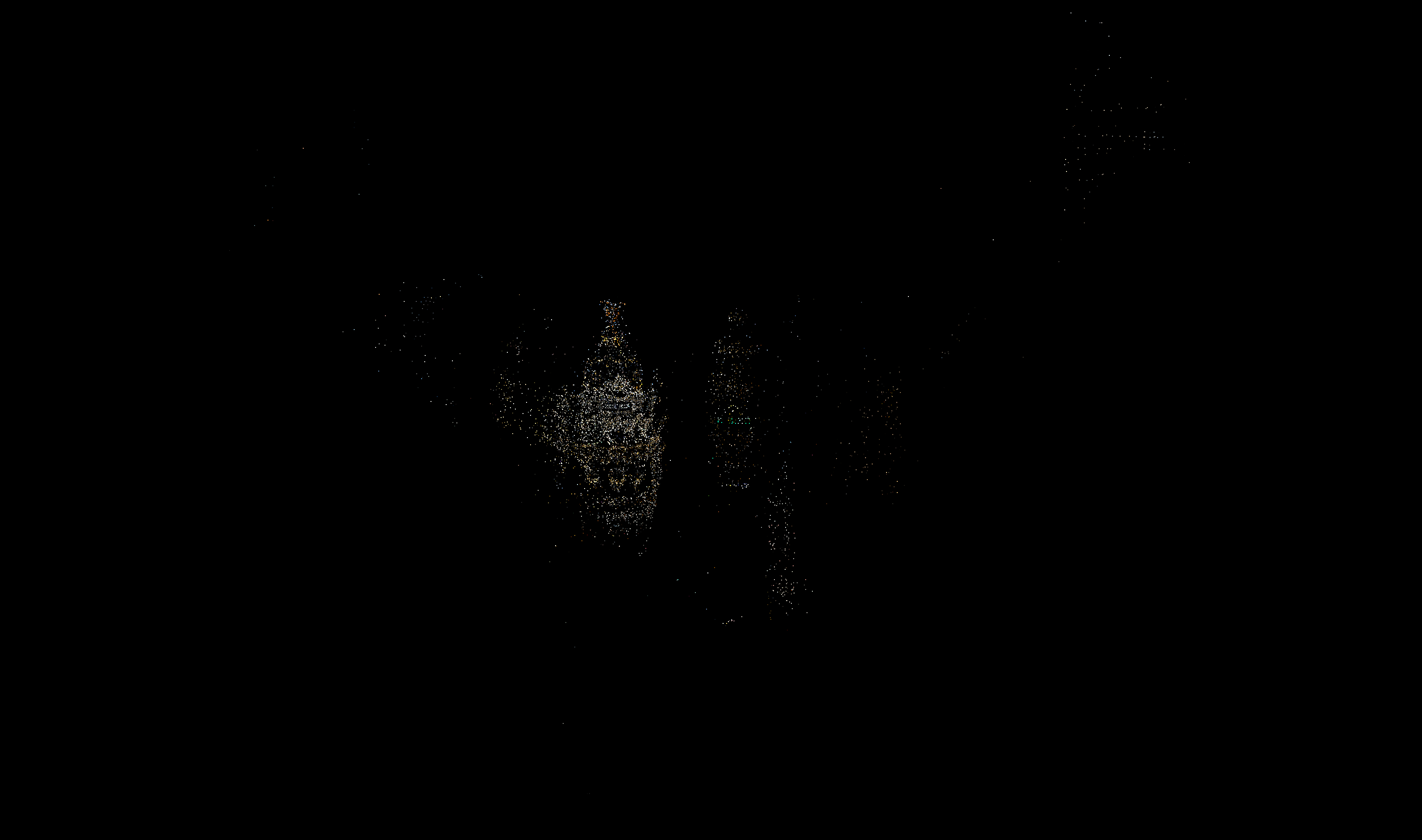} & \includegraphics[width=0.19\textwidth]{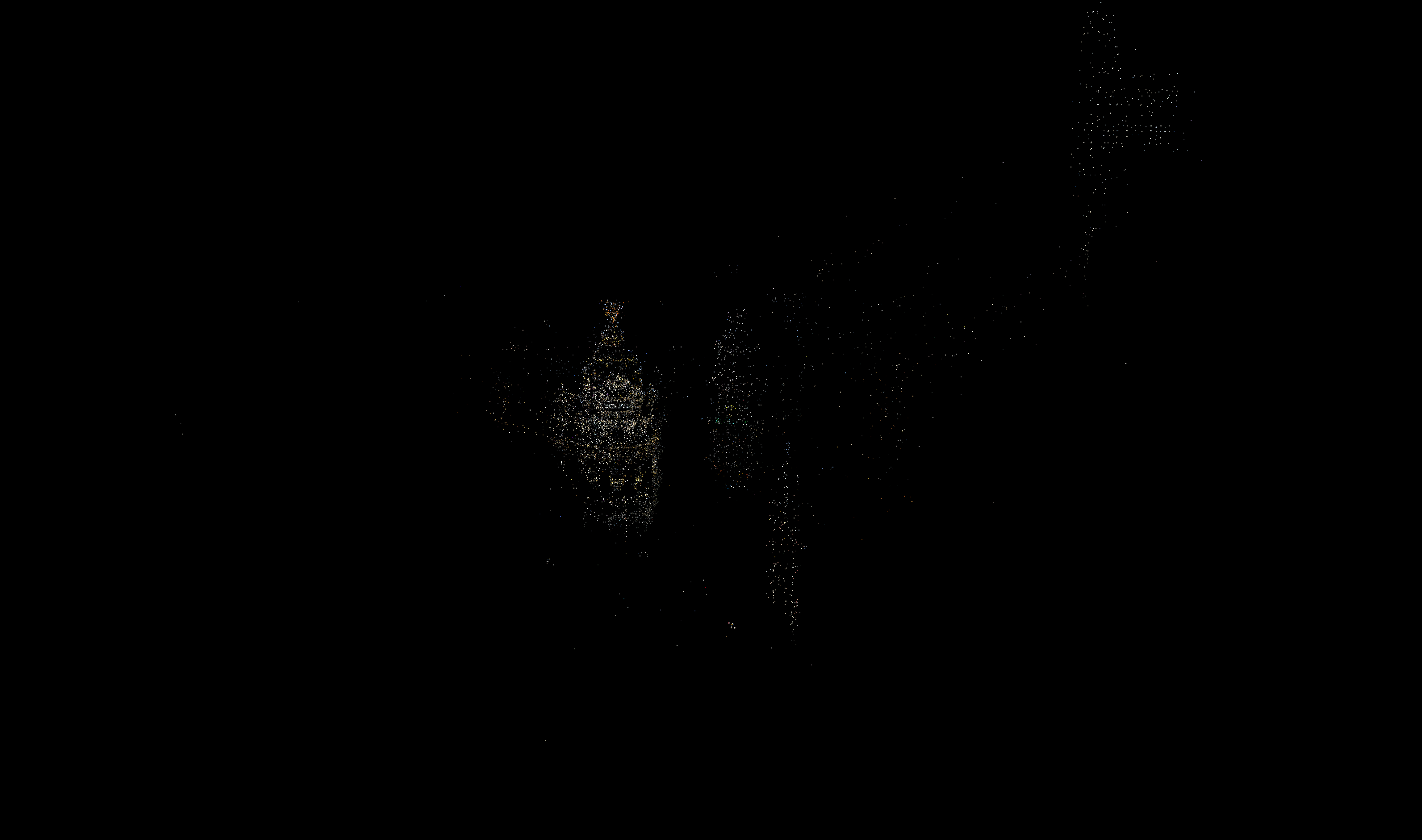} & \includegraphics[width=0.19\textwidth]{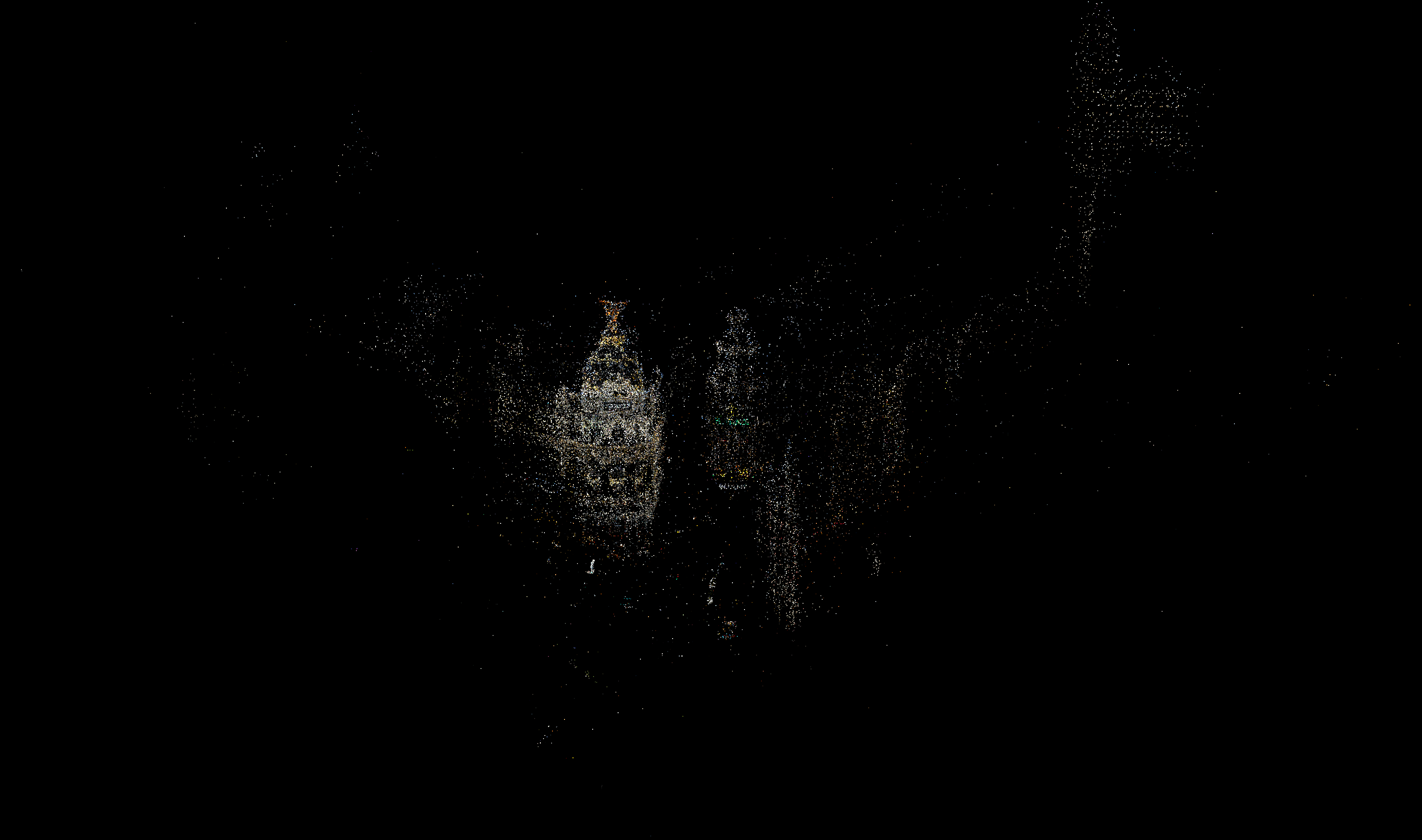}\vspace{20pt} \\
		\multicolumn{5}{c}{\large Gendarmenmarkt}\\
		BRIEF & SIFT & HardNet & SOSNet & ``Embed"\\ \includegraphics[width=0.19\textwidth]{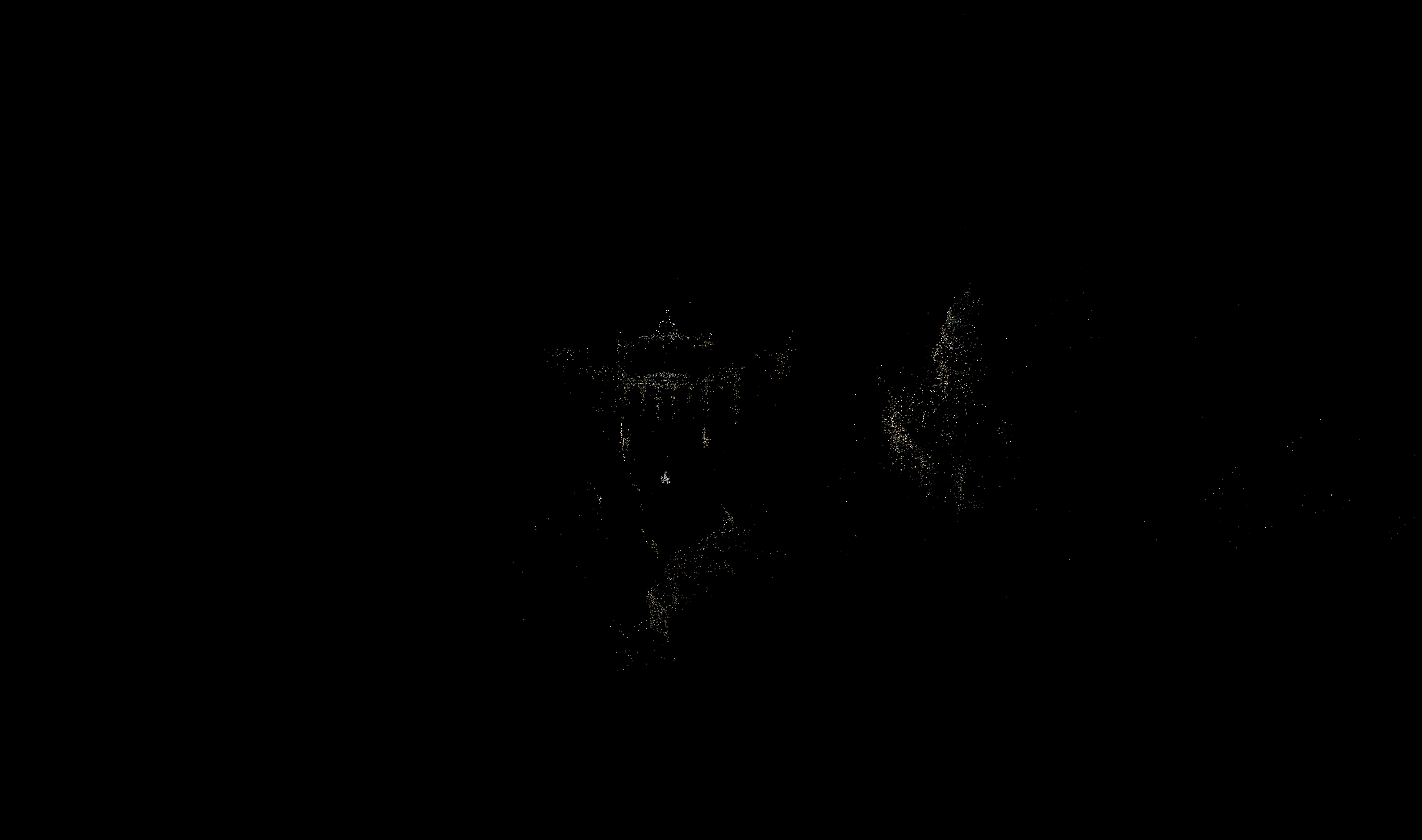} & \includegraphics[width=0.19\textwidth]{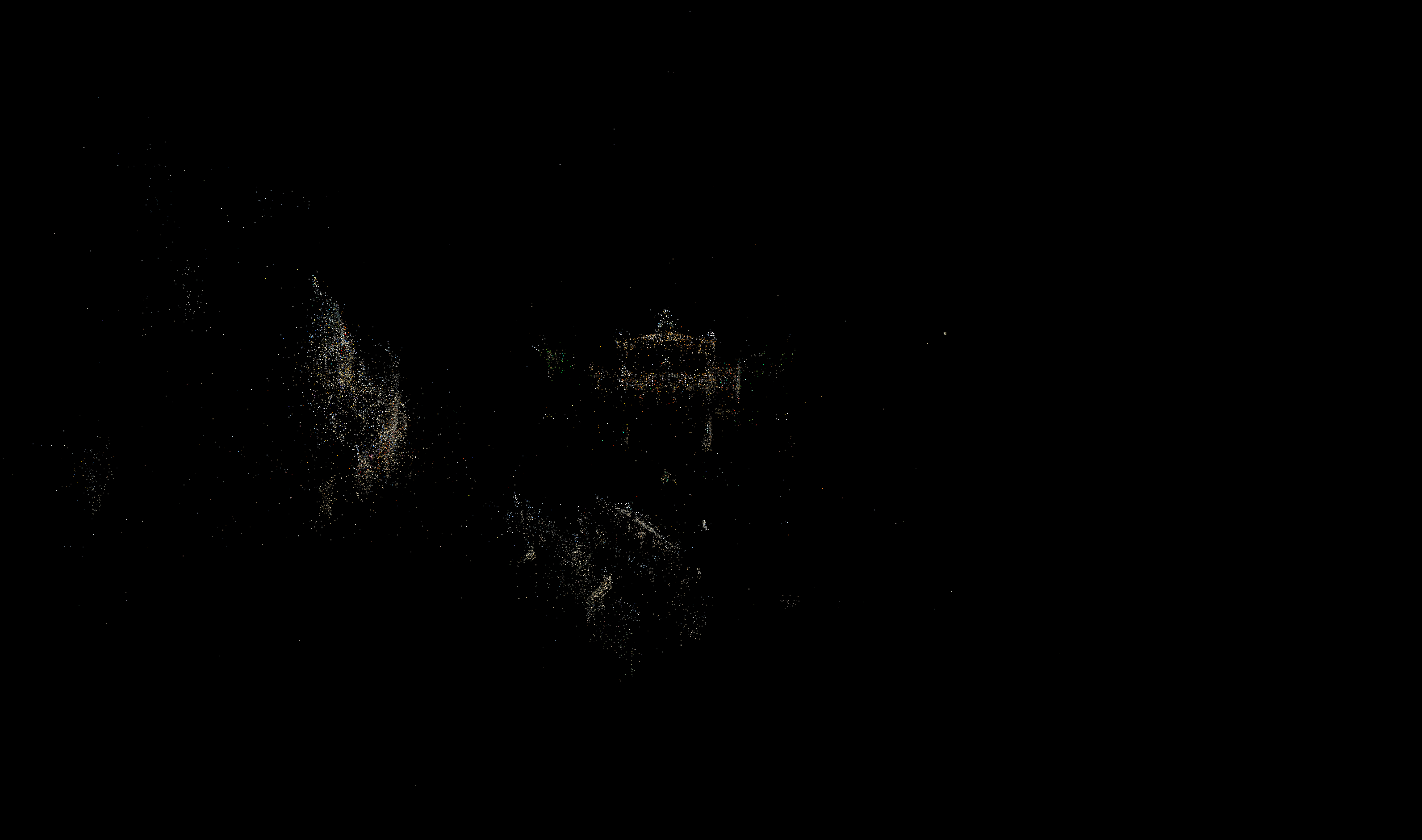} & \includegraphics[width=0.19\textwidth]{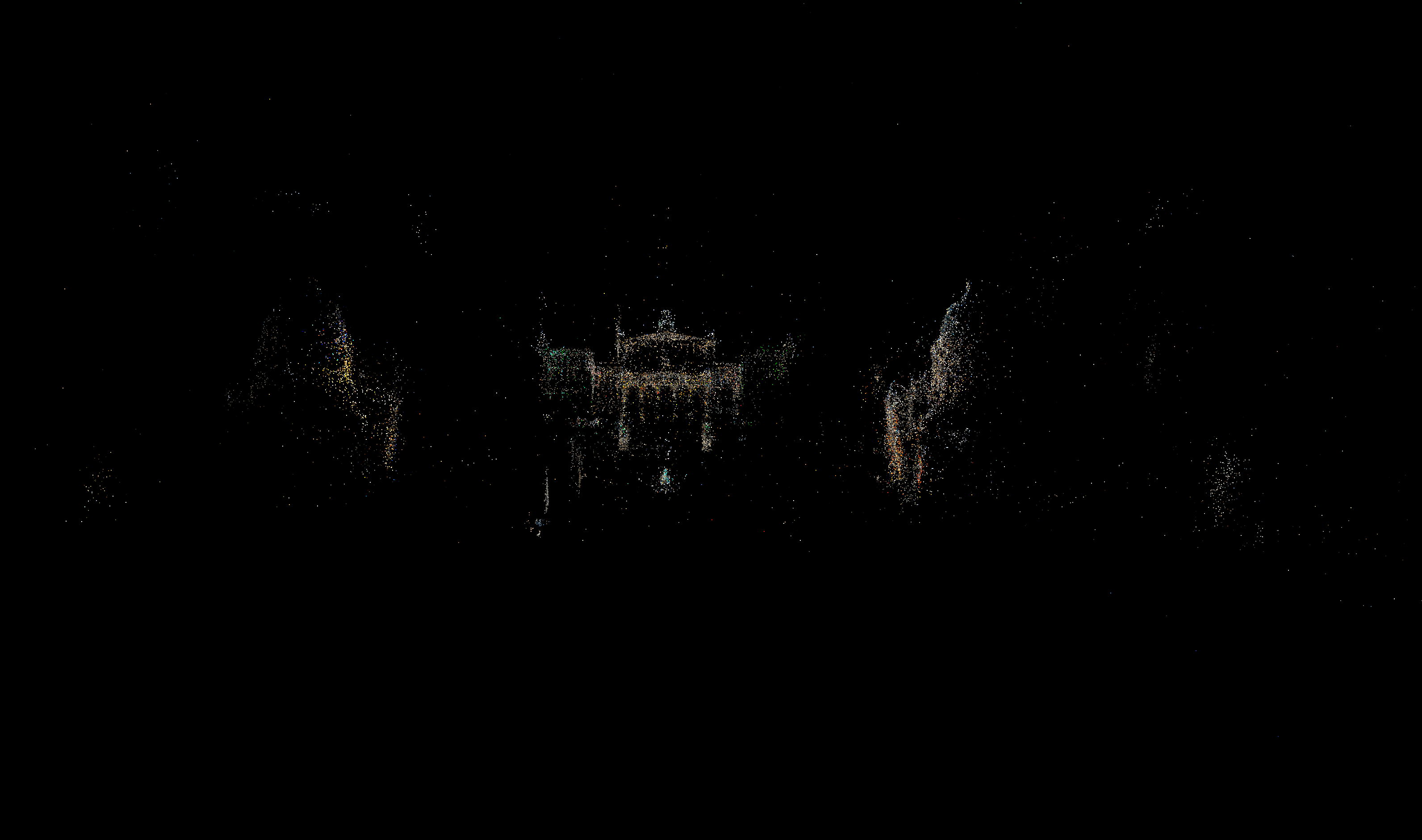} & \includegraphics[width=0.19\textwidth]{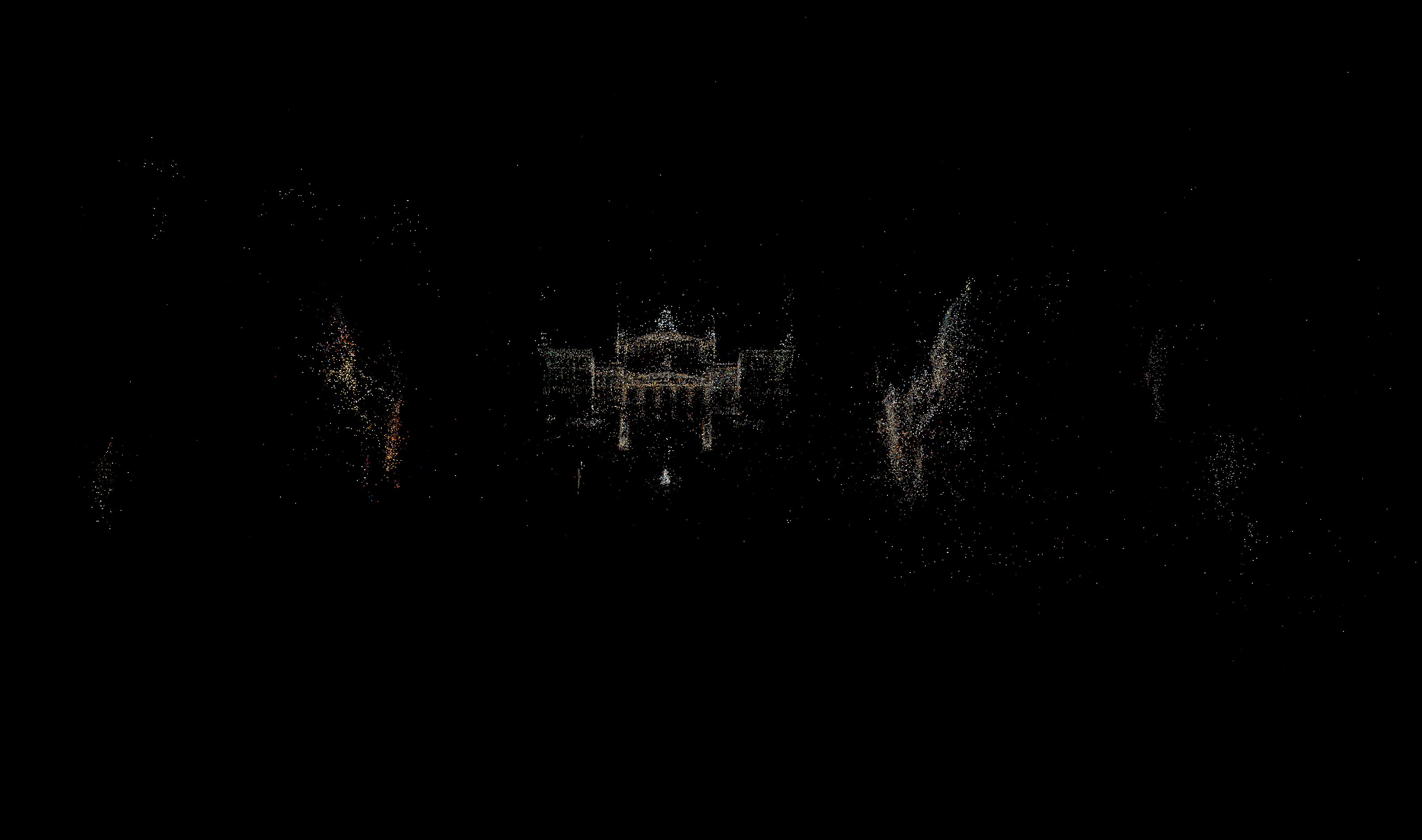} & \includegraphics[width=0.19\textwidth]{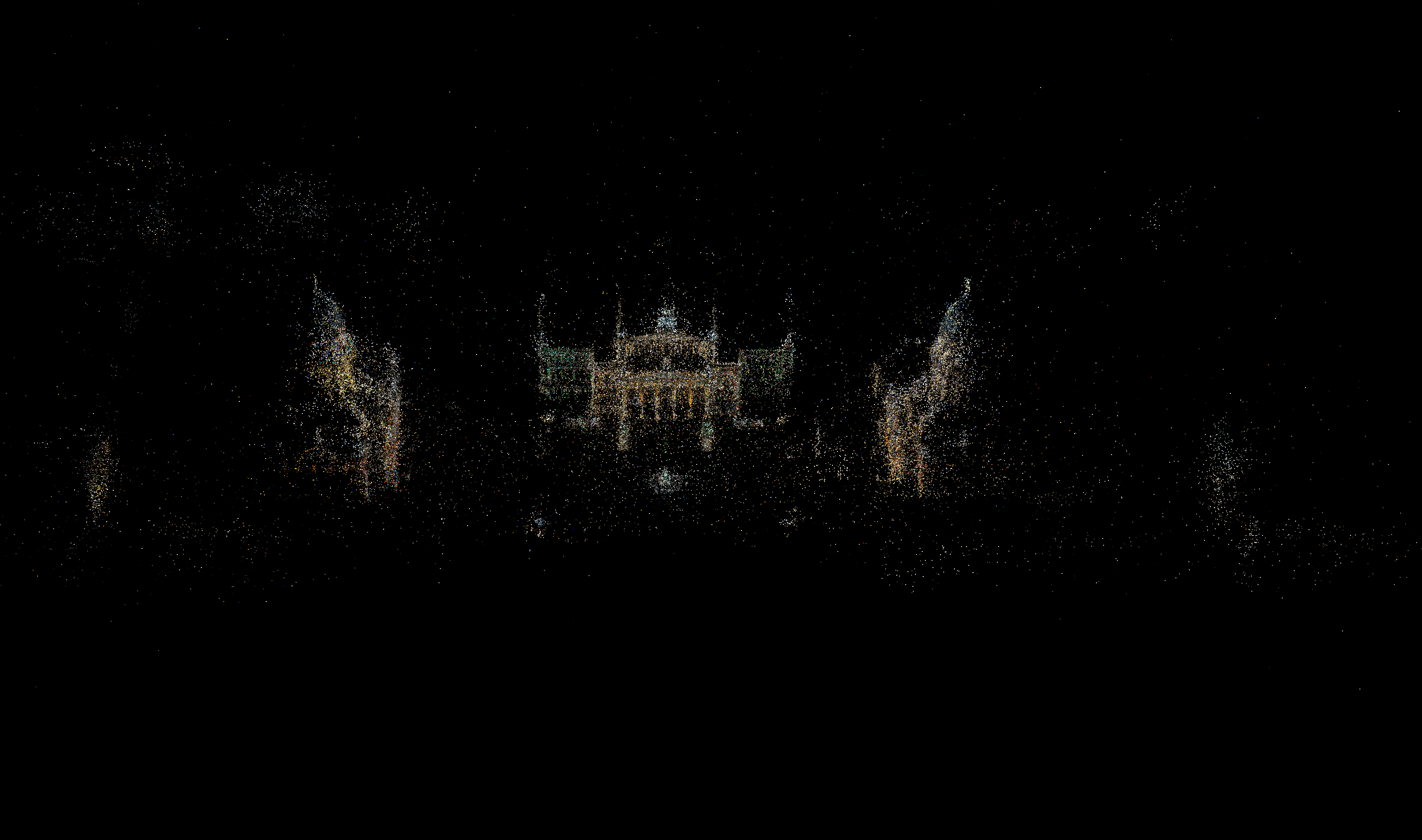}\vspace{20pt} \\
		\multicolumn{5}{c}{\large Tower of London}\\
		BRIEF & SIFT & HardNet & SOSNet & ``Embed"\\ \includegraphics[width=0.19\textwidth]{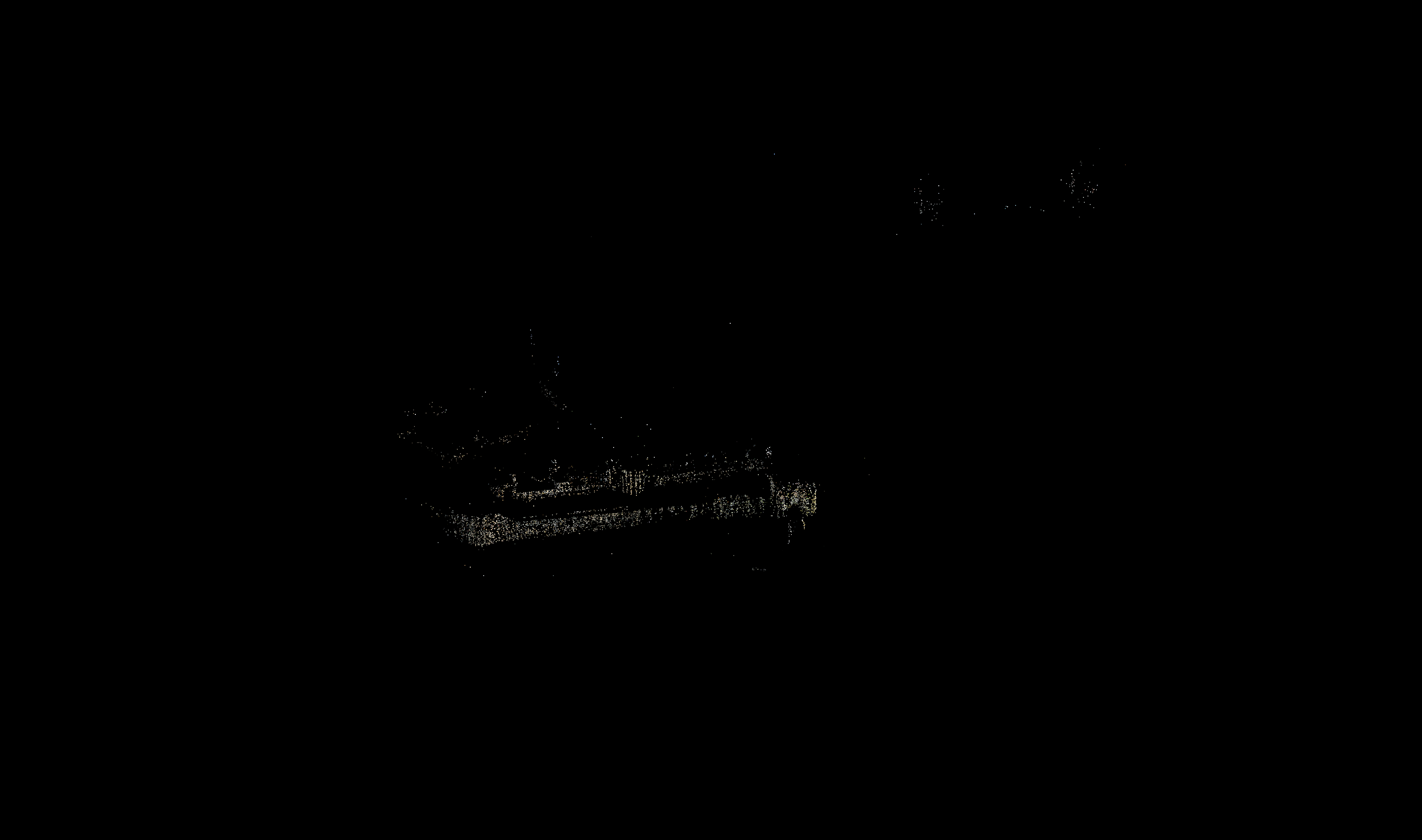} & \includegraphics[width=0.19\textwidth]{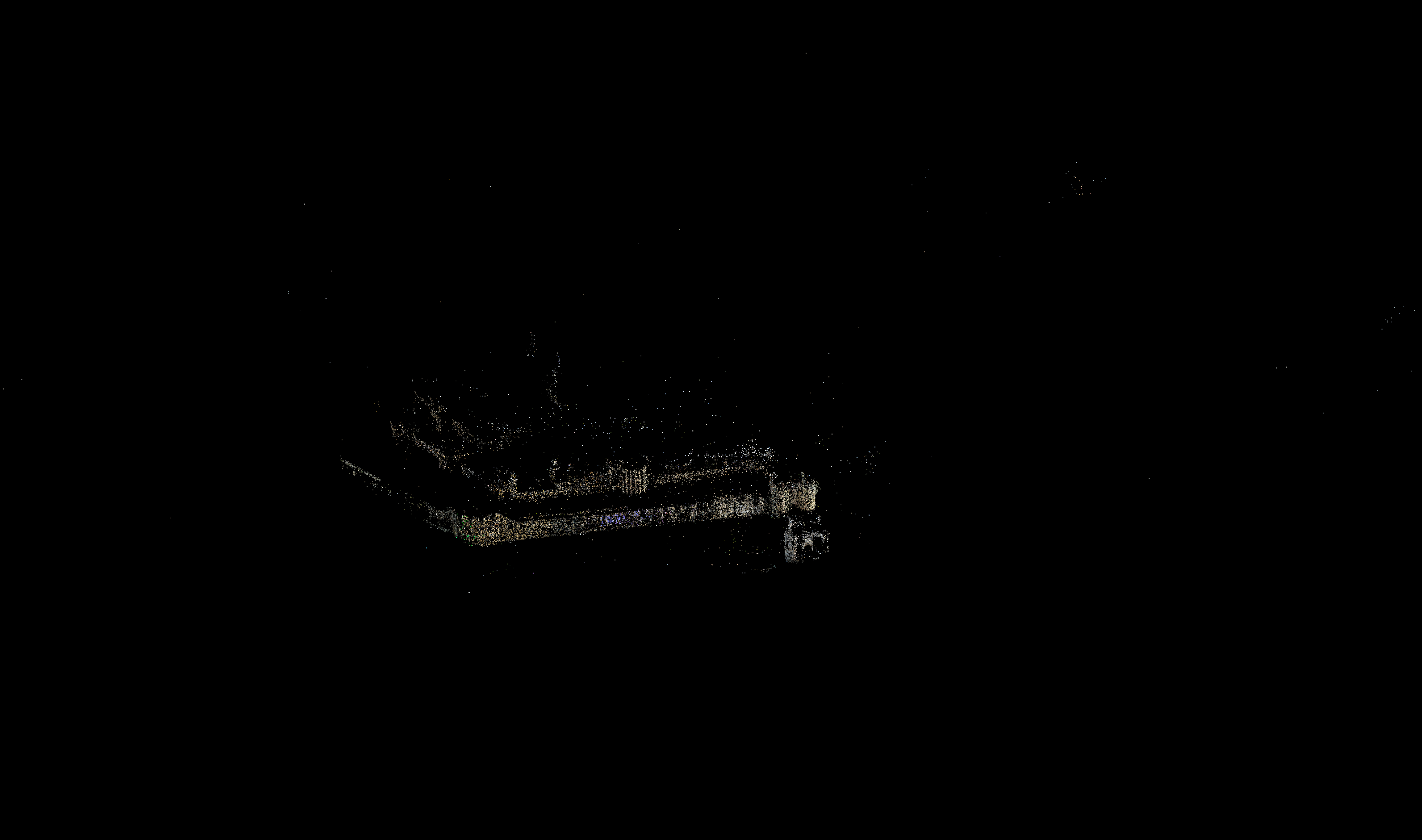} & \includegraphics[width=0.19\textwidth]{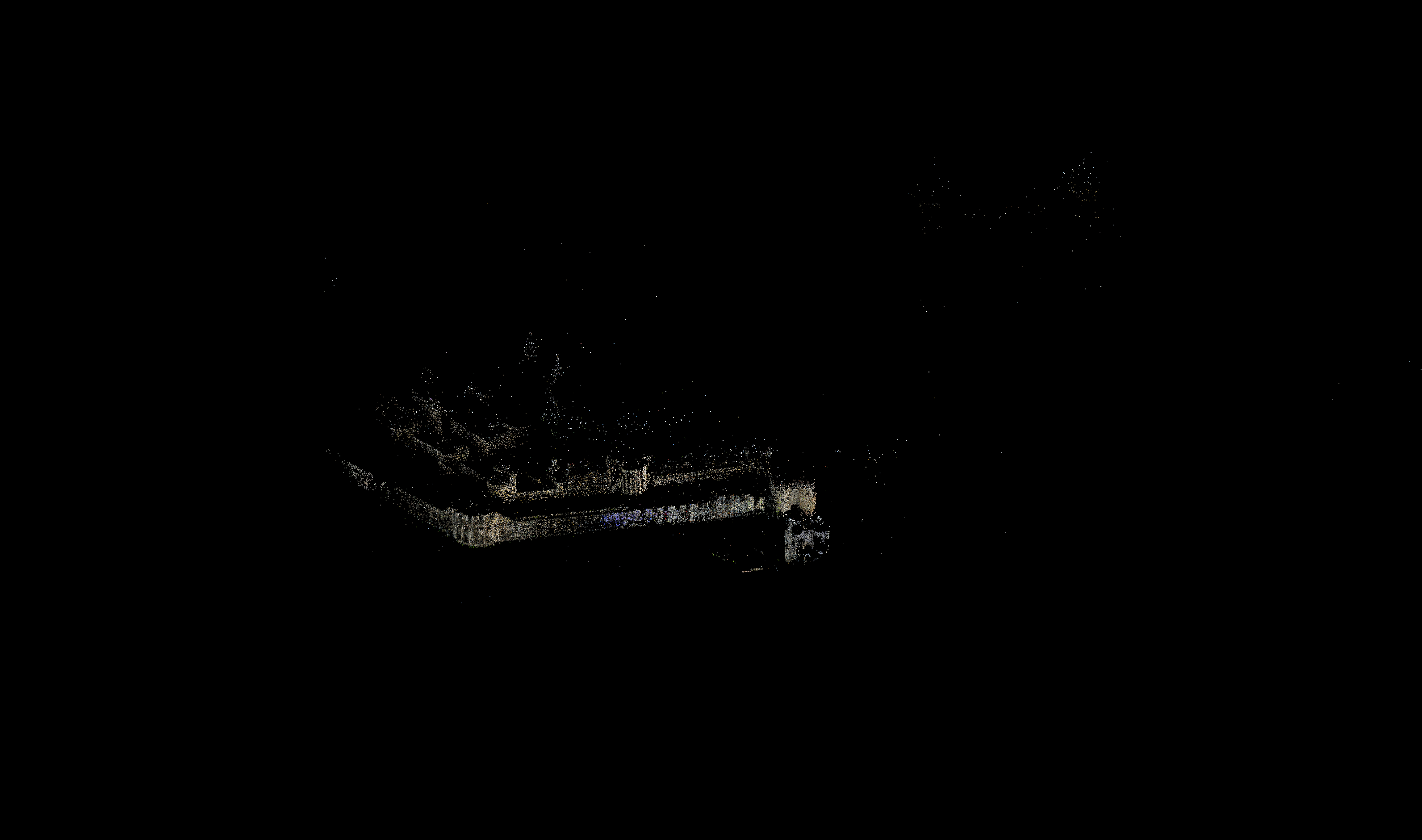} & \includegraphics[width=0.19\textwidth]{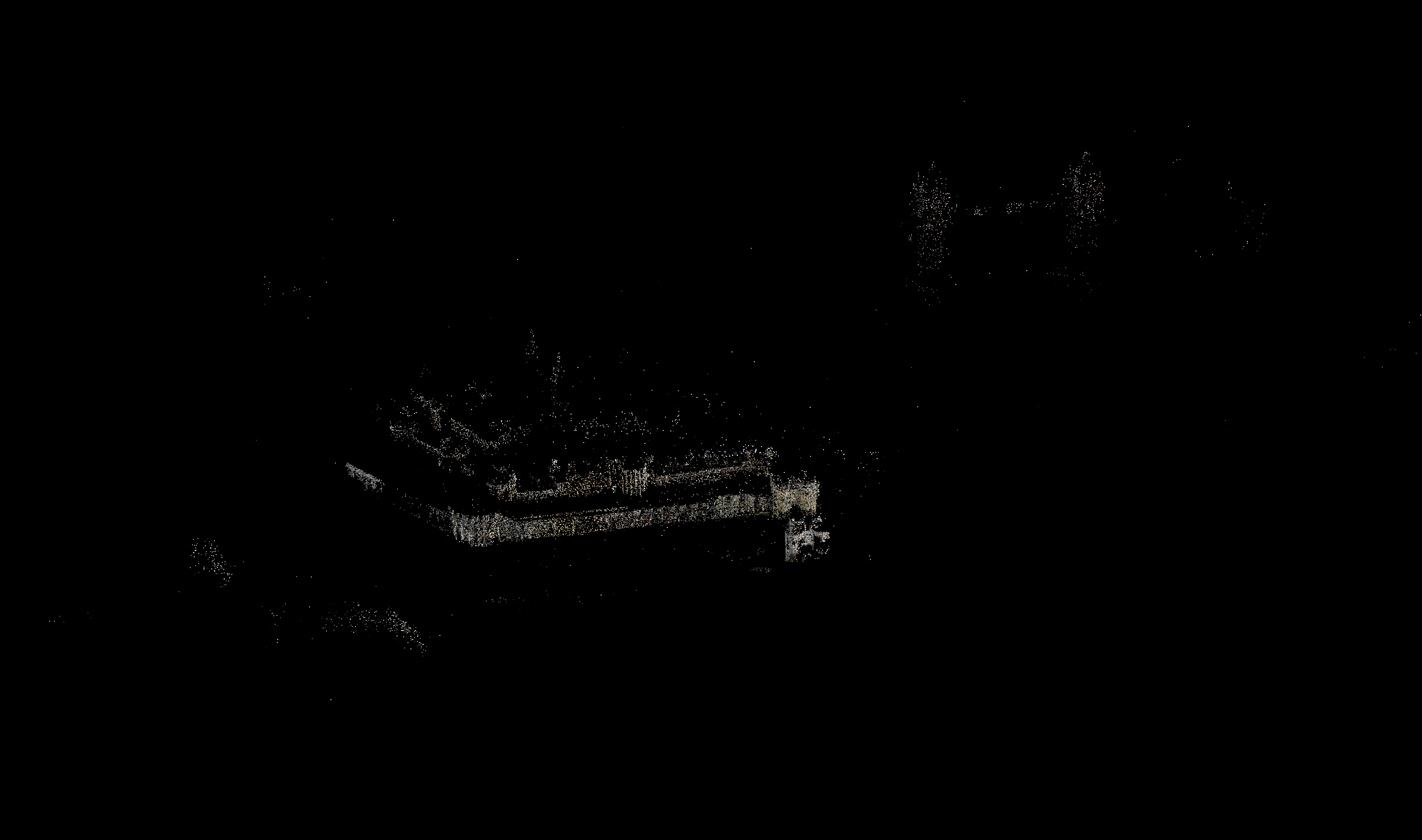} & \includegraphics[width=0.19\textwidth]{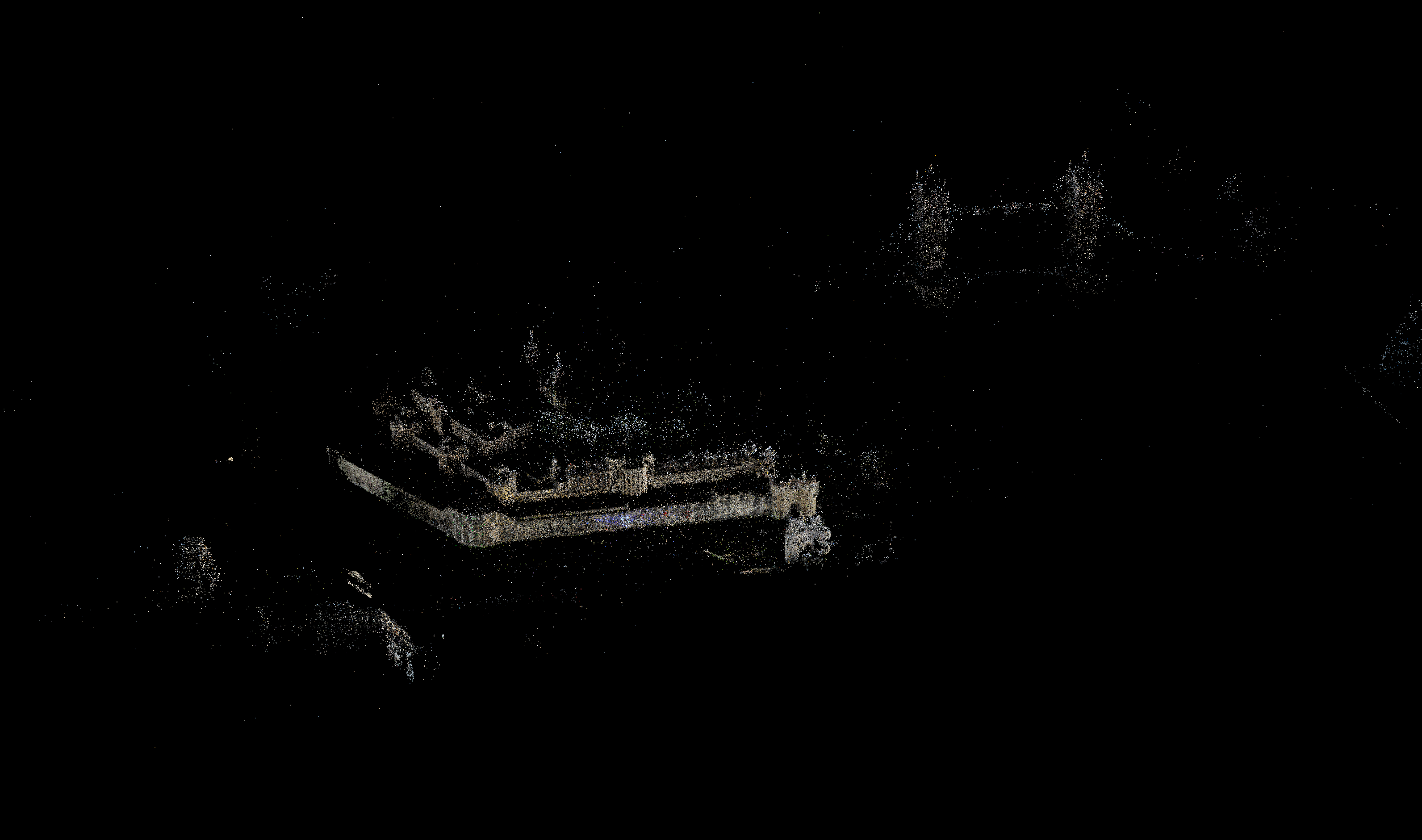} \\
	\end{tabular}
	\caption{{\bf Point cloud visualisations.} We show a qualitative comparison of the point-clouds obtained in the real-world setup (\ie, where each description algorithm only has access to a quarter of images) and the ones obtained using our method to match different descriptors.}
	\label{fig:sfm-qualitative}
\end{sidewaysfigure*}

We show a qualitative comparison of point-clouds in Figure~\ref{fig:sfm-qualitative}.
We compare the real-world point-clouds (\ie, where each description algorithm only has access to a quarter of images) with the proposed crowd-sourced reconstruction.
Our method is able to successfully match descriptors of different types yielding significantly more complete 3D models.
On the most difficult landmark containing strongly symmetric structures and multiple night images (Gendarmenmarkt), we notice that some individual reconstructions are unable to recover the correct ground-truth scene geometry (notably BRIEF and SIFT).

\begin{figure}
	\centering
	\includegraphics[width=.8\columnwidth]{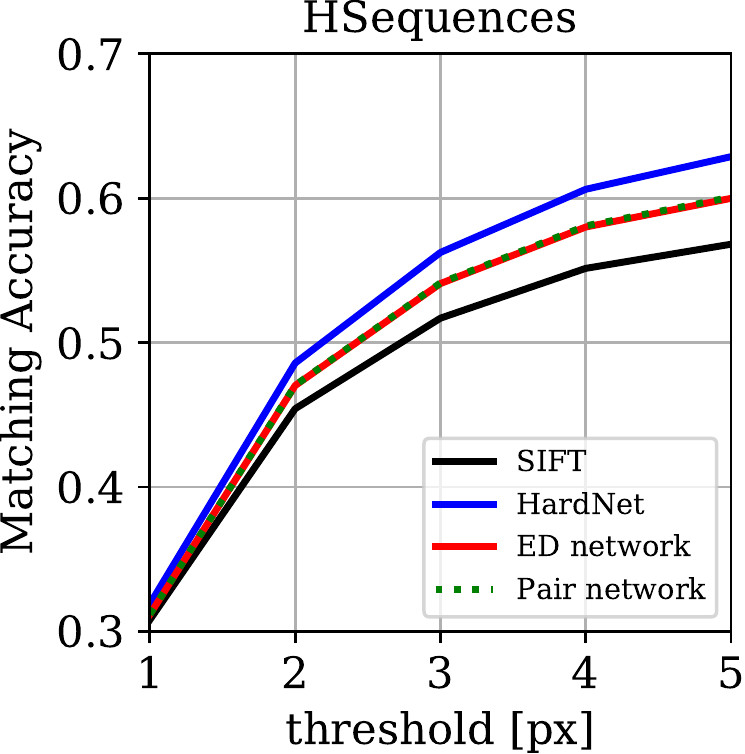}
	\vspace{-7.5pt}
	\caption{{\bf Pair vs. encoder-decoder.} We report the performance of SIFT to HardNet translation on the full sequences from the HPatches dataset. The encoder-decoder (ED) network performs on par with the pair network despite being trained for $4$ description algorithms at once.}
	\label{fig:pair-vs-enc-dec-ablation}
\end{figure}
	\section{Ablation study}

In this section, we study the impact of architecture and losses on our data-driven translation approach.
For this purpose, we consider the full sequences of the well-known HPatches dataset~\cite{Balntas2017HPatches}.
Following the protocol introduced by Dusmanu~\etal~\cite{Dusmanu2019D2}, we report the mean matching accuracy of a mutual nearest-neighbor matcher for different thresholds up to which a match is considered correct.
In each sequence, the first image is treated as query and matched against the other five.
For translation experiments, the query descriptors are translated from a source description algorithm (\eg, SIFT~\cite{Lowe2004Distinctive}) to a target one (\eg, HardNet~\cite{Mishchuk2017Working})) and matched against natively extracted descriptors (\eg, HardNet) in the other images.

\subsection{Naive matching}

\begin{figure}
	\centering
	\includegraphics[width=.75\columnwidth]{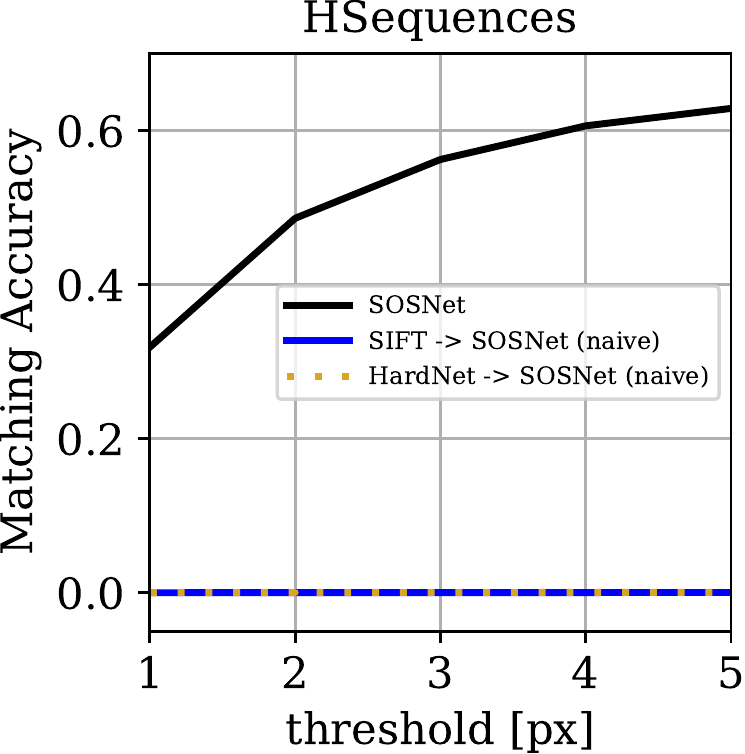}
	\vspace{-7.5pt}
	\caption{{\bf Naive matching.} Matching different descriptors by running nearest neighbor search from one descriptor space to the other does not yield any correct matches.}
	\label{fig:naive-matching}
\end{figure}

We first try naively matching different descriptors by running nearest neighbor search from one descriptor space to the other.
Results are reported in Figure~\ref{fig:naive-matching}.
BRIEF cannot be matched against SOSNet due to the different dimensionality.
SIFT does not yield any correct matches when matched directly against SOSNet.
This is also valid for HardNet, despite using the same backbone architecture and training data as SOSNet.
Thus, it is impossible to naively match different descriptors and, without cross-descriptor matching, the final 3D models would be disconnected.

\subsection{Pair vs. encoder-decoder}

We compare a pair network trained specifically for SIFT to HardNet translation with an encoder-decoder network trained for all $4$ description algorithms.
We use the same dataset and hyper-parameters.
We set the number of weights of the pair network equal to that of the encoder of SIFT concatenated with the decoder of HardNet.
Results are reported in Figure~\ref{fig:pair-vs-enc-dec-ablation}.
The performance of both approaches is similar.
However, the encoder-decoder network can be trained once no matter the number of description algorithms and has the advantage of a joint embedding.

\subsection{Number of description algorithms}

\begin{figure}
	\centering
	\includegraphics[width=.75\columnwidth]{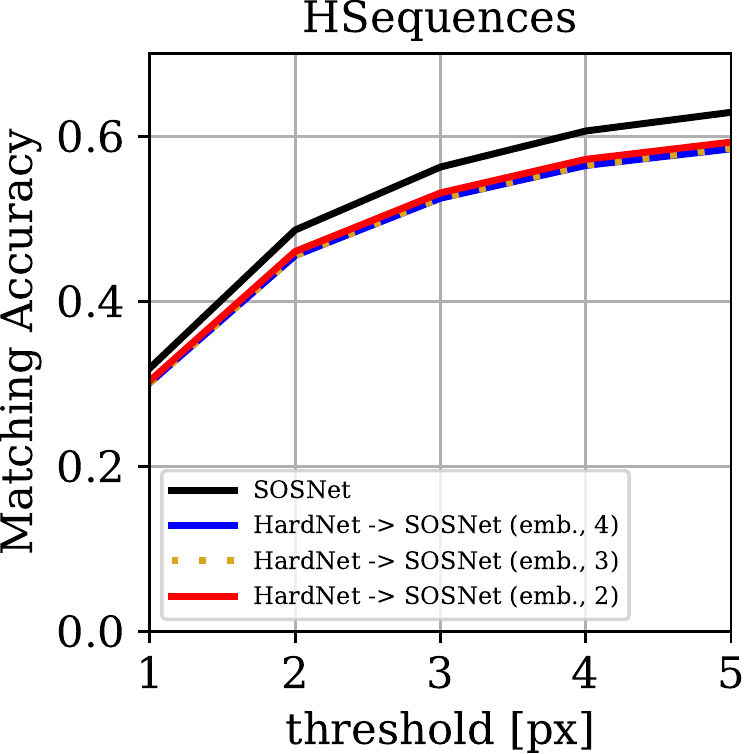}
	\vspace{-7.5pt}
	\caption{{\bf Number of description algorithms.} The performance gain is minimal when training the encoder-decoder approach using the learned descriptors only.}
	\label{fig:num-descriptors}
\end{figure}

In Figure~\ref{fig:num-descriptors}, we show an ablation based on the number of different description algorithm used during training.
We report the matching performance when matching HardNet to SOSNet features in the joint space.
We consider $3$ variants of the encoder-decoder architecture trained with different algorithm subsets: $4$ was trained with all descriptors (BRIEF, SIFT, HardNet, SOSNet), $3$ with SIFT, HardNet, SOSNet, and $2$ only with HardNet and SOSNet.
As can be seen, the performance gain is marginal when training exclusively with the learned methods.
We believe the performance loss when compared to raw descriptors is due to enforcing consistency between different methods.

\subsection{Loss}

We investigate the effect of different losses on the encoder-decoder approach.

\begin{figure}
	\centering
	\includegraphics[width=.75\columnwidth]{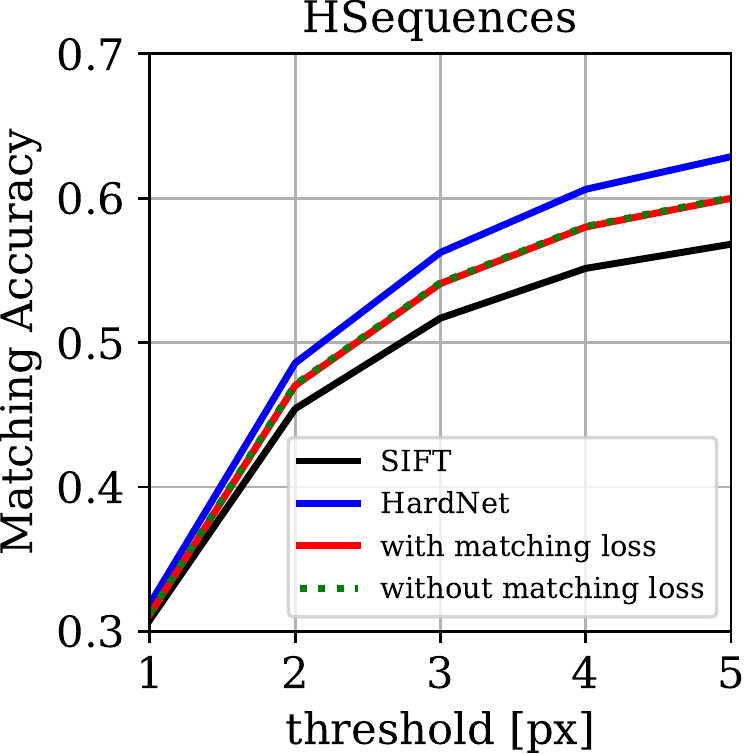}
	\vspace{-7.5pt}
	\caption{{\bf Matching loss.} We report the performance of SIFT to HardNet translation on the full sequences from the HPatches dataset. The matching loss makes the joint space suitable for establishing correspondences and does not have a negative impact on pair-wise translation.}
	\label{fig:matching-loss}
\end{figure}

\begin{figure}
	\begin{minipage}{0.48\columnwidth}
		\centering
		\includegraphics[width=\textwidth]{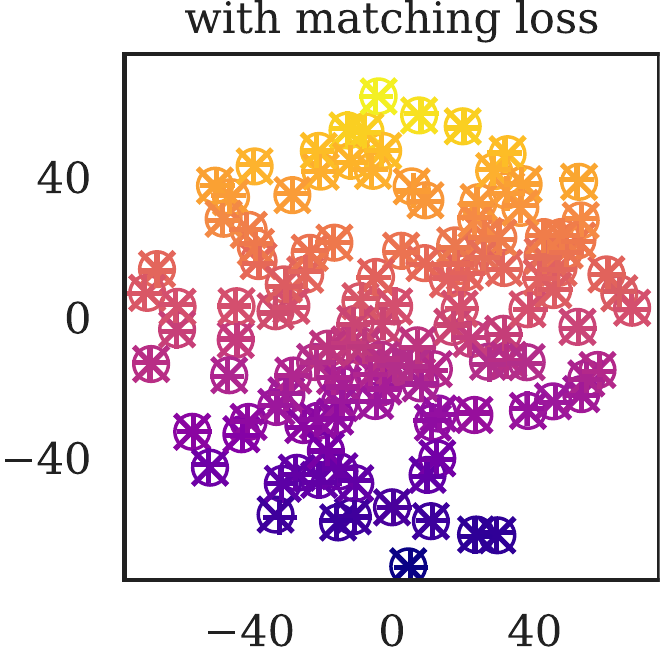}
	\end{minipage}
	\hfill
	\begin{minipage}{0.48\columnwidth}
		\centering
		\includegraphics[width=\textwidth]{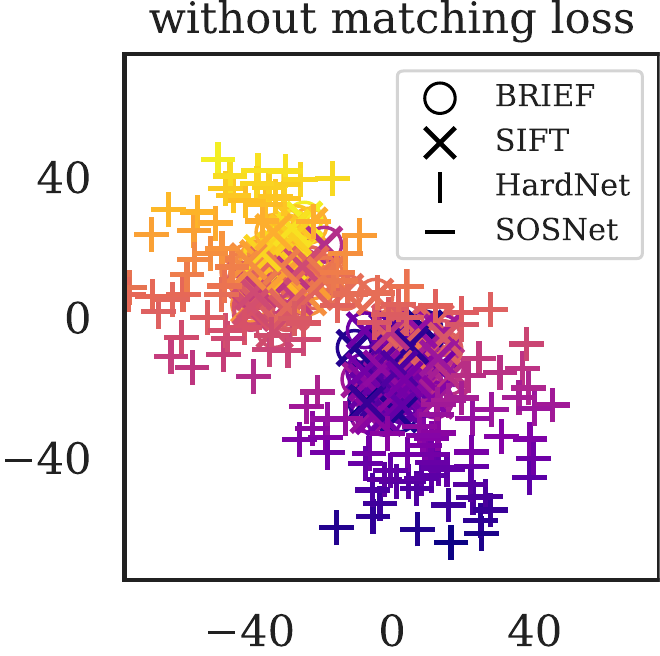}
	\end{minipage}
	\caption{{\bf t-SNE visualization of the joint space.} We visualize the embedding of $128$ training patches with different description algorithms. Without matching loss, the handcrafted and learned descriptors are not coherent.}
	\label{fig:t-sne-emb}
\end{figure}

\customparagraph{Matching loss.}
We first study the usefulness of the matching loss.
For this purpose, we randomly select $512$ patches from our training dataset.
We extract the $4$ descriptors from each patch and map them to the joint space using their associated encoders.
Finally, we use t-SNE~\cite{Maaten2008Visualizing} for visualization.
For clarity, we only plot $128$ descriptors of each type in Figure~\ref{fig:t-sne-emb}.
Training without a matching loss yields a representation that cannot be used for cross-descriptor matching.
However, HardNet and SOSNet seem coherent suggesting that learned descriptors focus on similar information.
When leveraging the matching loss, all descriptors are well aligned.
Furthermore, as shown by Figure~\ref{fig:matching-loss}, enforcing matchability in the joint space does not have a significant impact on the pair-wise translation.

\customparagraph{Final loss.}
We study three variations of the final loss used for training.
First, we consider the formulation presented in the main paper which takes into account interactions between all encoders and decoders:
\begin{equation}
	\mathcal{L}^T_\text{quadratic} = \frac{1}{\lvert \mathcal{A} \rvert^2} \sum_{A_i, A_j \in \mathcal{A}^2} \mathcal{L}^T_{i \rightarrow j} \enspace ,
\end{equation}
\begin{equation}
	\mathcal{L}^M_\text{quadratic} = \frac{1}{\lvert \mathcal{A} \rvert^2} \sum_{A_i, A_j \in \mathcal{A}^2} \mathcal{L}^M_{i \rightarrow j}.
\end{equation}
Second, the translation loss can be replaced by the traditional auto-encoder loss~\cite{Bourlard1988Auto} defined as:
\begin{equation}
	\mathcal{L}^T_\text{auto-encoder} = \frac{1}{\lvert \mathcal{A} \rvert} \sum_{A_i \in \mathcal{A}} \mathcal{L}^T_{i \rightarrow i} \enspace ,
\end{equation}
while the matching loss is kept as is:
\begin{equation}
	\mathcal{L}^M_\text{auto-encoder} = \frac{1}{\lvert \mathcal{A} \rvert^2} \sum_{A_i, A_j \in \mathcal{A}^2} \mathcal{L}^M_{i \rightarrow j} \enspace .
\end{equation}
Third, we propose a linear relaxation of our losses as:
\begin{equation}
	\mathcal{L}^T_\text{linear} = \frac{1}{\lvert \mathcal{A} \rvert} \sum_{A_i \in \mathcal{A}} \mathcal{L}^T_{i \rightarrow \sigma(i)} \enspace ,
\end{equation}
\begin{equation}
	\mathcal{L}^M_\text{linear} = \frac{1}{\lvert \mathcal{A} \rvert} \sum_{A_i \in \mathcal{A}} \mathcal{L}^M_{i \rightarrow \sigma(i)} \enspace ,
\end{equation}
with $\sigma$ a permutation of $\{1, \dots, \lvert \mathcal{A} \rvert\}$ chosen randomly at every optimization iteration.
In each case, the final loss is a weighted sum of the translation and matching losses: 
\begin{equation}
	\mathcal{L}_* = \mathcal{L}^T_* + \alpha \mathcal{L}^M_* \enspace .
\end{equation}

\begin{figure}
	\centering
	\includegraphics[width=.75\columnwidth]{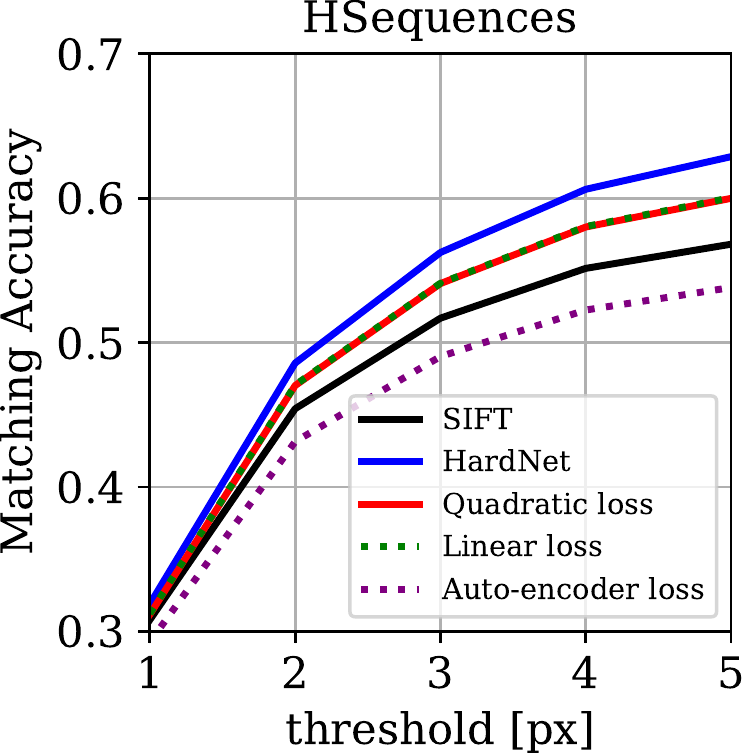}
	\vspace{-7.5pt}
	\caption{{\bf Loss ablation.} We report the performance of SIFT to HardNet translation on the full sequences from the HPatches dataset. We train our encoder-decoder approach with different losses. Taking into account the interaction between encoders and decoders of different description algorithms is required for better performance.}
	\label{fig:loss-ablation}
\end{figure}

We train the encoder-decoder approach for all $4$ description algorithms (\ie, BRIEF, SIFT, HardNet, SOSNet) with the same architecture and hyper-parameters using each loss independently.
We report the results for SIFT to HardNet translation in Figure~\ref{fig:loss-ablation}.
The auto-encoder loss performs poorly as it does not consider the interaction between the encoders and decoders of different description algorithms.
To speed up the training process (especially for larger collections of algorithms), one can use the linear variant of our losses as it yields similar performance.

\setlength{\tabcolsep}{2.0pt}
\setlength{\tabcolsep}{\tabcolsepdefault}
	
	\FloatBarrier	
	{\small
		\bibliographystyle{ieee_fullname}
		\bibliography{shortstrings,references}
	}
	
\end{document}